\documentclass[runningheads]{llncs}

\usepackage{eccv}

\usepackage{eccvabbrv}

\usepackage{graphicx}
\usepackage{booktabs}

\usepackage{amsmath}
\usepackage{multirow}
\usepackage{multicol}
\usepackage{adjustbox}
\usepackage{colortbl}
\usepackage{algorithm}
\usepackage{algorithmic}
\usepackage{transparent}
\usepackage{tikz}
\usepackage{tikz,tikz-3dplot}
\usetikzlibrary{calc,fit,spy}
\usepackage{bm}

\usepackage{tabularx}

\definecolor{mycolor}{RGB}{215, 215, 215}
\definecolor{myrowcolor}{RGB}{230, 230, 230}
\definecolor{mycolor2}{RGB}{225, 225, 255}

\usepackage[accsupp]{axessibility}  

\usepackage{hyperref}

\usepackage{orcidlink}

\newcommand{\ours}{{SNM-VFI}\xspace}

\begin{document}

\title{SNM-VFI: Symmetric Nonlinear Motion-Guided \\ Generative Video Frame Interpolation} 

\titlerunning{SNM-VFI: Motion-Guided Generative VFI}

\author{Jisoo Jeong\inst{1} \and Hong Cai\inst{1} \and Jamie Menjay Lin\inst{2} \and Hanno Ackermann\inst{1} \and \\ Hyeonjun Sim\inst{3}$^{\ddagger}$ \and Yinhao Zhu\inst{1} \and Yunxiao Shi\inst{1} \and Fatih Porikli\inst{1}
}

\authorrunning{J. Jeong et al.}

\institute{Qualcomm AI Research\inst{1}$^{\dagger}$, 
Qualcomm Technologies, Inc.\inst{2}, and Google\inst{3} \\
\email{\{jisojeon, hongcai, jmlin, hackerma\}@qti.qualcomm.com \\ flhy5836@gmail.com, \{yinhaoz, yunxshi, fporikli\}@qti.qualcomm.com}
}
\vspace{-14pt}

\maketitle

\begin{abstract}
    We propose \textbf{S}ymmetric \textbf{N}onlinear \textbf{M}otion-guided Generative \textbf{V}ideo \textbf{F}rame \textbf{I}nterpolation (\ours), a training-free framework for motion-controllable generative video frame interpolation with pre-trained optical flow and video diffusion models.
    Unlike conventional diffusion-based VFI methods that synthesize intermediate frames from random noise, \ours guides the generative process with correspondence-aware frames produced by a symmetric nonlinear motion model. Specifically, we first utilize a pre-trained optical flow model to construct multi-frame nonlinear flow-based intermediate frames and confidence maps. These flow-guided frames are then encoded as latent priors to initialize and iteratively guide a pre-trained Video Diffusion model, enabling the diffusion model to preserve dense motion correspondence while improving perceptual realism. To further enhance output quality, we employ confidence maps to fuse structurally reliable flow-based predictions with diffusion-generated details in uncertain regions such as occlusions and object boundaries. Extensive evaluations on challenging benchmarks, including DAVIS, Sintel, and KITTI, demonstrate that \ours achieves strong perceptual quality, competitive reconstruction accuracy, and robust temporal coherence across diverse motion scenarios.
\end{abstract}

\vspace{-10pt}


\vspace{-2mm}
\section{Introduction}
\label{sec:intro}

{\let\thefootnote\relax\footnotetext{{
\hspace{-6.5mm} $\dagger$ Qualcomm AI Research is an initiative of Qualcomm Technologies, Inc. 
}}}

{\let\thefootnote\relax\footnotetext{{
\hspace{-6.5mm} $\ddagger$ This work was done while at Qualcomm Technologies, Inc.
}}}

\begin{figure}[ht]
\begin{center}$
\centering
\begin{tabular}{c}
\includegraphics[width=\textwidth]{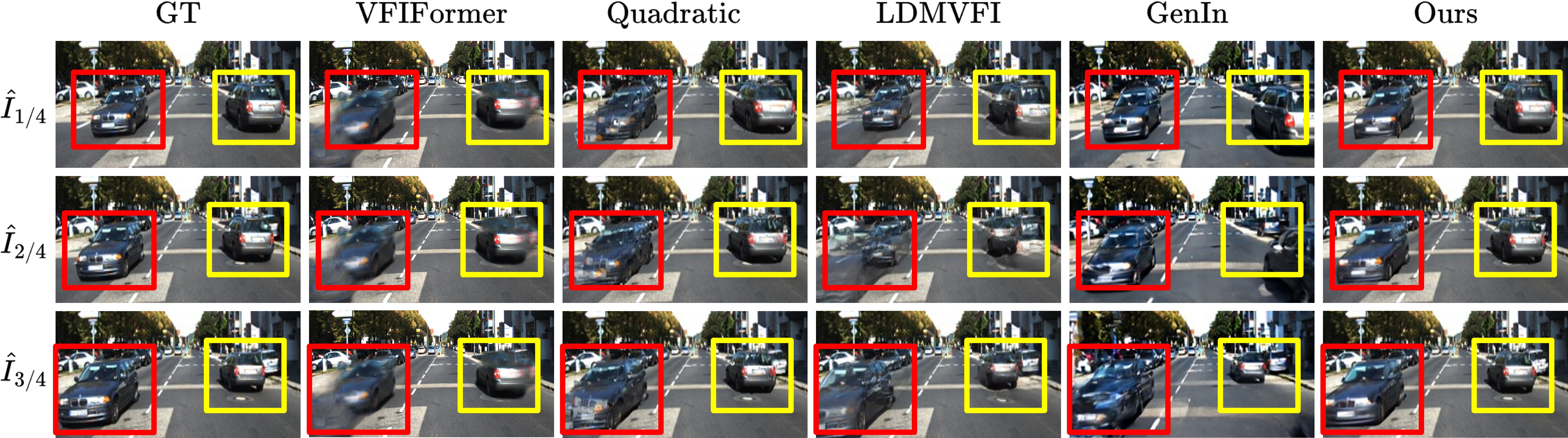} \\
\vspace{-30pt}
\end{tabular}$
\end{center}
\caption{Video Frame Interpolation results ($\hat{I}_{1/4}, \hat{I}_{2/4}, \hat{I}_{3/4}$) on the KITTI dataset. VFIFormer~\cite{lu2022video} shows interpolation results produced using a linear flow-based method, while LDMVFI~\cite{danier2024ldmvfi} and GenIn~\cite{wang2024generative} present results from diffusion-based VFI approaches. Quadratic~\cite{xu2019quadratic} represents a multi-frame VFI method, and the final column displays the results generated by our approach.  
}
\vspace{-15pt}
\label{fig:result_intro}
\end{figure}

Flow-based video frame interpolation (VFI) methods~\cite{kong2022ifrnet, han2022realflow, zhang2023extracting, li2023amt, jeong2024ocai} explicitly model pixel-level correspondences by estimating optical flow~\cite{hui2019lightweight, teed2020raft, huang2022flowformer} between frames. 
While effective in capturing motion trajectories, these approaches are fundamentally limited by their reliance on a linear motion assumption, \ie constant velocity between frames. This simplification restricts their ability to handle complex, accelerated, or nonlinear motion patterns that frequently occur in real-world scenes~\cite{xu2019quadratic, liu2020enhanced}. Moreover, flow-based methods often struggle with occlusions, imprecise object boundaries, and non-rigid deformations (as shown in the $2^{nd}$ and $3^{rd}$ columns of Fig~\ref{fig:result_intro}).

In contrast, recent advances in generative modeling have introduced diffusion models into the domain of Video Frame Interpolation, demonstrating impressive capabilities in producing visually compelling and perceptually realistic video content~\cite{danier2024ldmvfi, feng2024explorative, wang2024generative, jain2024video, zhang2025motion}.
However, these models typically operate on a per-frame or short-sequence basis, making them prone to temporal inconsistencies and incapable of capturing dense motion correspondences essential for smooth and coherent video synthesis (as shown in the $4^{th}$ and $5^{th}$ columns of Fig~\ref{fig:result_intro}). These limitations highlight the necessity for a video frame interpolation framework that effectively integrates the perceptual richness of diffusion models with the precise motion guidance and correspondence modeling offered by flow-based approaches.

In this paper, we propose \textbf{S}ymmetric \textbf{N}onlinear \textbf{M}otion Guided Generative \textbf{V}ideo \textbf{F}rame \textbf{I}nterpolation (\ours), a novel framework that unifies the complementary strengths of both paradigms to achieve superior interpolation quality with robust temporal coherence. Importantly, \ours operates efficiently by leveraging pre-trained optical flow models~\cite{teed2020raft, xu2022gmflow, huang2022flowformer} and off-the-shelf video diffusion models~\cite{blattmann2023stable, feng2024explorative, wang2024generative}, eliminating the need for task-specific training of the generative component.

\begin{figure}[t]
\begin{center}$
\centering
\begin{tabular}{c}
\includegraphics[width=\textwidth]{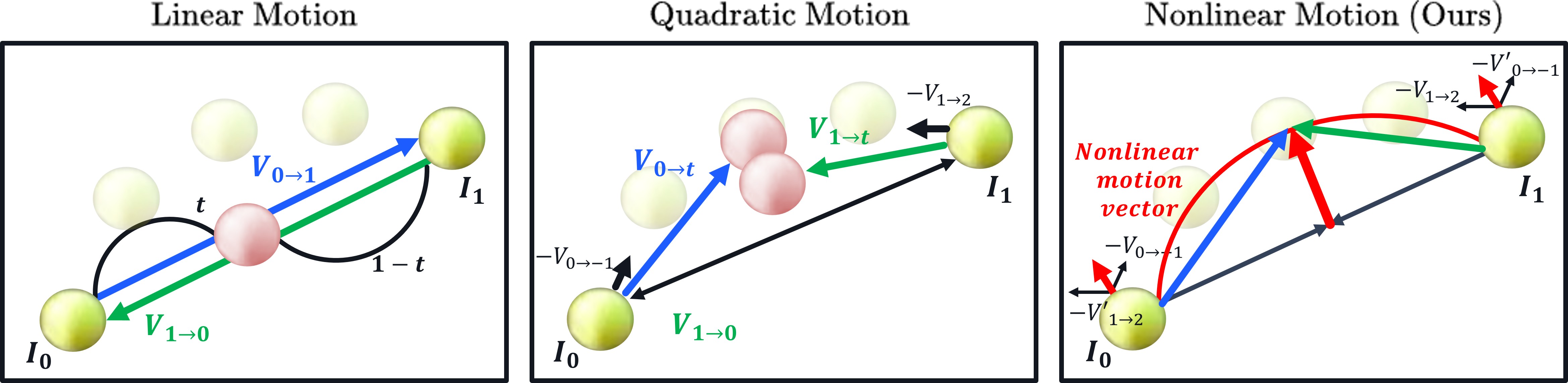} 

\end{tabular}$
\end{center} 
\vspace{-18pt}
\caption{Linear motion (left), Quadratic Motion (middle), and Symmetric Nonlinear Motion (right). Linear motion assumes constant velocity, causing the optical flows ($V_{0 \rightarrow t}$ and $V_{1 \rightarrow t}$) to meet at a single point, but failing to capture complex real-world dynamics. Quadratic motion incorporates velocity and acceleration at $I_{0}$ and $I_{1}$ by leveraging additional optical flows ($V_{0 \rightarrow -1}$, $V_{1 \rightarrow 2}$), yet forward and backward flows do not always align at an intermediate point in this framework. In contrast, our symmetric nonlinear motion approach also uses ($V_{0 \rightarrow -1}$, $V_{1 \rightarrow 2}$), but applies them symmetrically at both $I_{0}$ and $I_{1}$ while explicitly accounting for occlusion. This enables more accurate nonlinear modeling and ensures alignment of flows at any intermediate point. }
\label{fig:nonlinear}
\vspace{-7pt}
\end{figure}

To make a pre-trained video diffusion model usable for low-level temporal visual processing, we first construct reliable motion-conditioned generative priors via symmetric nonlinear flow interpolation.
As illustrated in Fig.~\ref{fig:nonlinear} (left), linear motion models often fail to capture the complexity of real-world dynamics. Previous attempts~\cite{xu2019quadratic, liu2020enhanced} incorporated additional frames (e.g., $I_{-1}$ or $I_{2}$) to improve motion estimation, but frequently produce misaligned flow vectors ($V_{0 \rightarrow t}$ and $V_{1 \rightarrow t}$), resulting in blur artifacts (Fig.~\ref{fig:nonlinear} middle and Fig.~\ref{fig:result_intro} the rear-right of the car in the yellow box of Quadratic). Moreover, these methods do not account for occlusions, further degrading interpolation quality. Our symmetric approach resolves this by jointly utilizing both past (-$V_{0 \rightarrow -1}$) and next (-$V_{1 \rightarrow 2}$) optical flows to compute two optical flows ($V_{0 \rightarrow t}$ and $V_{1 \rightarrow t}$). This design ensures robust alignment of warped flows at any intermediate point (Fig.~\ref{fig:nonlinear} right) and effectively handles occlusions, leading to significantly improved interpolation fidelity.

In addition to the more powerful motion modeling, our framework tackles the issue of temporal inconsistency that commonly affects diffusion-based video generation. Conventional diffusion-based VFI methods typically begin from random noise, which often leads to outputs lacking long-range temporal coherence. To mitigate this, we introduce flow-guided intermediate frames as both the initialization and dynamic guidance throughout the diffusion process. This integration of correspondence-aware motion priors ensures that the generated frames maintain temporal alignment while taking advantage of the high perceptual quality offered by generative models. Finally, we combine the best of both worlds and devise a fusion strategy to combine the outputs from our symmetric nonlinear motion-based VFI and our flow-guided diffusion-based VFI. More specifically, we replace the low-confidence areas in our flow-based interpolated frame, such as occluded regions and object boundaries where correspondences may not be available, with high-quality diffusion-generated pixels. As we shall see, this fusion strategy leads to measurable gains in perceptual quality.

In summary, our main contributions are as follows:
\begin{itemize}

\item We propose \ours, a training-free generative VFI framework that adapts pre-trained video diffusion models for low-level temporal visual processing using explicit motion priors.

\item We introduce flow-guided latent initialization and iterative diffusion guidance, which inject correspondence-aware motion cues throughout denoising.



\item We design a symmetric nonlinear motion prior that produces reliable flow-based intermediate frames and confidence maps, enabling confidence-aware fusion of accurate flow predictions and realistic diffusion-generated details in uncertain regions.



\item We validate \ours through extensive experiments on DAVIS, Sintel, and KITTI, demonstrating strong perceptual quality, competitive reconstruction accuracy, and robust temporal coherence across diverse motion scenarios.

\end{itemize}

\section{Related Work}
\label{sec:related}


\subsection{Flow-Based Video Frame Interpolation (VFI)}

Most flow-based video frame interpolation techniques estimate optical flow between input frames ($I_0$ and $I_1$) and warp the frames to synthesize intermediate frames. These methods can be categorized by their warping strategy: backward warping ($w_{b}$), which maps pixels from the target frame back to the source frames, and forward warping ($w_{f}$), which projects source pixels into the target frame.


\noindent \textbf{Backward-warping-based VFI:} 
Backward warping methods~\cite{kong2022ifrnet, huang2022real, lu2022video, li2023amt, zhang2023extracting, seo2025bim} typically involves predicting two optical flows ($V_{t \rightarrow 0}$ and $V_{t \rightarrow 1}$) from the target intermediate frame ($I_{t}$) to the input frames ($I_{0}$ and $I_{1}$) using a specifically trained VFI network. These flows are then used to perform a backward warping operation on $I_0$ and $I_1$ to obtain warped images ($w_{b}(I_{0},V_{t \rightarrow 0}), w_{b}(I_{1},V_{t \rightarrow 1})$). Finally, a weighting mask is employed in a network to blend these warped images ($\hat{I}_{t} = M \cdot w_{b}(I_{0},V_{t \rightarrow 0}) + (1-M) \cdot w_{b}(I_{1},V_{t \rightarrow 1}$)). A limitation of this approach is that the optical flows ($V_{t \rightarrow 0}$ and $V_{t \rightarrow 1}$) are estimated without direct access to the intermediate frame ($I_{t}$), which can result in inaccurate flow predictions, especially in cases involving large motion. This often leads to degraded interpolation quality for fast-moving or non-rigid objects.

\noindent \textbf{Forward-warping-based VFI:} 
Forward warping methods~\cite{niklaus2020softmax, han2022realflow, jeong2024ocai} estimate bidirectional optical flows ($V_{0 \rightarrow 1}$ and $V_{1 \rightarrow 0}$) between two input frames ($I_{0}$ and $I_{1}$), and approximate $V_{0 \rightarrow t}$ and $V_{1 \rightarrow t}$ via linear scaling: $t \cdot V_{0 \rightarrow t}$ and $(1-t) \cdot V_{1 \rightarrow t}$. Pixels are then projected into the intermediate frame using forward warping ($w_{f}(I_{0},V_{0 \rightarrow t}), w_{f}(I_{1},V_{1 \rightarrow t}$)). While effective for large motion due to accurate flow estimation, forward warping faces two key challenges: 1) the conflict problem, where multiple source pixels map to the same target location, and 2) the hole problem, where certain target pixels receive no source mapping. The conflict issue is typically addressed by incorporating auxiliary cues such as learning weight map, depth, or occlusion, while holes are filled using warped outputs from the alternate source frame. These strategies enhance robustness, particularly in scenarios involving complex or large-scale motion.

\noindent \textbf{Multi-frame VFI: }
To mitigate the limitations of the linear motion assumption inherent in two-frame-based interpolation, several methods have explored the use of multiple frames to generate more accurate intermediate results. One such approach~\cite{shang2023joint} structurally extends the model to accept multiple input frames (\eg, $T$ frames) and produces a larger set of output frames (\eg, $S$ $\times$ $T$), thereby capturing richer temporal dynamics and improving interpolation quality. Alternatively, quadratic motion estimation techniques~\cite{xu2019quadratic, liu2020enhanced} aim to enhance optical flow accuracy by incorporating motion information from both past and future frames. These methods estimate not only velocity but also acceleration, allowing for more precise modeling of nonlinear motion. 
However, discrepancies in velocity and acceleration between frames $I_{0}$ and $I_{1}$ can cause the interpolated flow vectors $V_{0 \rightarrow t}$ and $V_{1 \rightarrow t}$ to diverge, leading to misaligned target positions and blur artifacts. Furthermore, inaccurate flow estimation in occluded regions degrades interpolation quality, necessitating more robust motion modeling strategy. 

\noindent \textbf{OCAI (our baseline):} 
OCAI~\cite{jeong2024ocai} employs both forward and backward warping strategies to achieve high-quality intermediate frame generation using only a pre-trained optical flow model. Specifically, OCAI decomposes the optical flow ($V_{0 \rightarrow 1}$) and estimates the optical flow ($V_{t \rightarrow 1}$) from $I_t$ to $I_{1}$ through forward warping. 
\vspace{-4pt}
\begin{align}\label{decompose}
\begin{split}
    V_{0 \rightarrow 1}(x) &= V_{0 \rightarrow t}(x) + V_{t \rightarrow 1}(x + V_{0 \rightarrow t}(x)), \\
    V_{t \rightarrow 1}(x) &= \omega_{f}(V_{0 \rightarrow 1}(x) - V_{0 \rightarrow t}(x),\, V_{0 \rightarrow t}(x)).
\end{split}
\end{align}
OCAI addresses the conflict problem by introducing an occlusion-aware weighting mask and hole problem by leveraging optical flow consistency, which fills missing area using information from the complementary flow field. 

Furthermore, OCAI computes a confidence map based on forward-backward flow consistency~\cite{jeong2023distractflow}, enabling more reliable fusion and refinement of interpolated frames.
Finally, the intermediate frame ($I_{t}$) is synthesized through backward warping, guided by a confidence map~\cite{jeong2024ocai}.
\begin{equation} \label{combine}
\small
{I}_{t} = \frac{C_{t,\,0}}{C_{t,\,0} + C_{t,\,1}} w_b(I_{0},V_{t \rightarrow 0}) + \frac{C_{t,\,1}}{C_{t,\,0} + C_{t,\,1} } w_b(I_{1},V_{t \rightarrow 1}),
\vspace{-5pt}
\end{equation}

\subsection{Diffusion-Based Video Frame Interpolation}
Recently, diffusion-based approaches have gained considerable momentum in the video frame interpolation domain. Some methods~\cite{danier2024ldmvfi, jain2024video} condition the generative process on two input frames and synthesize intermediate frames by iteratively refining random Gaussian noise. In contrast, more advanced techniques, such as TRF~\cite{feng2024explorative} and GenIn~\cite{wang2024generative}, leverage off-the-shelf foundation models like Stable Video Diffusion~\cite{jain2024video} to generate entire video sequences from a pair of input frames. These methods independently process each input frame to generate corresponding latent representations and then merge the two latent trajectories at each diffusion step.
\begin{equation} \label{diffusion}
\begin{split}
\small
\bm{z}^{k-1,s} = \Phi (\bm{z}^{k}, c_s, k), \quad \bm{z}^{k-1,e} = \Phi (\bm{z}^{k}, c_e, k), \\
z_{n}^{k-1} = \alpha_{n} \cdot z_{n}^{k-1, s} + (1-\alpha_{n}) \cdot z_{N-n-1}^{k-1, e}.
\end{split}
\end{equation}
where \textit{z} denotes the latent representation of video frames, $\Phi$ represents the denoising network, $k$ corresponds to each denoising timestep, and $c_s$ and $c_e$ refer to the conditioning inputs derived from the start and end frames, respectively. GenIn further enhances temporal coherence by introducing a rotated temporal self-attention mechanism that aligns motion across latent features more effectively. Despite their ability to produce visually compelling and perceptually realistic frames, diffusion-based methods exhibit notable limitations. They often introduce unintended variations in brightness and contrast, and due to the lack of explicit pixel-level correspondence modeling, object motion may become inconsistent across frames. This can lead to temporal discontinuities and fragmented motion trajectories, undermining the overall coherence of the generated video.

\section{Proposed Approach}
\label{sec:method}

\begin{figure}[t]
\centering
\includegraphics[width=0.98\linewidth]{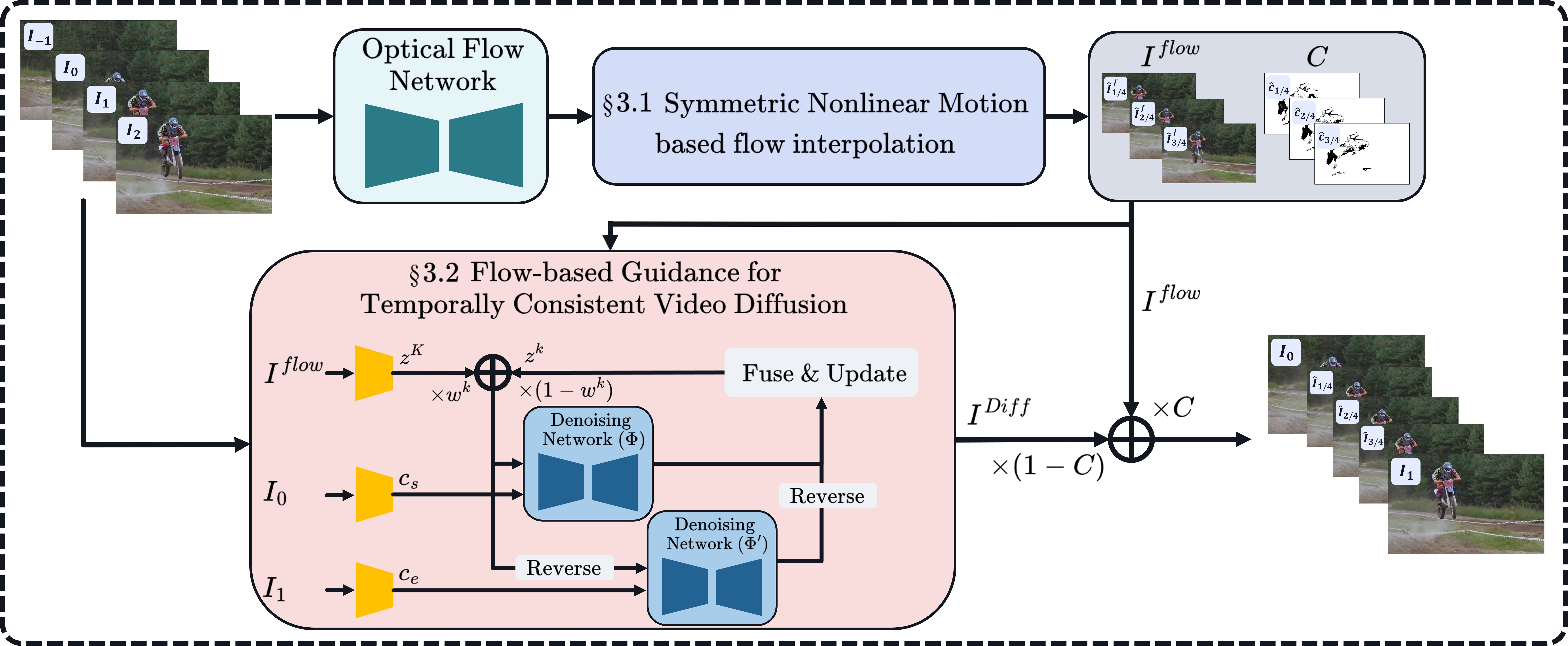}
\vspace{-5pt}
\caption{
Overview of SNM-VFI as a motion-guided generative visual processing framework. \ours first constructs correspondence-aware intermediate frames and confidence maps using symmetric nonlinear motion priors (Sec.~\ref{sub1}). These flow-guided frames are encoded as latent priors to initialize and iteratively guide a pre-trained video diffusion model, producing temporally coherent generative interpolations (Sec.~\ref{sub2}). Finally, confidence-aware fusion combines reliable flow-based structures with diffusion-generated details for enhanced interpolation quality (Sec.~\ref{sub3}).
}
\vspace{-3pt}
\label{fig:overview}
\end{figure}

Our proposed approach, \ours, significantly improves the quality and robustness of video frame interpolation by integrating the complementary strengths of optical flow and generative diffusion models. In Section~\ref{sub1}, we employ a multi-frame flow prediction strategy that explicitly models complex, nonlinear motion. This enables the generation of high-fidelity intermediate frames through advanced flow-based interpolation. In Section~\ref{sub2}, flow-guided intermediate frames serve as conditional inputs to the diffusion model, promoting temporal coherence and perceptual realism. Section~\ref{sub3} describes the use of confidence maps to fuse flow- and diffusion-based outputs, selectively integrating reliable regions from each to ensure structural precision and visual consistency.

\subsection{Symmetric Nonlinear Motion Based Flow Interpolation}
\label{sub1}
Following the flow decomposition principle utilized in OCAI (Eq.~$\ref{decompose}$), we first decompose the optical flow $V_{0 \rightarrow 1}$ into two directional components: $V_{0 \rightarrow t}$ and $V_{t \rightarrow 1}$. To capture nonlinear motion, we approximate $V_{0 \rightarrow t}$ not only using the base flow $V_{0 \rightarrow 1}$, but also by incorporating motion cues from symmetric neighborhood flows, specifically $V_{0 \rightarrow -1}$ and $V_{1 \rightarrow 2}$. This enriched motion context enables our model to better identify and represent complex, nonlinear motion patterns:
\begin{equation} \label{eq:our0tapp}
\footnotesize
\begin{split}
V_{0 \rightarrow t}(x) &= t \cdot V_{0 \rightarrow 1}(x)
                        + \alpha \cdot t \cdot (1-t) \cdot \frac{-V_{0 \rightarrow -1} + w_b(-V_{1 \rightarrow 2},V_{0 \rightarrow 1})}{2}, 
\end{split}
\end{equation}
where $\alpha$ is a nonlinear hyperparameter that controls the magnitude of the nonlinearity and $w_b$ is backward warping. The parameter $\alpha$ is set to 0.5. To address occluded regions, we compute a comprehensive occlusion map~\cite{meister2018unflow} and replace the flow within these areas using a simplified and more stable linear motion approximation, thereby enhancing reliability in regions with uncertain motion: 
\begin{equation} \label{our_occ}
\begin{split}
M_{0 \rightarrow t} = M_{0 \rightarrow 1} \cdot M_{0 \rightarrow -1} \cdot w_{b}(M_{1 \rightarrow 2}, V_{0 \rightarrow 1}),\\
V_{0 \rightarrow t}(p) = V^{OCAI}_{0 \rightarrow t}(p) \quad \text{if} \ M_{0 \rightarrow t}(p)=1,
\end{split}
\end{equation}
where $M$ denotes the matching map, defined as $M=1-O$, $O$ represents the occlusion map. Given our $V_{0 \rightarrow t}$ and Eq.~\ref{decompose}, we derive the optical flow  $V^{I_{0}}_{t \rightarrow 1}(x)$. Here, the superscript $I_{0}$ indicates that this flow originates from the input frame $I_{0}$. Note that we apply the same occlusion-aware weighting mask during forward warping to address the conflict problem.

Unlike OCAI, our method does not rely on optical flow consistency for hole filling, as we do not assume linear motion. Instead, we perform forward warping on the inverse flows $-V_{0 \rightarrow t}(x)$ and $-V_{1 \rightarrow t}(x)$ using the respective flows $V_{0 \rightarrow t}(x)$ and $V_{1 \rightarrow t}(x)$. This yields additional flows $V^{I_{0}}_{t \rightarrow 0}(x)$ and $V^{I_{1}}_{t \rightarrow 1}(x)$, which are then used to fill the hole regions as follows: 
\begin{equation} \label{holefill}
\begin{split}
\small
V^{I_1}_{t \rightarrow 1} &= w_{f}(-V_{1 \rightarrow t},V_{1 \rightarrow t}), \\
V_{t \rightarrow 1} &= (1-H)\cdot V^{I_0}_{t \rightarrow 1} + H \cdot V^{I_1}_{t \rightarrow 1},
\end{split}
\end{equation}
where H is hole mask used in OCAI~\cite{jeong2024ocai} and $w_f$ is forward warping. After we obtain final optical flows ($V_{t \rightarrow 1}$ and $V_{t \rightarrow 0}$), we compute the confidence maps and generate inter-frame $I_t$ using Eq.~\ref{combine}.

\subsection{Flow-Based Guidance for Temporally Consistent Video Diffusion}
\label{sub2}

Given the two real frames ($I_{0}$ and $I_{1}$), a set of intermediate frames ($\hat{I}_{t_1}, \hat{I}_{t_2}, ..., \hat{I}_{t_k}$) generated based on \ref{sub1}, we extract latent representations from the flow-based intermediate frames using a VAE encoder: ${z}_{n}^{K} \leftarrow {VAE}_{e}(\hat{I}_{n})$. We use this latent representation as the initial input for the diffusion process, replacing random Gaussian noise, and proceed with the subsequent denoising steps.
At each denoising step, we incorporate the flow-guided latent representation at the feature level as follows: \vspace{-6pt}
\begin{equation} \label{fusion}
z^{k-1} \leftarrow w^{k-1} \cdot z^{K} + (1-w^{k-1}) \cdot z^{k-1}_{diff}, 
\vspace{-4pt}
\end{equation}
where $z^{K}$ denotes the latent representation extracted from the flow-based intermediate frames, while $z^{k-1}_{diff}$ represents the output of the diffusion step. The weighting function $w^{k}$ is defined as $k/K$, where $K$ denotes the total number of diffusion steps and $k$ decreases from $K$ to $0$ during sampling. Thus, the flow-guided latent $z^K$ has a stronger influence in early denoising stages, while the diffusion output $z^{k-1}_{diff}$ gradually dominates in later steps to refine perceptual details.
Through feature-level fusion, we inject flow information into the intermediate stages of the diffusion process. After completing the denoising steps, the final latent output is decoded using a VAE decoder ($VAE_{d}$) as follows:
\begin{equation} \label{vae_decoder}
\vspace{-5pt}
\begin{split}
I^{\text{Diff}}\leftarrow VAE_{d}(z^{0})
\end{split}
\vspace{-4pt}
\end{equation}

\subsection{Confidence-Aware Fusion of Flow and Diffusion Outputs}
\label{sub3}
We fuse the two intermediate frames, one from the flow-based output and the other from the diffusion-based output, using a confidence map computed from Section~\ref{sub1}, as follows:
\begin{equation} \label{final_fusion}
I_{t} = C_{t} \cdot I^{\text{Flow}}_{t} + (1-C_{t})\cdot I^{\text{Diff}}_{t}.
\end{equation}

\section{Experiments}
\label{sec:exp}


\subsection{Experimental Setup}
\textbf{Datasets:} 
We evaluate \ours against state-of-the-art video frame interpolation (VFI) methods using three widely adopted benchmarks: DAVIS~\cite{pont20172017}, Sintel~\cite{butler2012naturalistic}, and KITTI~\cite{geiger2013vision, menze2015object}. Experiments are conducted under two interpolation scenarios:
1) $\times 2$ Interpolation:
The task is to reconstruct the middle frame $I_{1}$ using the model. Most existing algorithms rely only on the two frames ($I_{0}, I_{2}$), where some methods~\cite{xu2019quadratic, liu2020enhanced}, including ours, utilize four frames ($I_{-2}, I_{0}, I_{2}, I_{4}$) to generate the middle frame. 
2) $\times 4$ Interpolation: 
The task is to generate the intermediate frames $I_{1}, I_{2}, I_{3}$ between given frames $I_{0}$ and $I_{4}$ using the model. Similar to the $\times$2 Interpolation setting, most existing algorithms utilize only the two input frames ($I_{0}, I_{4}$), whereas others, including ours, use four frames ($I_{-4}, I_{0}, I_{4}, I_{8}$) to synthesize the intermediate frames. More details can be found in the supplementary material. 

\textbf{Models:} 
To ensure a fair comparison with existing VFI approaches, we adopt the official implementations and publicly released model weights, primarily those pre-trained on the Vimeo90k~\cite{xue2019video} dataset. For forward-warping based methods such as RIPR~\cite{han2022realflow} and OCAI~\cite{jeong2024ocai}, we use the RAFT~\cite{teed2020raft} optical flow model with weights trained on the FlyingChairs~\cite{dosovitskiy2015flownet} and FlyingThings3D~\cite{mayer2016large} datasets.
In \ours, we employ the same RAFT model used in baselines like RIPR and OCAI. In addition, we integrate the GenIn~\cite{wang2024generative} model, using its publicly released fine-tuned weights\footnote{The GenIn model generates a total of 25 intermediate frames ($\hat{I}_0$, $\hat{I}_{1}$,... $\hat{I}_{24}$). For evaluation, the $\hat{I}_{12}$ frame is compared with the true middle frame, the $\hat{I}_{6}$ frame with the $I_{1/4}$ frame, and the $\hat{I}_{18}$ frame with the $I_{3/4}$ frame.}. Note that all components in our pipeline are used as-is, without any additional training or fine-tuning. Additional details are provided in the supplementary material.

\begin{table*}[t!]
\begin{center}
\caption{Video Frame Interpolation (VFI) results for the $\times$2 interpolation setting on the DAVIS, Sintel, and KITTI datasets. We report standard image quality metrics: PSNR ($\uparrow$), SSIM ($\uparrow$), LPIPS ($\downarrow$), and FID score ($\downarrow$)), where arrows indicate whether higher or lower values are better. \textcolor{red}{Red}/\textcolor{blue}{Blue}: \textcolor{red}{Best} and \textcolor{blue}{Second Best} results.}
\label{tab:middle2}
\vspace{-15pt}
\adjustbox{max width=1.0\textwidth}
{
\begin{tabular}{|l||c|c|c|}
\hline
 \multirow{2}*{ \cellcolor{mycolor2} Methods} 
\cellcolor{mycolor2} & \cellcolor{mycolor2} DAVIS--2 & \cellcolor{mycolor2} Sintel--2 & \cellcolor{mycolor2} KITTI--2 \\
\cline{2-4}
\multirow{-2}*{ \cellcolor{mycolor2} Methods} & \cellcolor{mycolor2} \hspace{1pt}PSNR\hspace{1pt}/\hspace{1pt}SSIM\hspace{1pt}/\hspace{1pt}LPIPS\hspace{1pt}/ \ FID \ \ \ & \cellcolor{mycolor2} \hspace{1pt}PSNR\hspace{1pt}/\hspace{1pt}SSIM\hspace{1pt}/\hspace{1pt}LPIPS\hspace{1pt}/ \ FID \ \ \ & \cellcolor{mycolor2} \hspace{1pt}PSNR\hspace{1pt}/\hspace{1pt}SSIM\hspace{1pt}/\hspace{1pt}LPIPS\hspace{1pt}/ \ FID \ \ \ \\
\hline
IFRNet~\cite{kong2022ifrnet} \scriptsize{(CVPR 2022)} & 27.3168/0.8749/0.1347/\ 46.2731&29.0813/0.8898/0.1336/110.0396&21.5024/0.7620/0.2168/44.4112 \\
VFIFormer~\cite{lu2022video} \scriptsize{(CVPR 2022)} & \textcolor{blue}{27.5316}/\textcolor{blue}{0.8825}/0.1415/\ 47.9819 & \textcolor{blue}{29.6084}/\textcolor{blue}{0.8990}/\textcolor{blue}{0.1185}/110.1337&\textcolor{blue}{22.4012}/\textcolor{red}{0.7884}/0.2268/40.4422 \\
AMT~\cite{li2023amt} \scriptsize{(CVPR 2023)} & 27.5251/0.8783/0.1331/\ 46.5251&29.3449/0.8931/0.1260/102.5036&21.8201/0.7715/0.2220/42.7778 \\
EMA-VFI~\cite{zhang2023extracting} \scriptsize{(CVPR 2023)} & \textcolor{red}{27.6178}/0.8781/0.1456/\ 47.3722&\textcolor{red}{29.7058}/0.8960/0.1373/111.4100&21.8389/0.7681/0.2536/49.7861\\
BiM-VFI~\cite{seo2025bim} \scriptsize{(CVPR 2025)} &27.1588/0.8765/0.1471/\ \textcolor{red}{35.9279}&28.9174/0.8867/0.1355/\ \textcolor{blue}{70.8399}&21.8127/0.7646/0.2343/27.5182 \\
RIPR~\cite{han2022realflow} \scriptsize{(ECCV 2022)} &25.8446/0.8638/0.1364/\ 47.3994&27.7324/0.8813/0.1275/\ 88.0443&21.1753/0.7381/0.1945/35.1997\\
OCAI~\cite{jeong2024ocai} \scriptsize{(CVPR 2024)} &26.8707/0.8749/\textcolor{blue}{0.1327}/\ 43.7556&28.7139/0.8912/0.1237/\ 74.6818&22.0865/0.7604/\textcolor{blue}{0.1886}/\textcolor{blue}{27.0655}\\
LDMVFI~\cite{danier2024ldmvfi} \scriptsize{(AAAI 2024)} &25.9154/0.8461/0.1453/\ 51.0077&26.9267/0.8524/0.1529/104.6009&19.5199/0.6764/0.2475/47.0781\\
TRF~\cite{feng2024explorative} \scriptsize{(ECCV 2024)} & 15.0228/0.5397/0.4740/113.9287&16.1904/0.5782/0.4557/195.9670&13.0914/0.4464/0.4684/58.0689\\
GenIn~\cite{wang2024generative} \scriptsize{(ICLR 2025)} &19.6487/0.6936/0.2708/\ 68.7283&19.4844/0.6565/0.3330/134.3639&17.0116/0.5842/0.3255/38.7543\\
\rowcolor{myrowcolor} \textbf{\ours} &27.4797/\textcolor{red}{0.8928}/\textcolor{red}{0.1243}/\ \textcolor{blue}{43.0419} &29.3269/\textcolor{red}{0.9109}/\textcolor{red}{0.1134}/\ \textcolor{red}{70.7634} &\textcolor{red}{22.8125}/\textcolor{blue}{0.7868}/\textcolor{red}{0.1811}/\textcolor{red}{26.4216} \\
 \hline
\end{tabular}
}
\vspace{-15pt}
\end{center}
\end{table*}

\textbf{Evaluation Metrics:} 
We evaluate interpolation quality using three standard metrics: Peak Signal-to-Noise Ratio (PSNR), Structural Similarity Index Measure (SSIM)~\cite{wang2004image}, and Learned Perceptual Image Patch Similarity (LPIPS with VGG)~\cite{zhang2018unreasonable}. To assess perceptual realism, we also report the Fréchet Inception Distance (FID)~\cite{heusel2017gans} on the synthesized frames.

\subsection{$\times$2 Interpolation on DAVIS, Sintel, and KITTI}
Table~\ref{tab:middle2} presents the results of our experiments under the $\times$2 interpolation setting across three benchmark datasets: DAVIS, Sintel, and KITTI. On the DAVIS and Sintel datasets, Flow-based methods, particularly those employing backward warping (IFRNet, VFIFormer, AMT, EMA-VFI, Bim-VFI), achieve high scores in PSNR and SSIM, indicating strong pixel-level fidelity. BiM-VFI is the only method that achieves strong FID scores but performs significantly worse in PSNR, SSIM, and LPIPS compared to other backward warping methods.
Forward warping methods (RIPR, OCAI) generally perform better in LPIPS and FID than backward warping approaches, excluding BiM-VFI. \ours achieves the best SSIM and LPIPS scores and remains highly competitive in PSNR and FID, with only a marginal gap compared to the top-performing model. Notably, our approach demonstrates a substantial improvement in LPIPS, outperforming all other methods by a significant margin.

On the KITTI dataset, backward warping methods continue to excel in PSNR and SSIM, while forward warping techniques perform well in LPIPS and FID. \ours achieves the best results in all metrics except SSIM, where it ranks second with a score very close to the top-performing approach. These results collectively demonstrate the robustness and effectiveness of our framework across diverse motion scenarios and evaluation criteria.

\begin{table*}[t!]
\begin{center}

\caption{Video Frame Interpolation (VFI) results on the DAVIS, Sintel, and KITTI datasets. The top section presents evaluations of the middle frame under the $\times$4 interpolation setting, while the bottom section shows results for all intermediate frames in the $\times$4 setting. We report standard image quality metrics: PSNR ($\uparrow$), SSIM ($\uparrow$), LPIPS ($\downarrow$), and FID score ($\downarrow$)), where arrows indicate whether higher or lower values are better. \textcolor{red}{Red}/\textcolor{blue}{Blue}: \textcolor{red}{Best} and \textcolor{blue}{Second Best} results.
}
\label{tab:middle}
\vspace{-15pt}
\adjustbox{max width=1.0\textwidth}
{
\begin{tabular}{|l||c|c|c|}
\hline
 \multirow{2}*{ \cellcolor{mycolor2} Methods} 
\cellcolor{mycolor2} & \cellcolor{mycolor2} DAVIS-4 & \cellcolor{mycolor2} Sintel-4 & \cellcolor{mycolor2} KITTI-4 \\
\cline{2-4}
\multirow{-2}*{ \cellcolor{mycolor2} Methods} & \cellcolor{mycolor2} \hspace{1pt}PSNR\hspace{1pt}/\hspace{1pt}SSIM\hspace{1pt}/\hspace{1pt}LPIPS\hspace{1pt}/ \ FID \ \ \ & \cellcolor{mycolor2} \hspace{1pt}PSNR\hspace{1pt}/\hspace{1pt}SSIM\hspace{1pt}/\hspace{1pt}LPIPS\hspace{1pt}/ \ FID \ \ \ & \cellcolor{mycolor2} \hspace{1pt}PSNR\hspace{1pt}/\hspace{1pt}SSIM\hspace{1pt}/\hspace{1pt}LPIPS\hspace{1pt}/ \ FID \ \ \ \\
\hline
\rowcolor{mycolor}\multicolumn{4}{|c|}{Middle Frame ($I_{0.5}$)}\\
\hline
IFRNet~\cite{kong2022ifrnet} \scriptsize{(CVPR 2022)} &22.9465/0.7795/0.2318/\ 81.4500&25.3899/0.8298/0.2045/164.6008&17.9034/0.6528/0.3088/66.1144\\
VFIFormer~\cite{lu2022video} \scriptsize{(CVPR 2022)} & \textcolor{red}{23.4146}/\textcolor{blue}{0.7932}/0.2238/\ 80.9076&\textcolor{red}{25.8165}/\textcolor{red}{0.8427}/0.1979/153.9648&\textcolor{blue}{18.5943}/\textcolor{blue}{0.6820}/0.3195/64.7331\\
AMT~\cite{li2023amt} \scriptsize{(CVPR 2023)} &23.2578/0.7860/0.2198/\ 78.8323&25.5592/0.8292/0.2015/161.1479&18.0569/0.6628/0.3143/64.1559\\
EMA-VFI~\cite{zhang2023extracting} \scriptsize{(CVPR 2023)} &23.2670/0.7843/0.2515/\ 84.2363&\textcolor{blue}{25.7122}/0.8345/0.2121/159.2066&18.2554/0.6623/0.3423/70.8467\\
BiM-VFI~\cite{seo2025bim} \scriptsize{(CVPR 2025)} &23.1200/0.7867/0.2232/\ \textcolor{red}{58.6681}&25.3769/0.8261/0.2048/\textcolor{red}{111.7590}&18.1651/0.6588/0.3185/45.8235\\
RIPR~\cite{han2022realflow} \scriptsize{(ECCV 2022)} &22.2779/0.7759/0.2040/\ 72.3845&24.2306/0.8106/0.1930/\textcolor{blue}{111.9524}&17.7725/0.6315/\textcolor{blue}{0.2761}/51.0808\\
OCAI~\cite{jeong2024ocai} \scriptsize{(CVPR 2024)} &23.0076/0.7855/\textcolor{blue}{0.2017}/\ 69.9548&25.3340/0.8292/\textcolor{blue}{0.1872}/113.2435&18.3811/0.6489/0.2740/46.0131\\
LDMVFI~\cite{danier2024ldmvfi} \scriptsize{(AAAI 2024)} &21.9603/0.7446/0.2348/\ 84.0895&23.6099/0.7892/0.2231/148.7152&16.8230/0.5818/0.3205/61.2554\\
TRF~\cite{feng2024explorative} \scriptsize{(ECCV 2024)} &14.9700/0.5386/0.4801/116.1452&15.8220/0.5725/0.4771/215.3137&12.8847/0.4382/0.4675/61.3676\\
GenIn~\cite{wang2024generative} \scriptsize{(ICLR 2025)} &18.2682/0.6511/0.3159/\ 79.9909&17.7754/0.6180/0.3729/149.5369&16.5727/0.5671/0.3341/\textcolor{red}{41.9435}\\
\hline
\rowcolor{myrowcolor} \textbf{\ours} &\textcolor{blue}{23.2814}/\textcolor{red}{0.7977}/\textcolor{red}{0.1961}/\ \textcolor{blue}{69.3161}&25.6858/\textcolor{blue}{0.8405}/\textcolor{red}{0.1811}/117.9414&\textcolor{red}{19.1511}/\textcolor{red}{0.6867}/\textcolor{red}{0.2632}/\textcolor{blue}{41.9844}\\
 \hline
 \hline
\rowcolor{mycolor}\multicolumn{4}{|c|}{All Frames ($I_{1/4}$, $I_{2/4}$, $I_{3/4}$)}\\
\hline
IFRNet~\cite{kong2022ifrnet} \scriptsize{(CVPR 2022)} &23.8182/0.8019/0.2138/\ 57.9793&25.9228/0.8456/0.1941/133.2429&18.4975/0.6737/0.2952/44.8743\\
VFIFormer~\cite{lu2022video} \scriptsize{(CVPR 2022)} &\textcolor{red}{24.2433}/\textcolor{blue}{0.8139}/0.2058/\ 59.8759&\textcolor{red}{26.2521}/\textcolor{red}{0.8546}/0.1871/124.3194&\textcolor{blue}{19.1139}/\textcolor{blue}{0.6997}/0.3073/45.1584\\
AMT~\cite{li2023amt} \scriptsize{(CVPR 2023)} &24.1223/0.8079/0.2006/\ 53.2007&26.0644/0.8451/0.1911/123.4832&18.6756/0.6823/0.2988/42.1071\\
EMA-VFI~\cite{zhang2023extracting} \scriptsize{(CVPR 2023)} &\textcolor{blue}{24.1316}/0.8069/0.2346/\ 61.4262&26.2659/0.8497/0.2047/132.0049&18.6655/0.6798/0.3207/43.0282\\
BiM-VFI~\cite{seo2025bim} \scriptsize{(CVPR 2025)} &23.9099/0.8069/0.2051/\ \textcolor{red}{40.1814}&25.8389/0.8372/0.1907/\ \textcolor{red}{86.5464}&18.7143/0.6757/0.3037/28.7029\\
RIPR~\cite{han2022realflow} \scriptsize{(ECCV 2022)} &22.9510/0.7959/0.1907/\ 51.6677&24.7052/0.8217/0.1816/\ 93.0850&18.2286/0.6475/0.2638/33.1112\\
OCAI~\cite{jeong2024ocai} \scriptsize{(CVPR 2024)} &23.7432/0.8063/\textcolor{blue}{0.1864}/\ 51.2046&25.8927/0.8402/\textcolor{blue}{0.1747}/\ 93.8831&18.8717/0.6650/\textcolor{blue}{0.2620}/29.3621\\
LDMVFI~\cite{danier2024ldmvfi} \scriptsize{(AAAI 2024)} &22.7786/0.7702/0.2126/\ 54.2721&24.2163/0.8057/0.2051/108.8361&17.3220/0.6017/0.2981/31.6903\\
TRF~\cite{feng2024explorative} \scriptsize{(ECCV 2024)} &15.3228/0.5500/0.4694/\ 85.5805&16.0730/0.5845/0.4614/173.3449&13.3383/0.4556/0.4536/37.9767\\
GenIn~\cite{wang2024generative} \scriptsize{(ICLR 2025)} &18.7968/0.6696/0.2967/\ 54.9852&18.5285/0.6414/0.3443/121.9460&17.1569/0.5855/0.3186/\textcolor{red}{25.6861}\\
\hline
\rowcolor{myrowcolor} \textbf{\ours} &24.0859/\textcolor{red}{0.8188}/\textcolor{red}{0.1814}/\ \textcolor{blue}{48.5302}&\textcolor{blue}{26.2414}/\textcolor{blue}{0.8510}/\textcolor{red}{0.1688}/\ \textcolor{blue}{92.4279}&\textcolor{red}{19.5978}/\textcolor{red}{0.7006}/\textcolor{red} {0.2519}/\textcolor{blue}{27.6261}\\
\hline
\end{tabular}
}
\vspace{-25pt}
\end{center}
\end{table*}


\subsection{$\times$4 Interpolation on DAVIS, Sintel, and KITTI}

\textbf{Middle-Frame ($I_{1/2}$) Evaluation:} 
Table~\ref{tab:middle} reports results under the $\times$4 interpolation setting using the same datasets. When evaluating the middle frame, we observe trends similar to those in the ×2 setting. Backward warping-based methods generally achieve high PSNR and SSIM scores (excluding BiM-VFI), while forward warping-based methods perform well in LPIPS and FID. Interestingly, GenIn records the best FID score on the KITTI dataset. This highlights the trade-offs between perceptual realism and pixel-level accuracy in diffusion-based models. Our proposed method consistently shows either the best or near-best performance across all metrics. It demonstrates strong generalization across datasets and interpolation settings, achieving competitive PSNR and SSIM scores while outperforming other methods in perceptual quality metrics such as LPIPS and FID.

\textbf{All-Frame ($I_{1/4}$, $I_{2/4}$, $I_{3/4}$) Evaluations: }
In the $\times$4 interpolation setting, we evaluate all generated intermediate frames ($I_{1/4}, I_{2/4}, I_{3/4}$) to assess the temporal generalization capabilities of each method. Traditional backward warping-based approaches are typically trained for fixed temporal positions (e.g., the midpoint) and struggle with arbitrary time steps. In our experiment, we adopted a hierarchical interpolation strategy: first, we generated the middle frame ($I_{2/4}$) using the input frames ($I_{0}, I_{1}$); then we synthesized ($I_{1/4}$) using ($I_{0}, I_{2/4}$), and ($I_{3/4}$) using ($I_{2/4}, I_{1}$). Across all experiments, we observed that the performance of all frames was consistently better than that of the middle frame ($I_{2/4}$). We hypothesize that this improvement stems from the relatively lower degree of nonlinearity in the motion between adjacent frames ($I_{0} \rightarrow I_{1/4}, I_{1} \rightarrow I_{3/4}$) compared to the central interpolation ($I_{0} \rightarrow I_{2/4}, I_{1} \rightarrow I_{2/4}$). The more linear nature of these shorter temporal intervals likely contributes to the enhanced interpolation quality observed in the all-frame experiment.

\begin{figure}[p]
\begin{center}$
\centering
\begin{tabular}{ c cc cc c}

& \multicolumn{2}{c}{\text{DAVIS}} & \multicolumn{2}{c}{\text{Sintel}} & 
\text{KITTI} \\

\hspace{-0.1cm} \rotatebox{90}{\hspace{3.5mm} \scriptsize{$I_{0}$}}
 & \hspace{-0.1cm} \includegraphics[width=2.33cm,height=1.3cm]{Fig/rebuttal/00011.jpg} & \hspace{-0.15cm} 
\includegraphics[width=2.33cm,height=1.3cm]{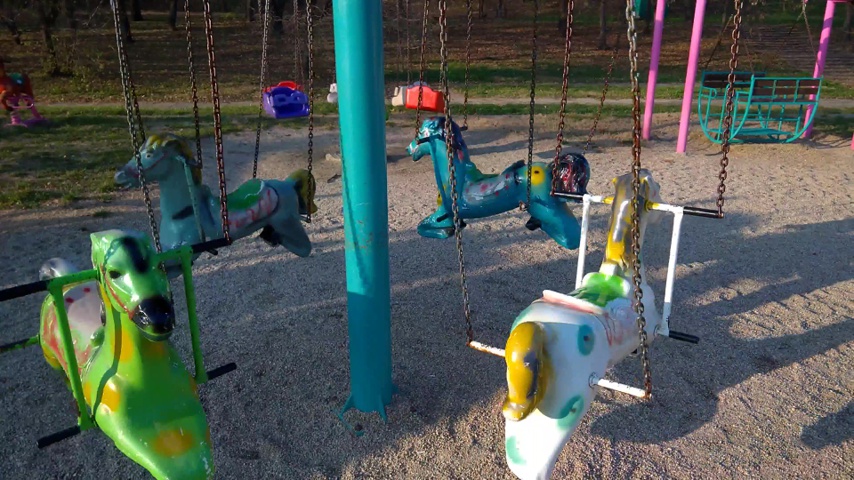}& \hspace{-0.15cm} 
\includegraphics[width=2.33cm,height=1.3cm]{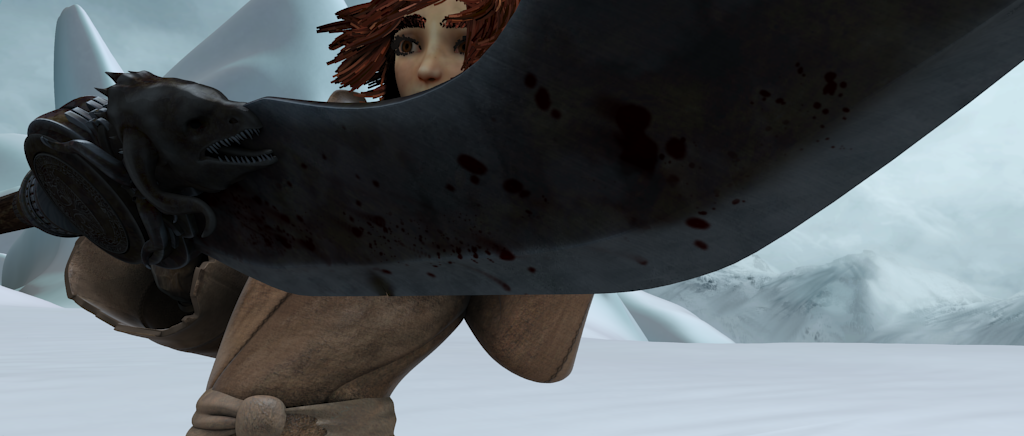}& \hspace{-0.15cm} 
\includegraphics[width=2.33cm,height=1.3cm]{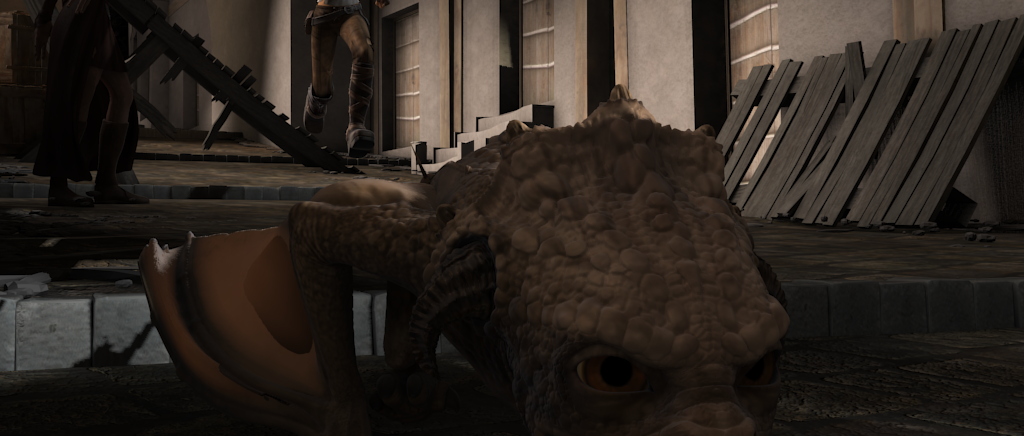}& \hspace{-0.15cm} 
\includegraphics[width=2.33cm,height=1.3cm]{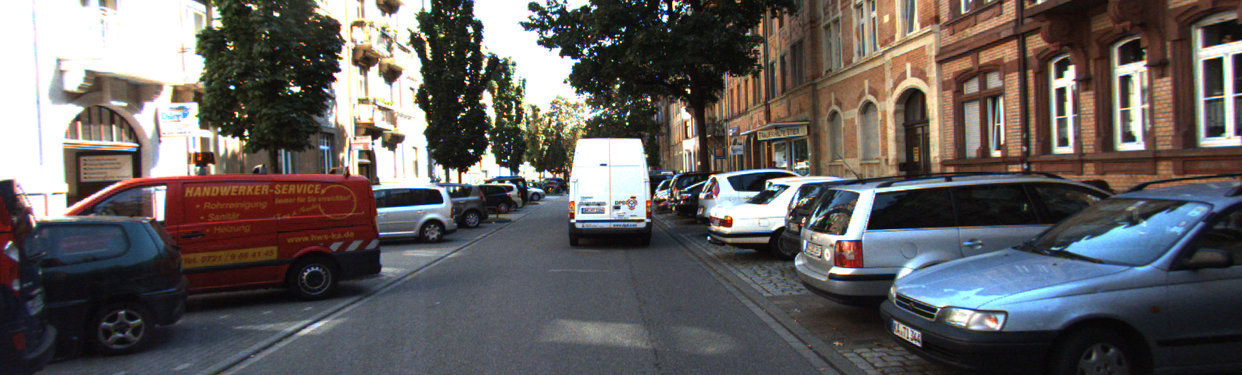}  \\
\hspace{-0.1cm} \rotatebox{90}{\hspace{3.5mm} \scriptsize{$I_{1}$}}
 & \hspace{-0.1cm} \includegraphics[width=2.33cm,height=1.3cm]{Fig/rebuttal/00013.jpg} & \hspace{-0.15cm} 
\includegraphics[width=2.33cm,height=1.3cm]{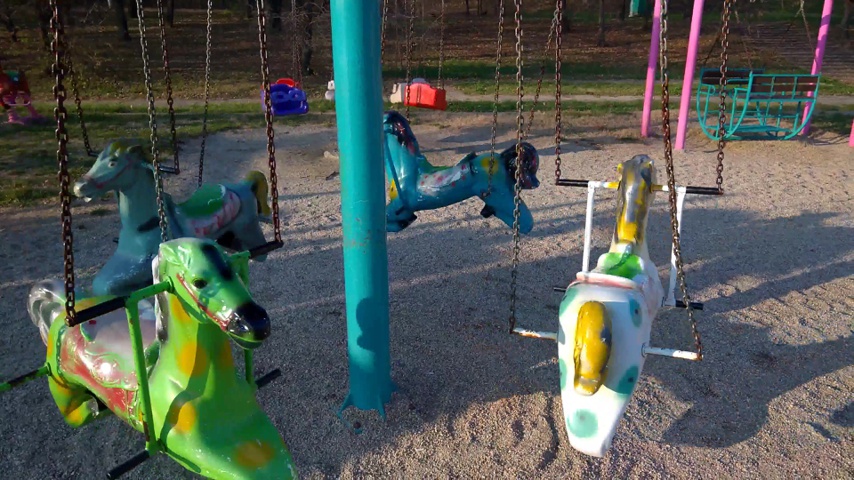}& \hspace{-0.15cm} 
\includegraphics[width=2.33cm,height=1.3cm]{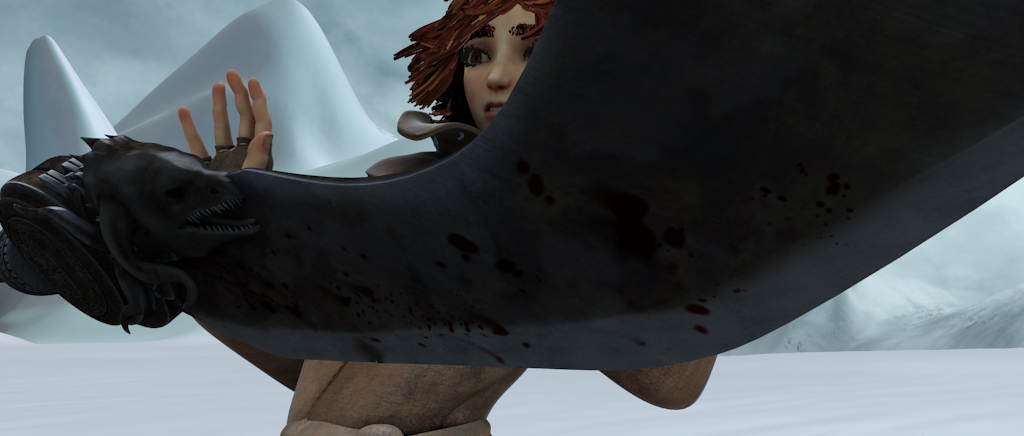}& \hspace{-0.15cm} 
\includegraphics[width=2.33cm,height=1.3cm]{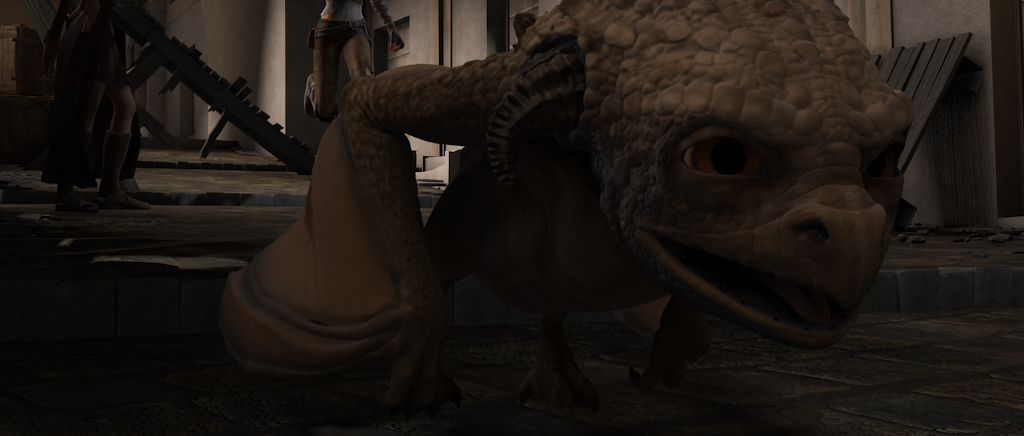}& \hspace{-0.15cm} 
\includegraphics[width=2.33cm,height=1.3cm]{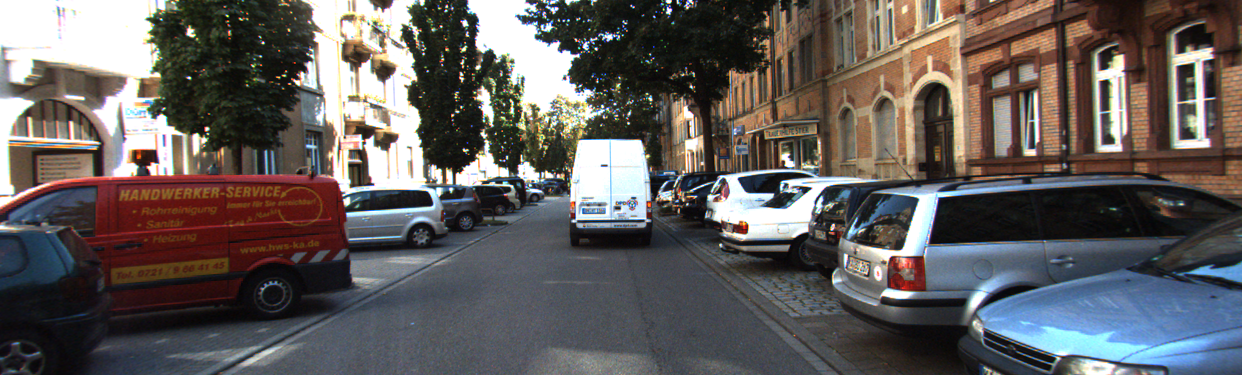}  \\
\hspace{-0.1cm} \rotatebox{90}{\hspace{2.5mm} \scriptsize{GT}}
 & \hspace{-0.1cm} \includegraphics[width=2.33cm,height=1.3cm]{Fig/results/davis_gt.jpg} & \hspace{-0.15cm} 
\includegraphics[width=2.33cm,height=1.3cm]{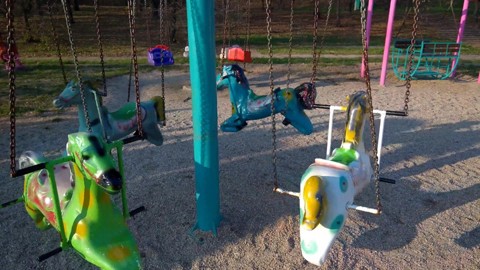}& \hspace{-0.15cm} 
\includegraphics[width=2.33cm,height=1.3cm]{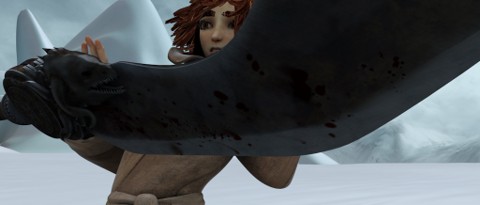}& \hspace{-0.15cm} 
\includegraphics[width=2.33cm,height=1.3cm]{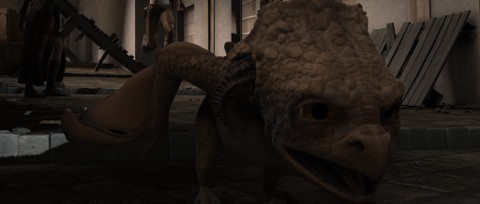}& \hspace{-0.15cm} 
\includegraphics[width=2.33cm,height=1.3cm]{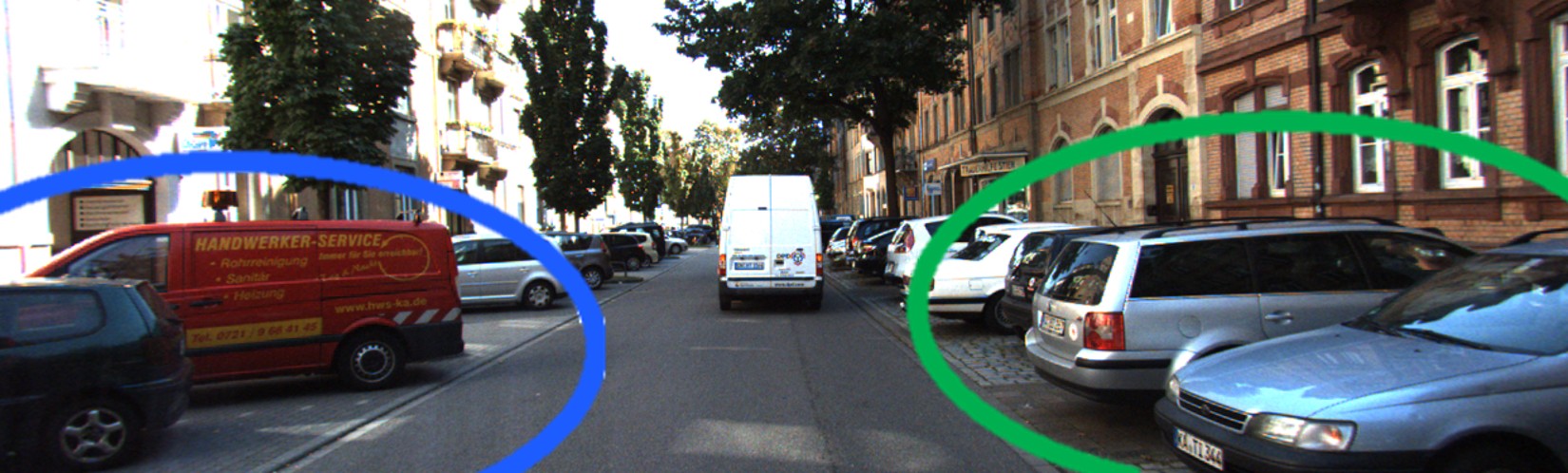}  \\
\hspace{-0.1cm} \rotatebox{90}{\hspace{-0.1mm} \tiny{VFIFormer}}
 & \hspace{-0.1cm} \includegraphics[width=2.33cm,height=1.3cm]{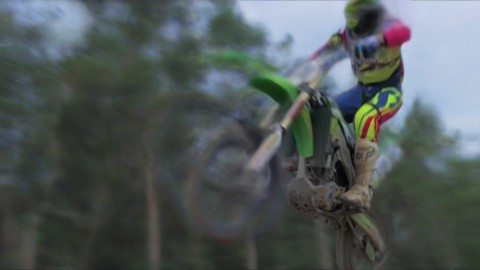} & \hspace{-0.15cm} 
\includegraphics[width=2.33cm,height=1.3cm]{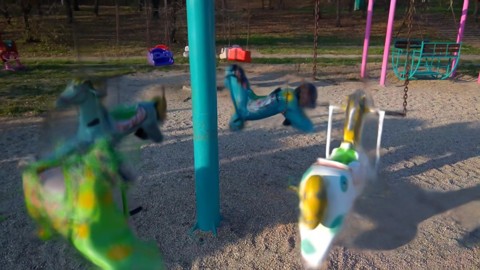}& \hspace{-0.15cm} 
\includegraphics[width=2.33cm,height=1.3cm]{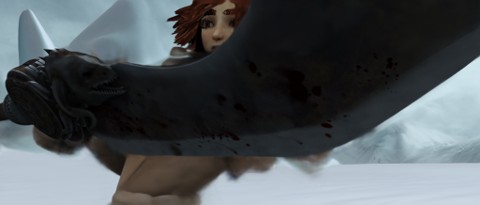}& \hspace{-0.15cm} 
\includegraphics[width=2.33cm,height=1.3cm]{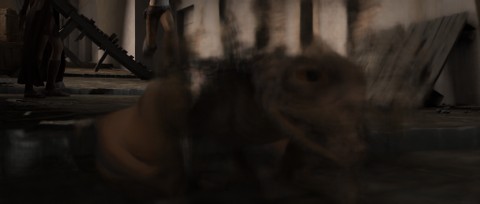}& \hspace{-0.15cm} 
\includegraphics[width=2.33cm,height=1.3cm]{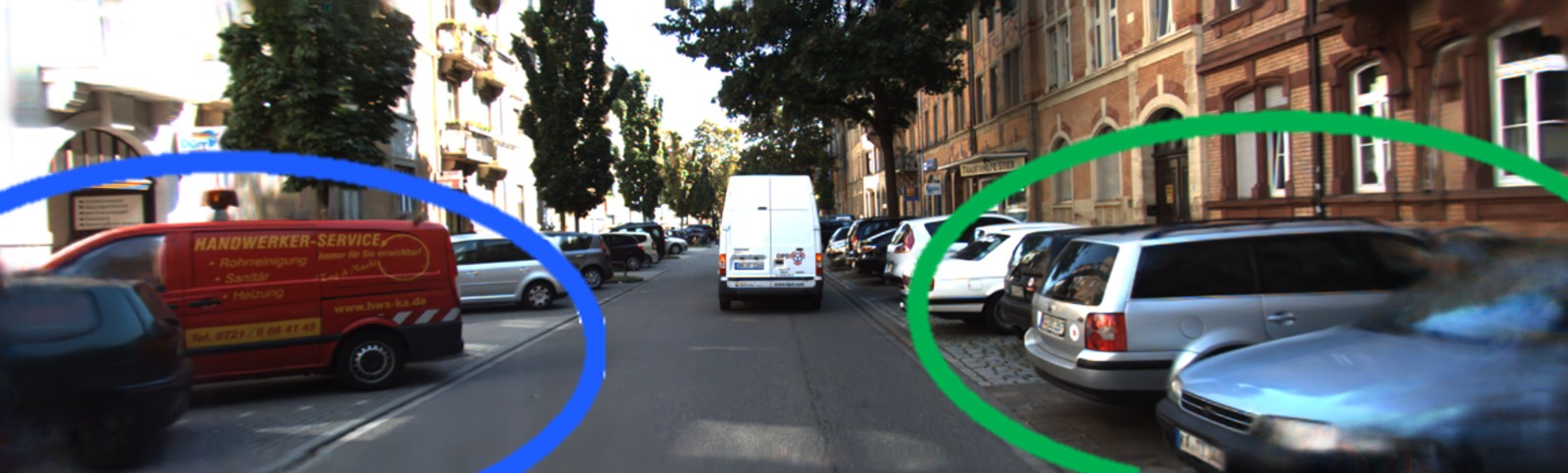}  \\
\hspace{-0.1cm} \rotatebox{90}{\hspace{0.5mm} \tiny{EMA-VFI}}
 & \hspace{-0.1cm} \includegraphics[width=2.33cm,height=1.3cm]{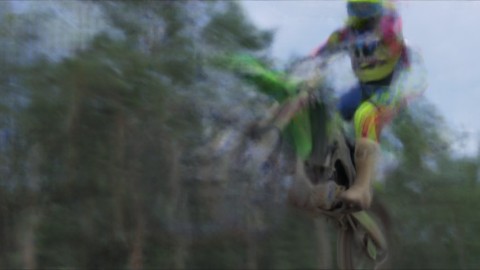} & \hspace{-0.15cm} 
\includegraphics[width=2.33cm,height=1.3cm]{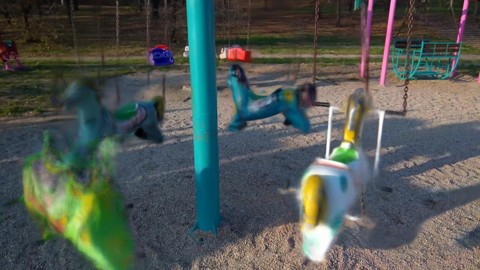}& \hspace{-0.15cm} 
\includegraphics[width=2.33cm,height=1.3cm]{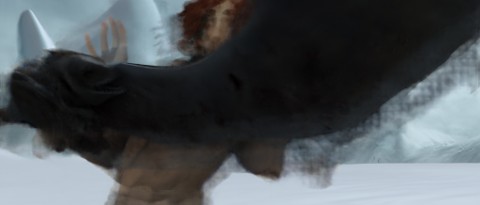}& \hspace{-0.15cm} 
\includegraphics[width=2.33cm,height=1.3cm]{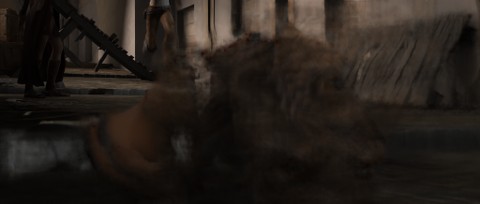}& \hspace{-0.15cm} 
\includegraphics[width=2.33cm,height=1.3cm]{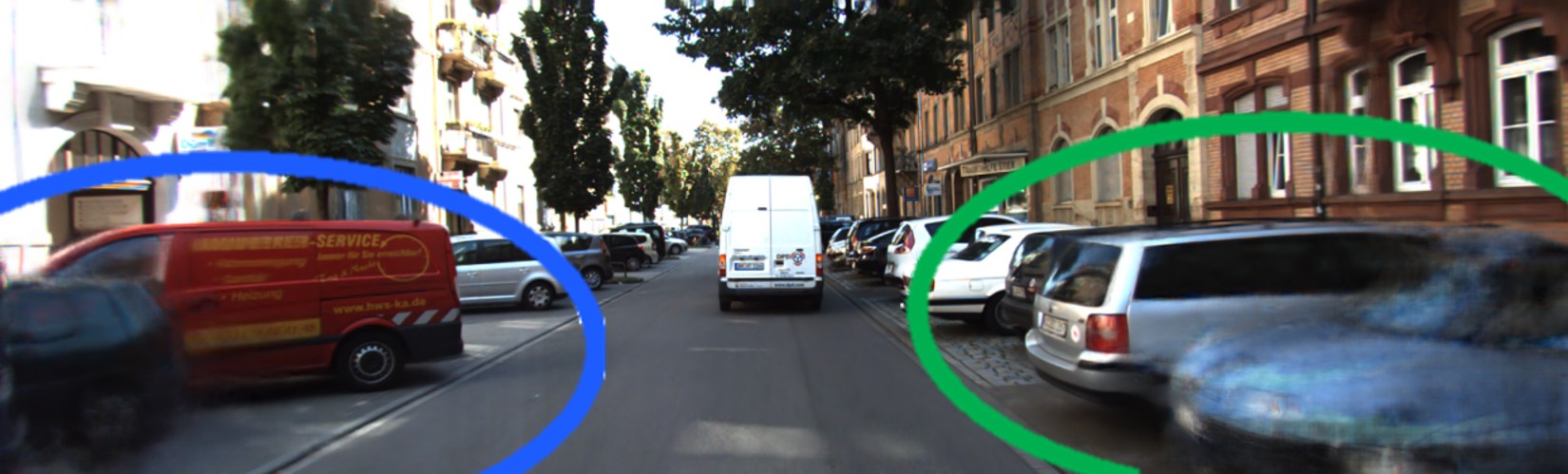}  \\
\hspace{-0.1cm} \rotatebox{90}{\hspace{1.5mm} \tiny{OCAI}}
 & \hspace{-0.1cm} \includegraphics[width=2.33cm,height=1.3cm]{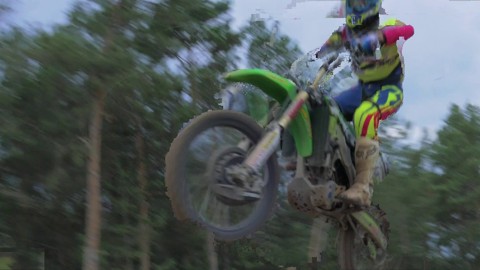} & \hspace{-0.15cm} 
\includegraphics[width=2.33cm,height=1.3cm]{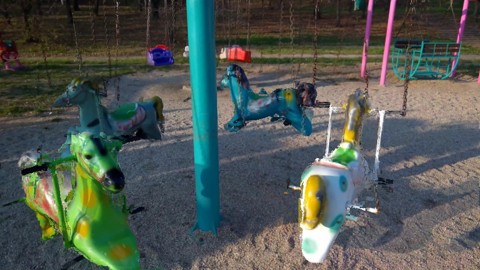}& \hspace{-0.15cm} 
\includegraphics[width=2.33cm,height=1.3cm]{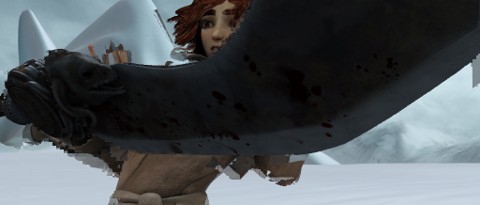}& \hspace{-0.15cm} 
\includegraphics[width=2.33cm,height=1.3cm]{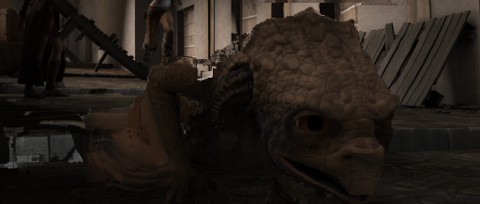}& \hspace{-0.15cm} 
\includegraphics[width=2.33cm,height=1.3cm]{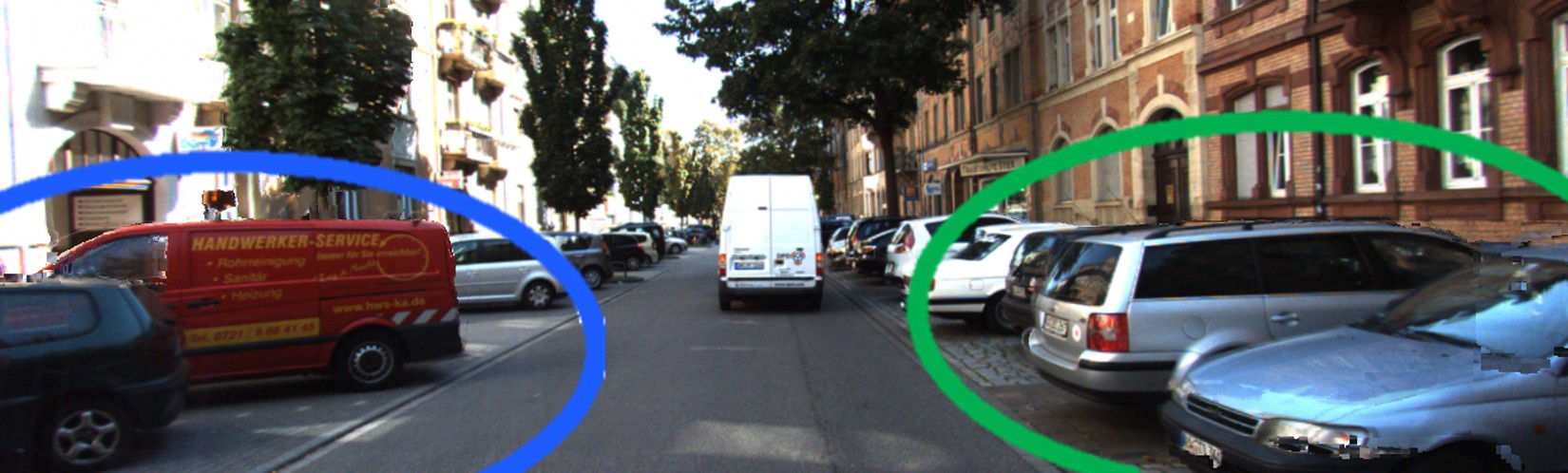}  \\
\hspace{-0.1cm} \rotatebox{90}{\hspace{1.0mm} \tiny{LDMVFI}}
 & \hspace{-0.1cm} \includegraphics[width=2.33cm,height=1.3cm]{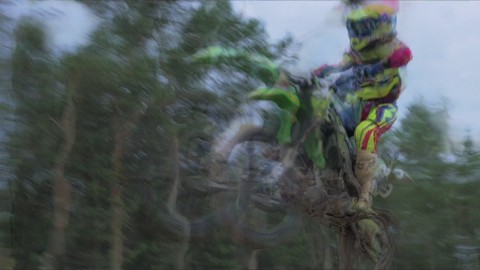} & \hspace{-0.15cm} 
\includegraphics[width=2.33cm,height=1.3cm]{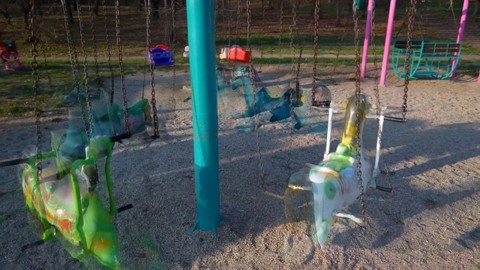}& \hspace{-0.15cm} 
\includegraphics[width=2.33cm,height=1.3cm]{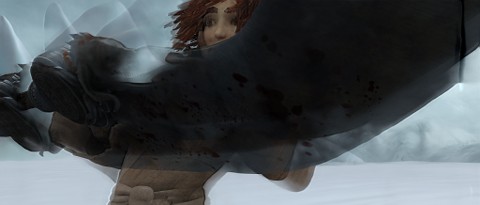}& \hspace{-0.15cm} 
\includegraphics[width=2.33cm,height=1.3cm]{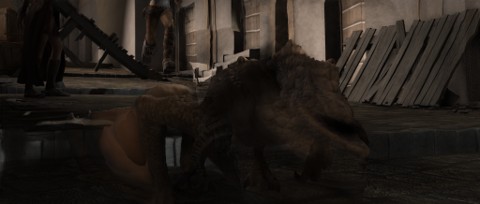}& \hspace{-0.15cm} 
\includegraphics[width=2.33cm,height=1.3cm]{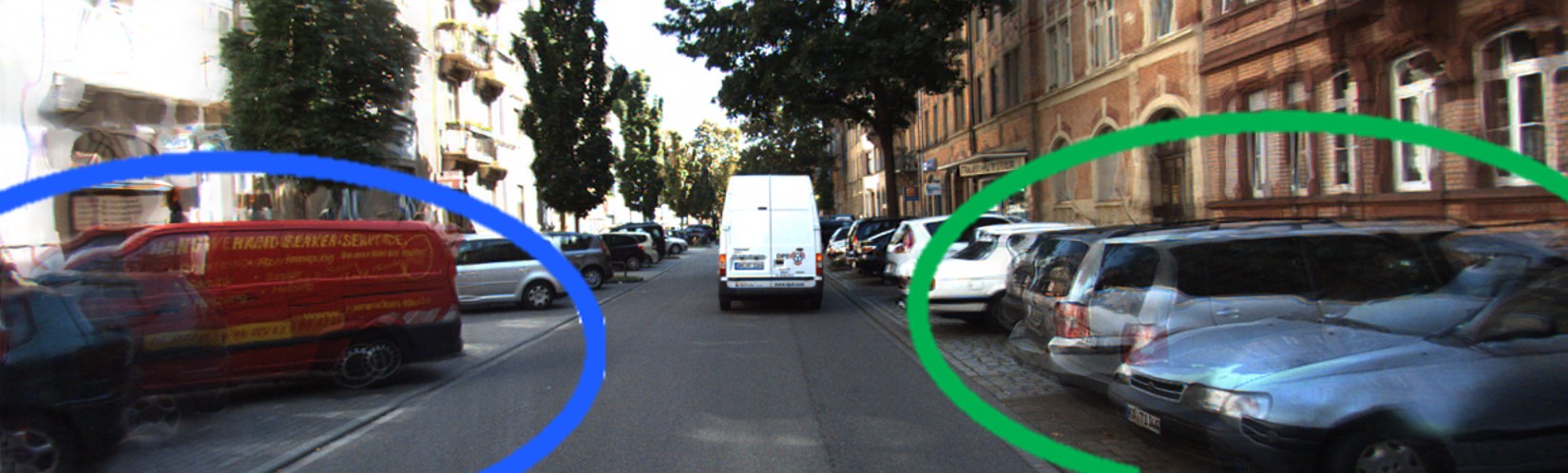}  \\
\hspace{-0.1cm} \rotatebox{90}{\hspace{1.5mm} \tiny{GenIn}}
 & \hspace{-0.1cm} \includegraphics[width=2.33cm,height=1.3cm]{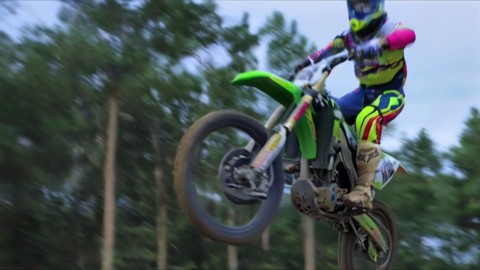} & \hspace{-0.15cm} 
\includegraphics[width=2.33cm,height=1.3cm]{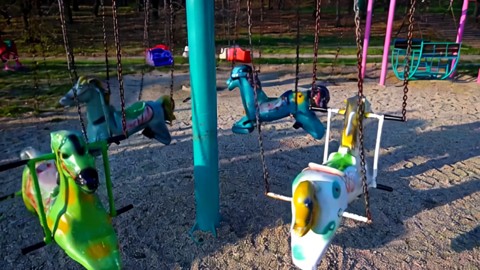}& \hspace{-0.15cm} 
\includegraphics[width=2.33cm,height=1.3cm]{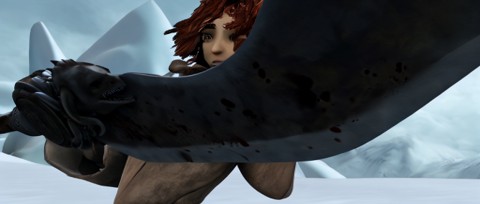}& \hspace{-0.15cm} 
\includegraphics[width=2.33cm,height=1.3cm]{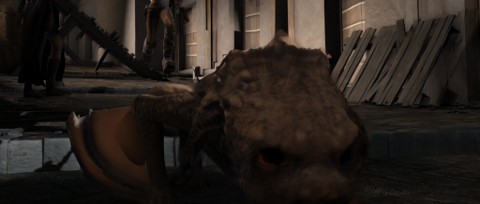}& \hspace{-0.15cm} 
\includegraphics[width=2.33cm,height=1.3cm]{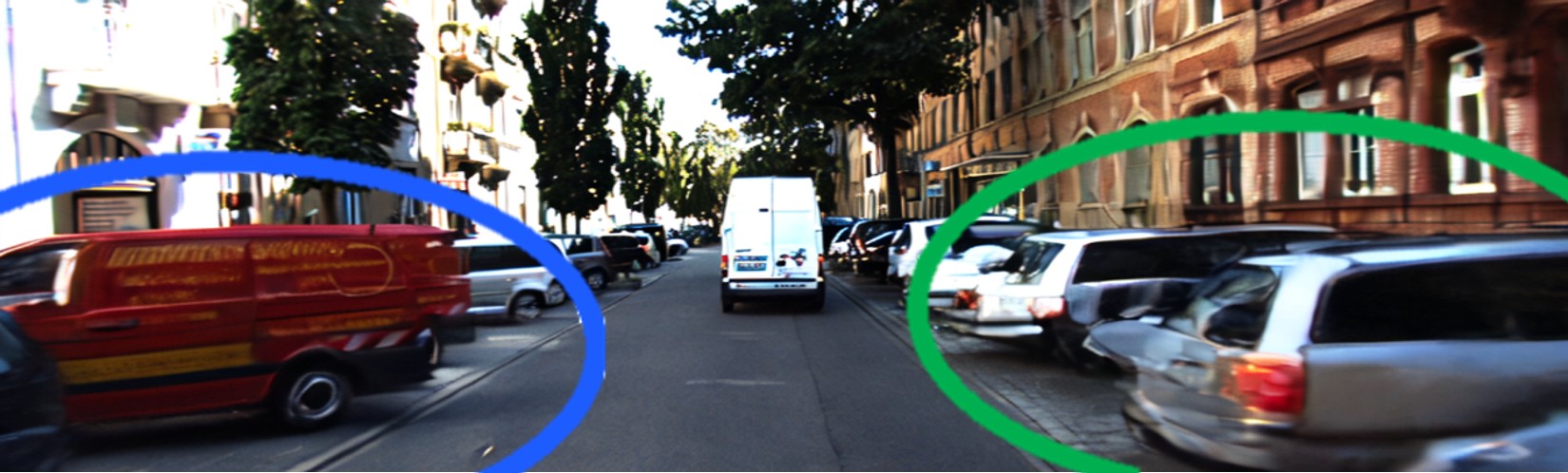}  \\
\hspace{-0.1cm} \rotatebox{90}{\hspace{0.7mm} \tiny{\ours}}
 & \hspace{-0.1cm} \includegraphics[width=2.33cm,height=1.3cm]{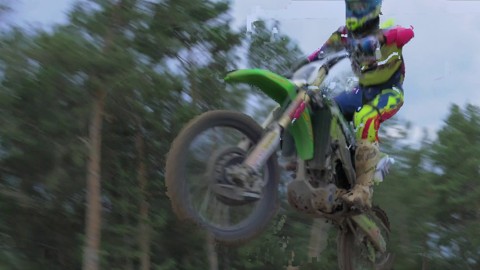} & \hspace{-0.15cm} 
\includegraphics[width=2.33cm,height=1.3cm]{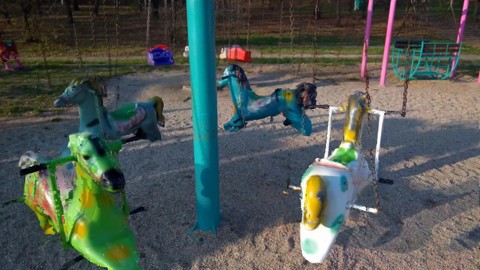}& \hspace{-0.15cm} 
\includegraphics[width=2.33cm,height=1.3cm]{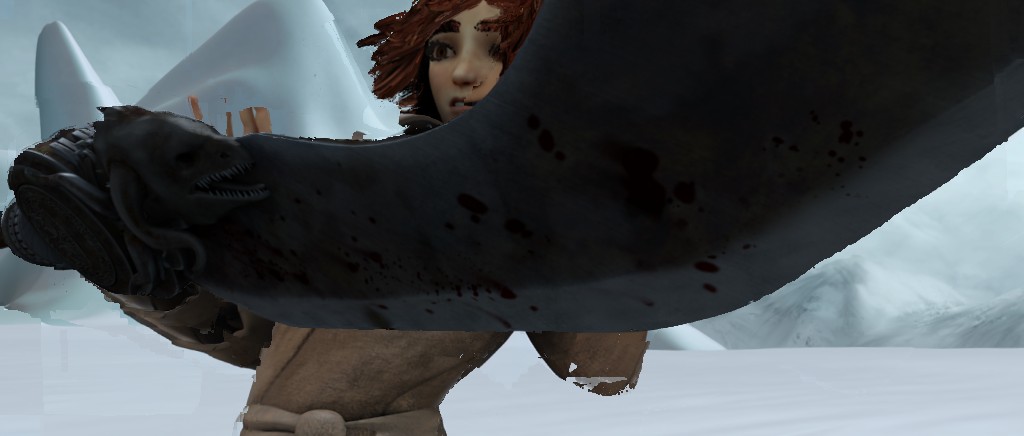}& \hspace{-0.15cm} 
\includegraphics[width=2.33cm,height=1.3cm]{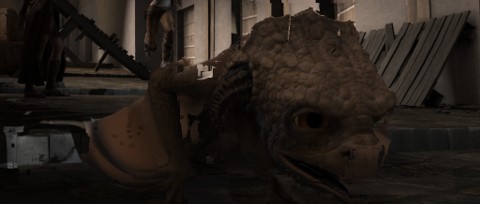}& \hspace{-0.15cm} 
\includegraphics[width=2.33cm,height=1.3cm]{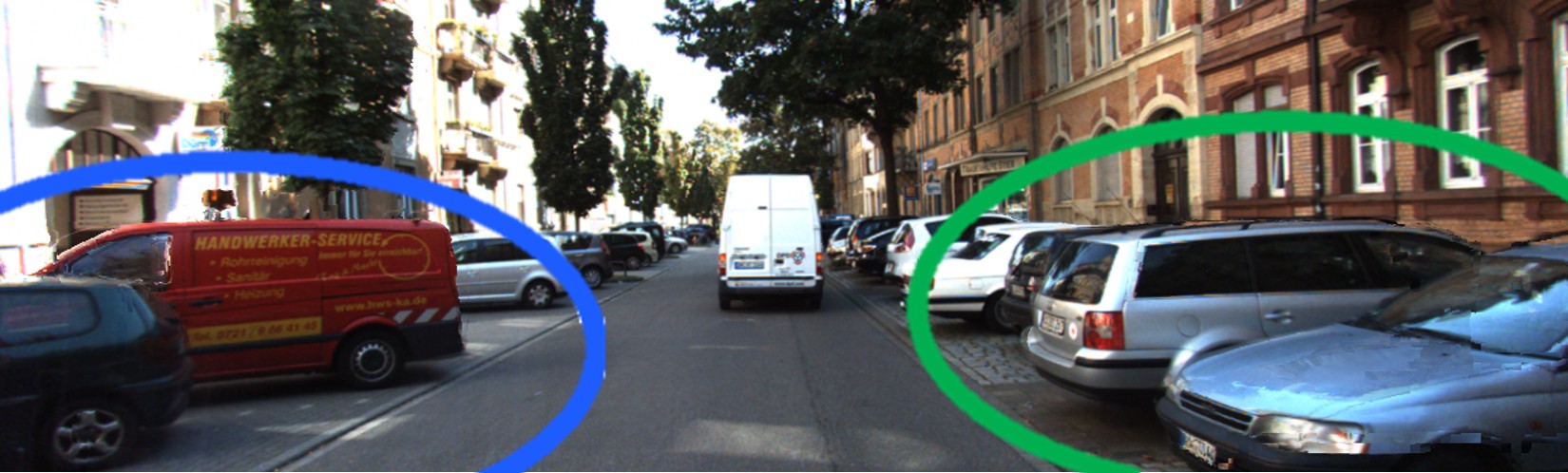}  \\

\vspace{-30pt}
\end{tabular}$
\end{center}
\caption{Video frame interpolation results ($\times$2 interpolation) on DAVIS, Sintel, and KITTI. The fourth to sixth rows show results from flow-based methods: VFIFormer and EMA-VFI (backward warping) and OCAI (forward warping). The seventh and eighth rows present results from diffusion-based methods: LDMVFI and GenIn. The final row displays the output of our proposed \ours method. 
}
\vspace{-10pt}
\label{fig:experiment_vfi}
\end{figure}

\vspace{-5pt}
\subsection{Qualitative Results}
\vspace{-2mm}
Fig.~\ref{fig:experiment_vfi} presents the $\times$2 interpolation results from state-of-the-art VFI methods and our proposed approach. 
Backward warping-based methods such as VFIFormer~\cite{lu2022video} and EMA-VFI~\cite{zhang2023extracting} struggle with large motion, producing blurred results despite high PSNR and SSIM scores in Table~\ref{tab:middle2}, especially in regions with complex motion.
Forward warping-based methods like OCAI demonstrate better robustness to large motion but tend to introduce noise around object boundaries. Diffusion‑based approaches can generate perceptually realistic frames, but variations in brightness and contrast frequently lead to outputs that appear plausible yet deviate from the ground truth frame, resulting in reduced reconstruction accuracy. Moreover, diffusion-based methods often lack robustness to large motion. For example, in GenIn~\cite{wang2024generative}, the second Sintel example shows almost no change from $I_{0}$ while the KITTI example hallucinates new contents (in blue and green boxes), resulting in noticeable blur and temporal inconsistency.

\begin{figure}[t]
\begin{center}$
\centering
\begin{tabular}{ c cc cc c}

& GT & VFIFormer & OCAI & GenIn & \ours \\

\hspace{-0.1cm} \rotatebox{90}{\hspace{3.5mm} \scriptsize{$I_{1/4}$}}
 & \hspace{-0.1cm} \includegraphics[width=2.33cm,height=1.3cm]{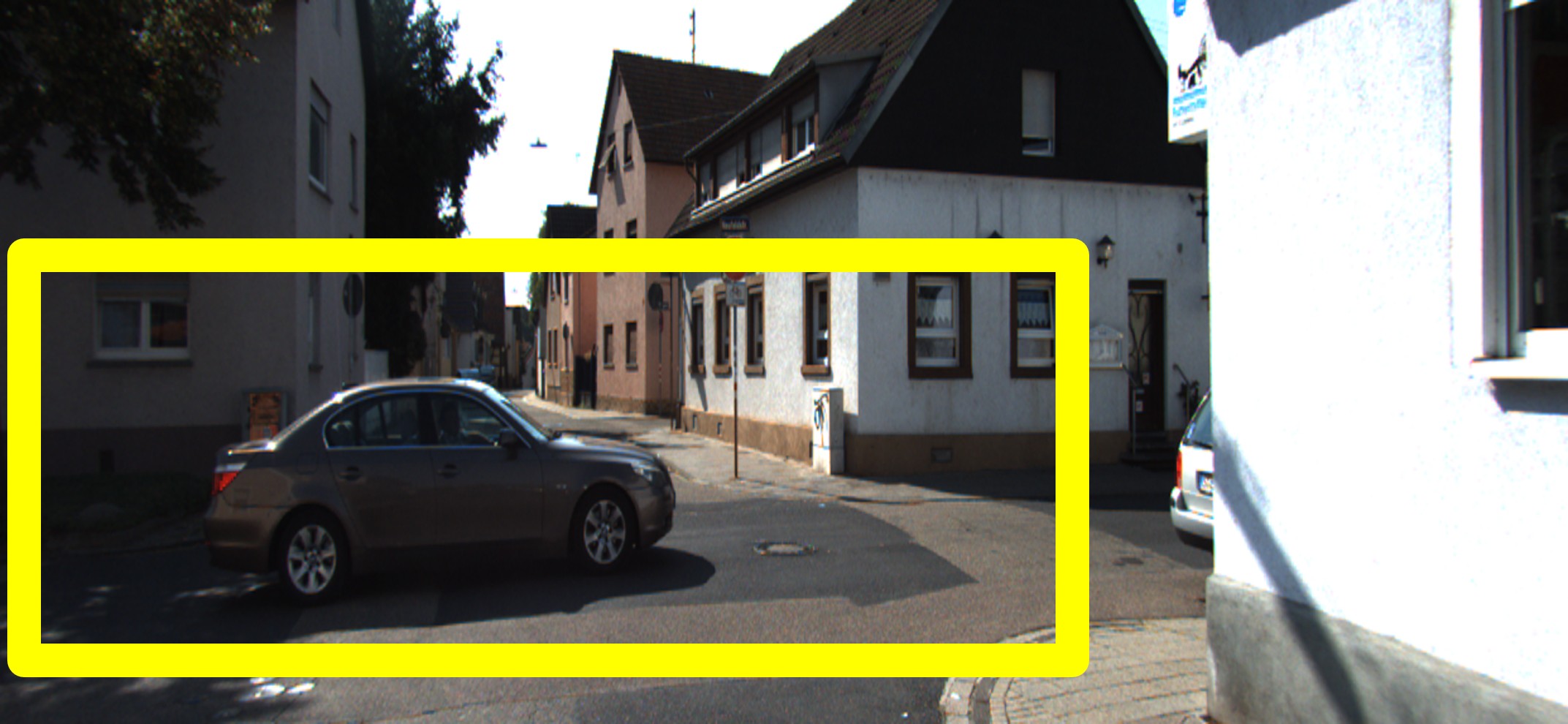} & \hspace{-0.15cm} 
\includegraphics[width=2.33cm,height=1.3cm]{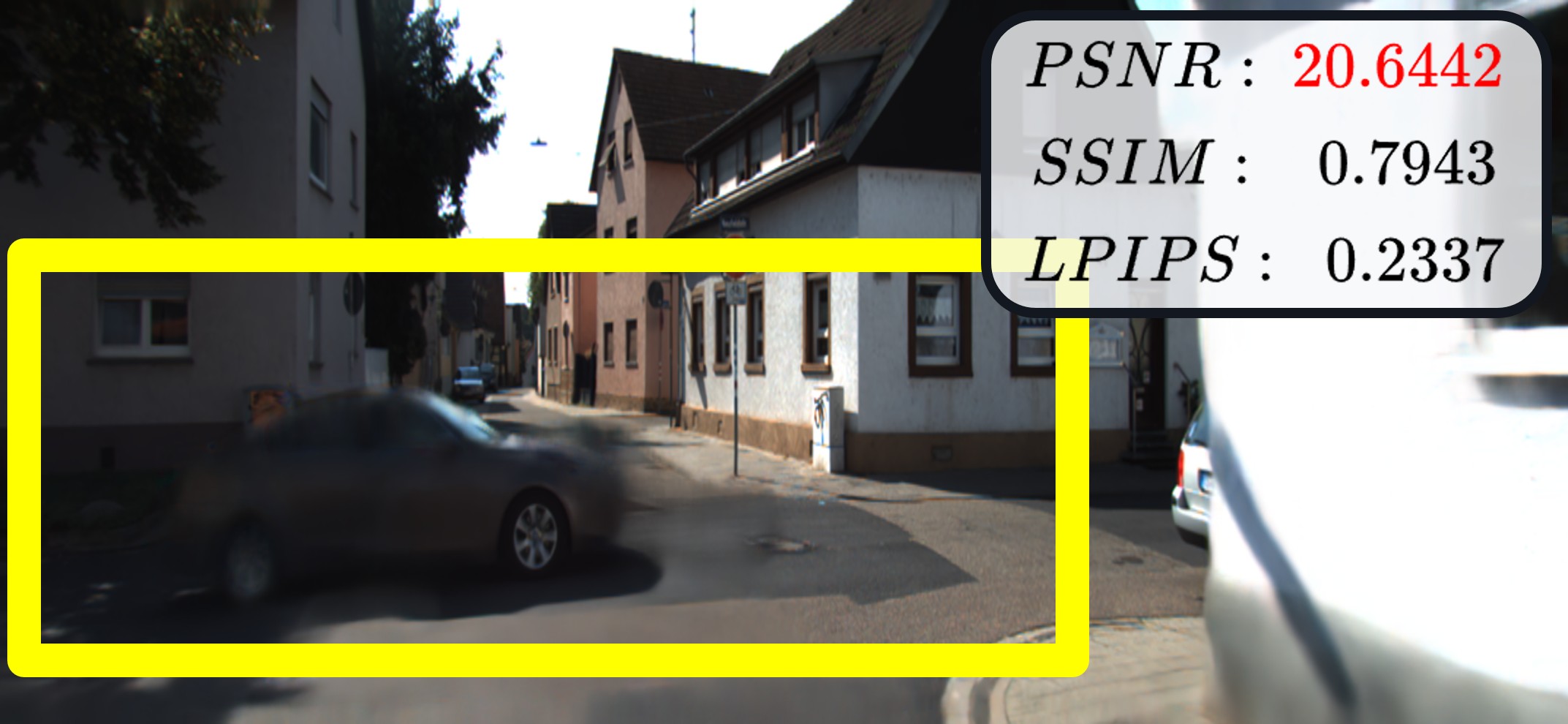}& \hspace{-0.15cm} 
\includegraphics[width=2.33cm,height=1.3cm]{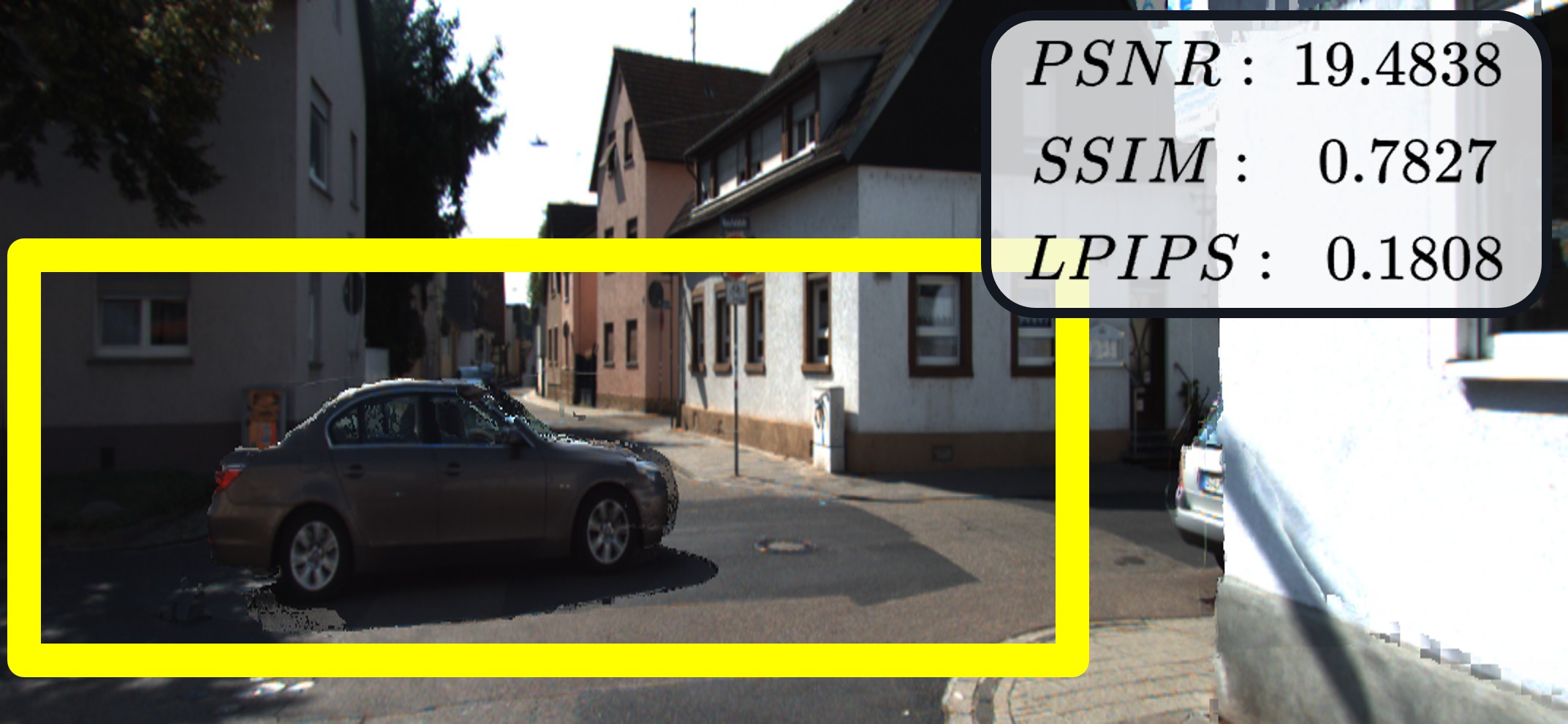}& \hspace{-0.15cm} 
\includegraphics[width=2.33cm,height=1.3cm]{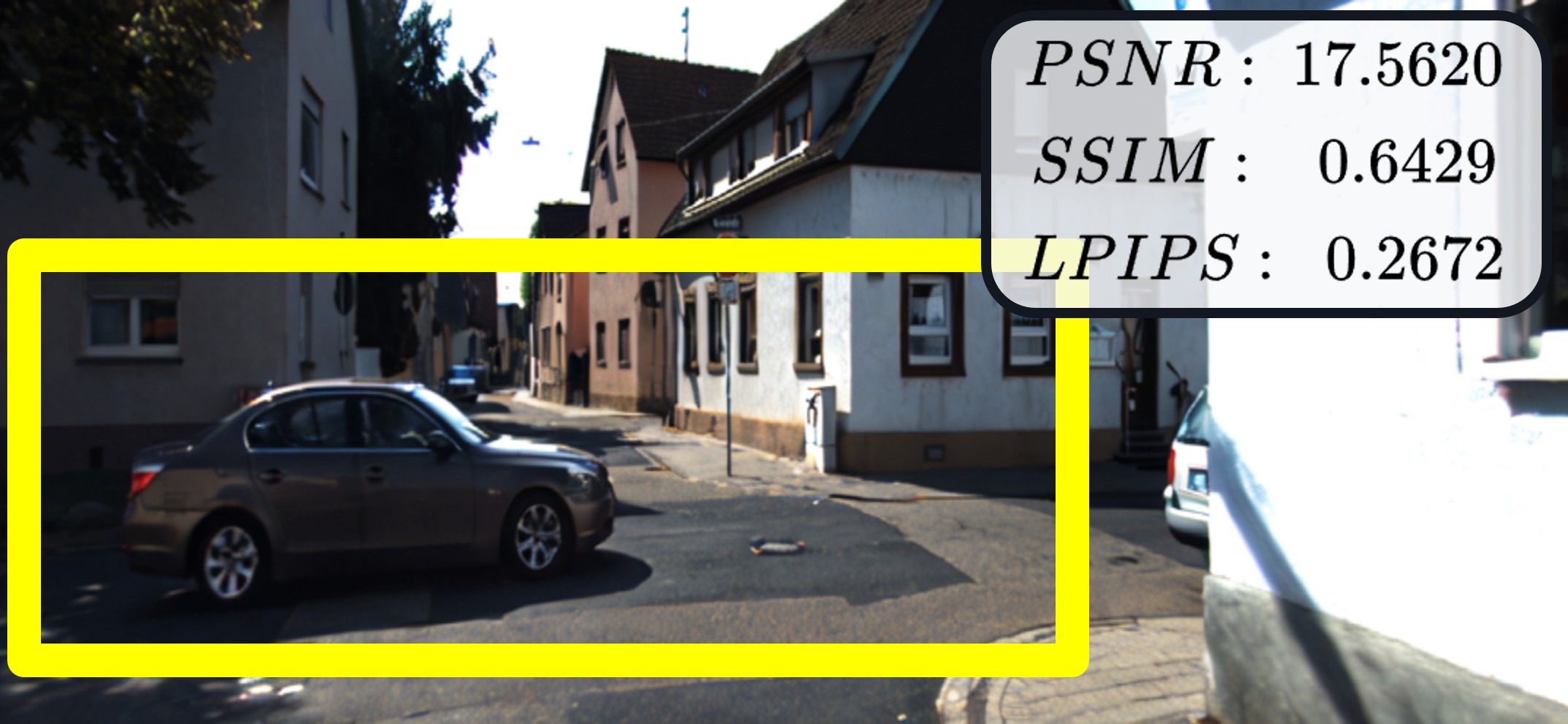}& \hspace{-0.15cm} 
\includegraphics[width=2.33cm,height=1.3cm]{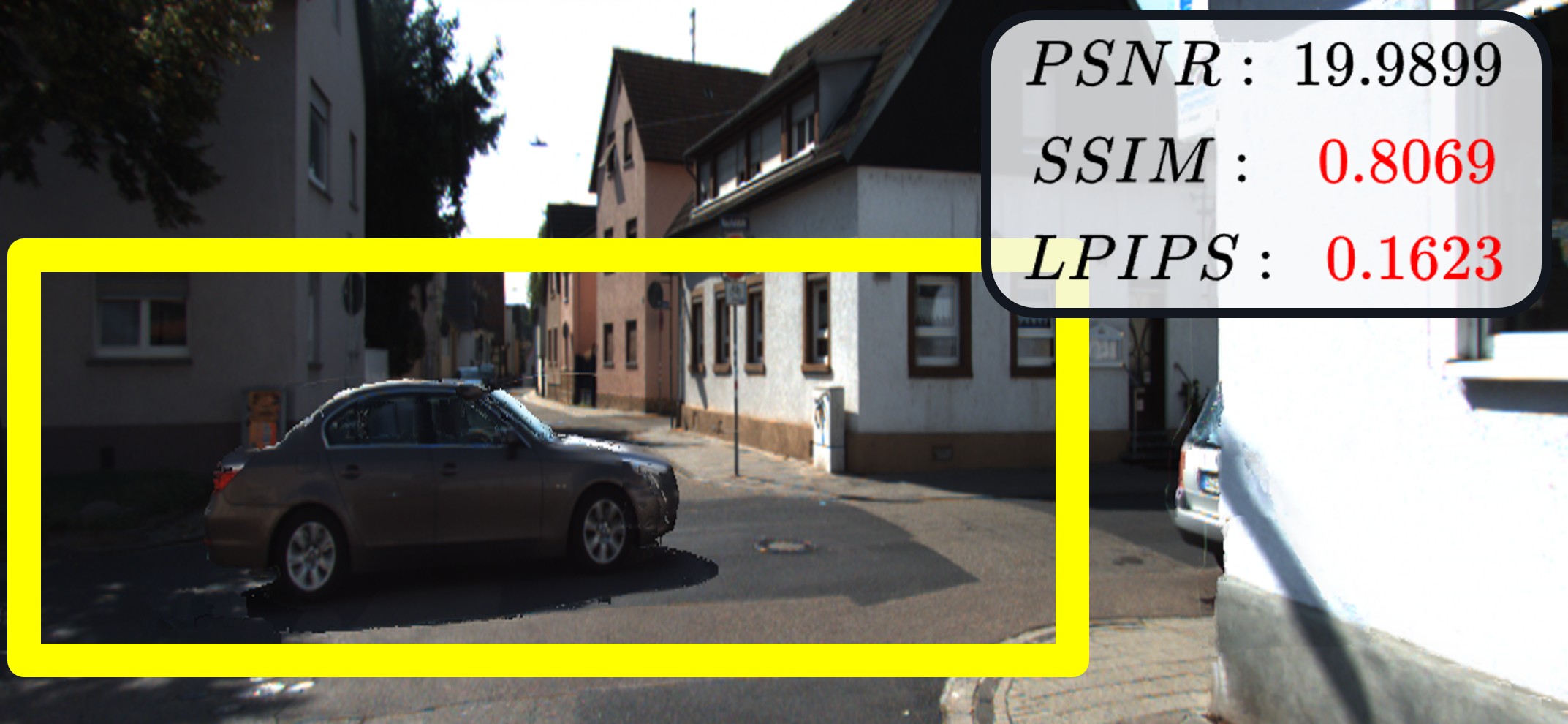}  \\

\hspace{-0.1cm} \rotatebox{90}{\hspace{3.5mm} \scriptsize{$I_{2/4}$}}
 & \hspace{-0.1cm} \includegraphics[width=2.33cm,height=1.3cm]{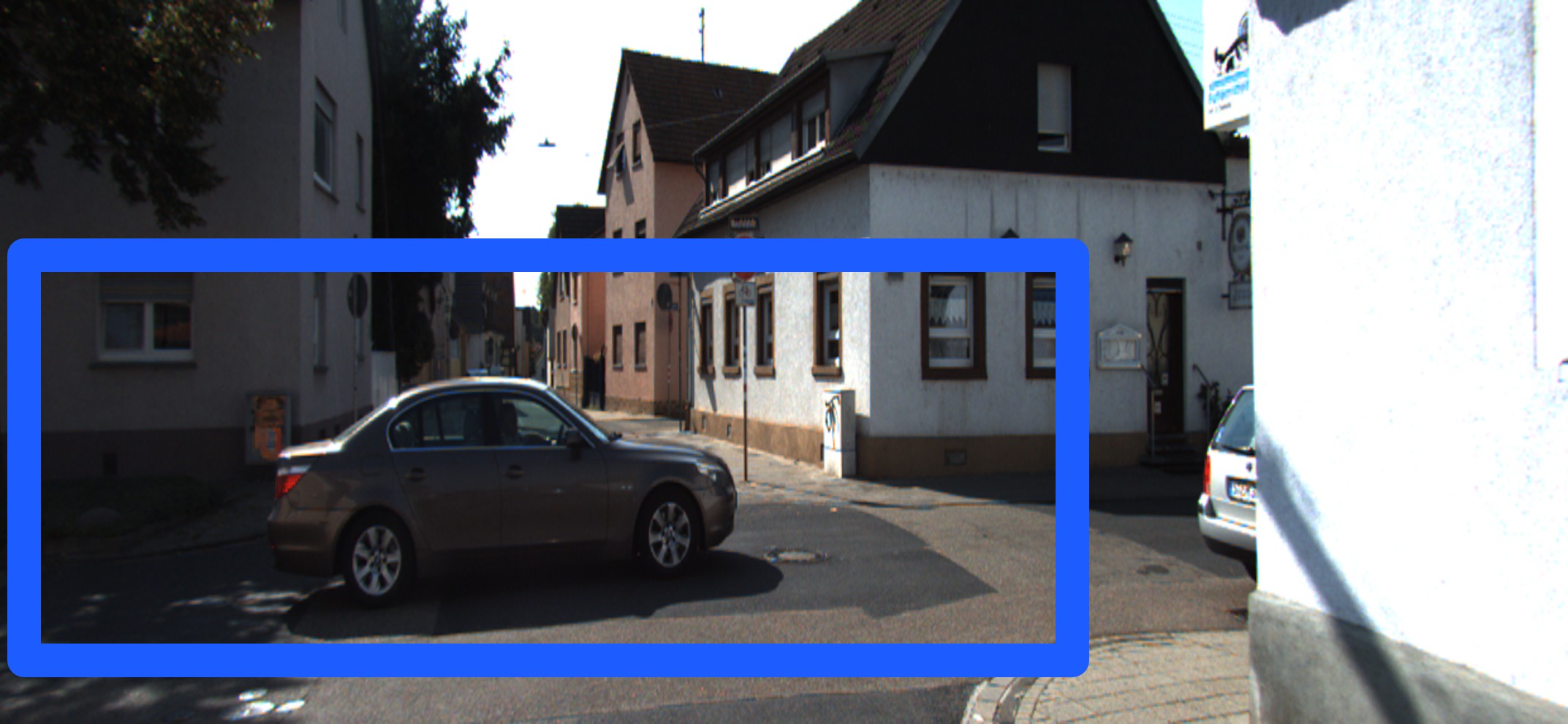} & \hspace{-0.15cm} 
\includegraphics[width=2.33cm,height=1.3cm]{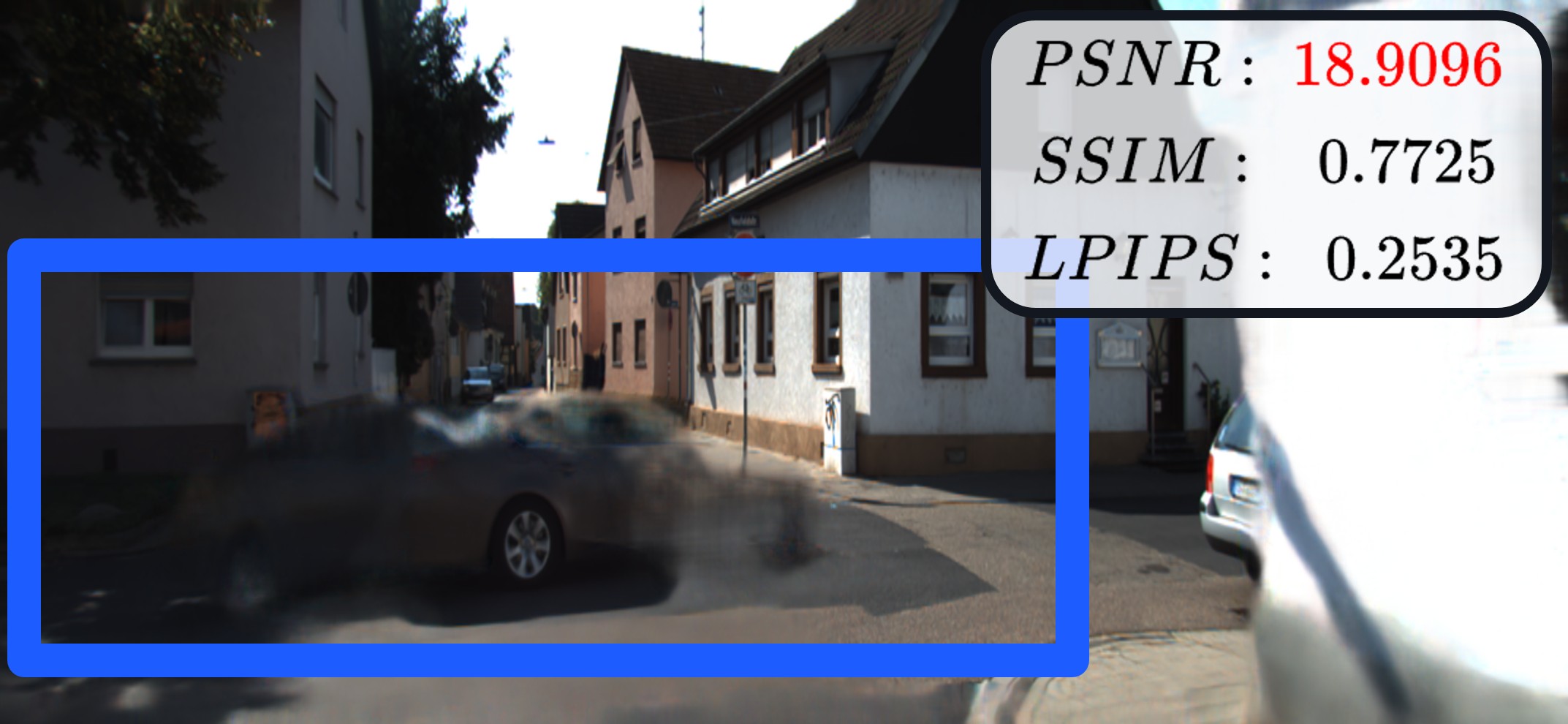}& \hspace{-0.15cm} 
\includegraphics[width=2.33cm,height=1.3cm]{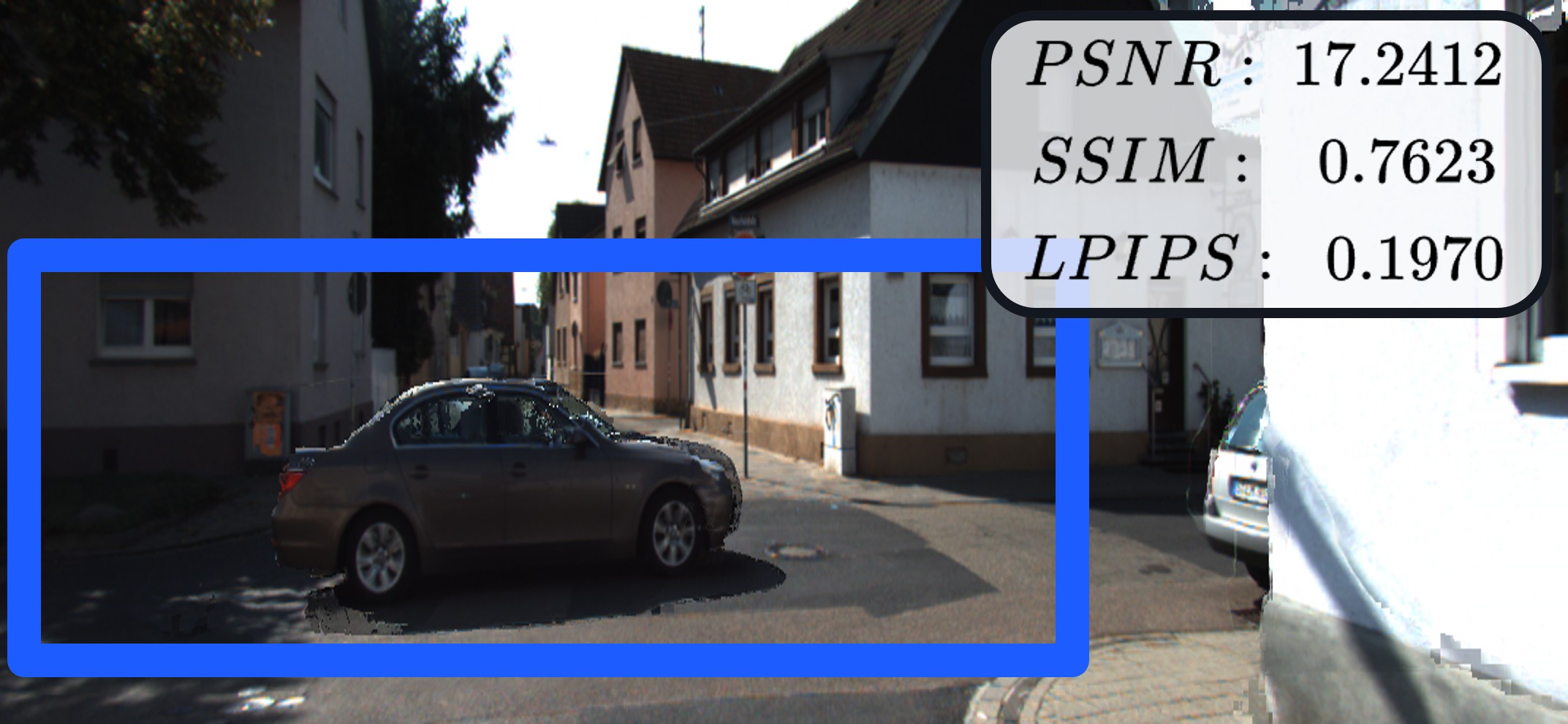}& \hspace{-0.15cm} 
\includegraphics[width=2.33cm,height=1.3cm]{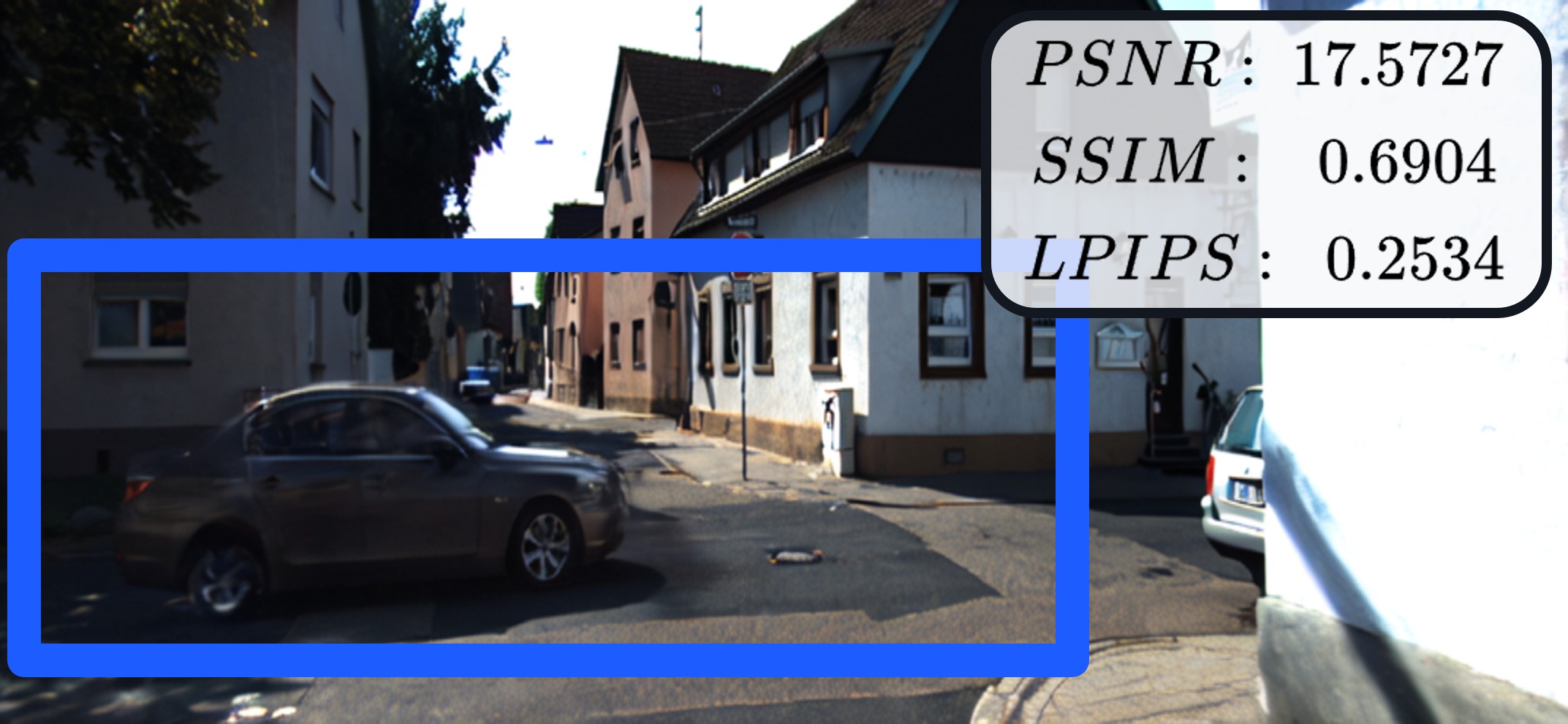}& \hspace{-0.15cm} 
\includegraphics[width=2.33cm,height=1.3cm]{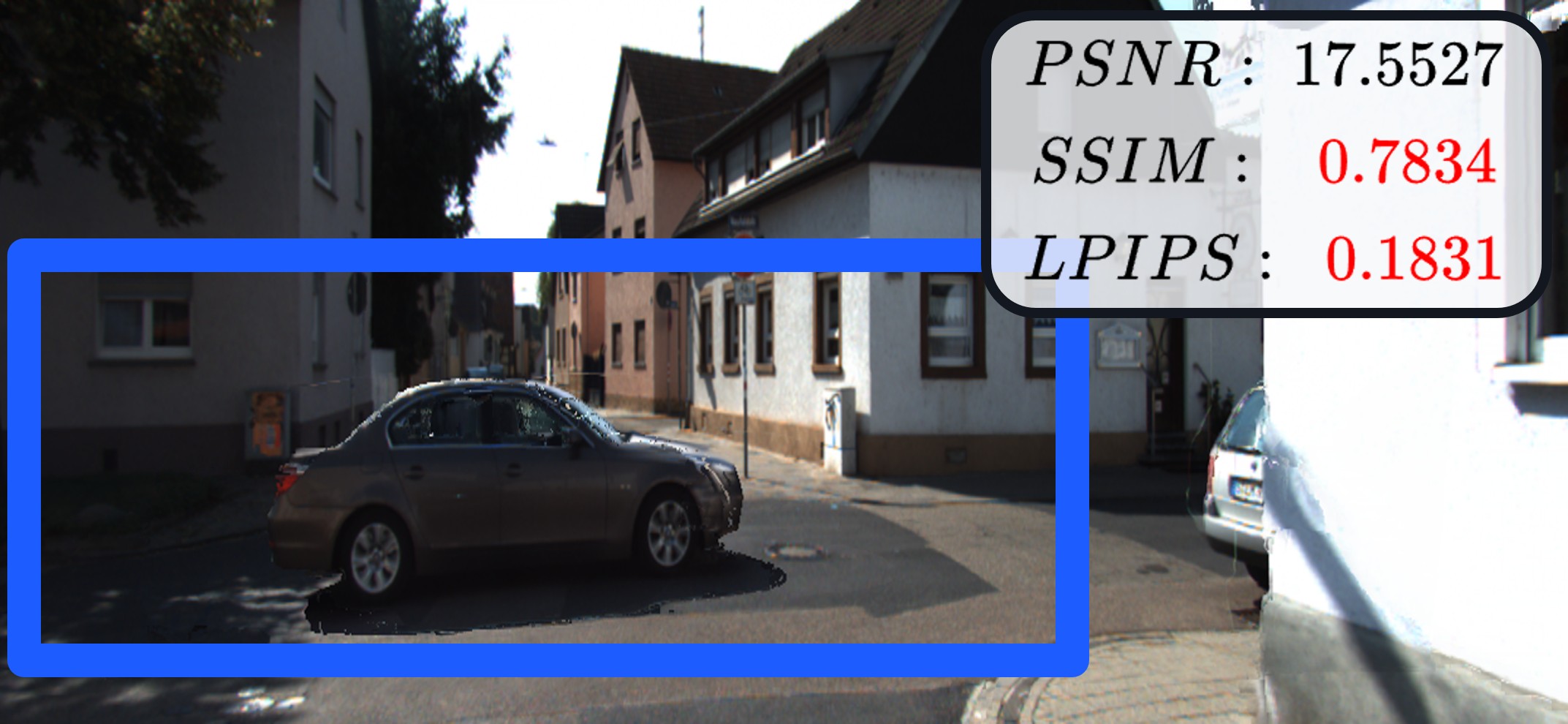}  \\

\hspace{-0.1cm} \rotatebox{90}{\hspace{3.5mm} \scriptsize{$I_{3/4}$}}
 & \hspace{-0.1cm} \includegraphics[width=2.33cm,height=1.3cm]{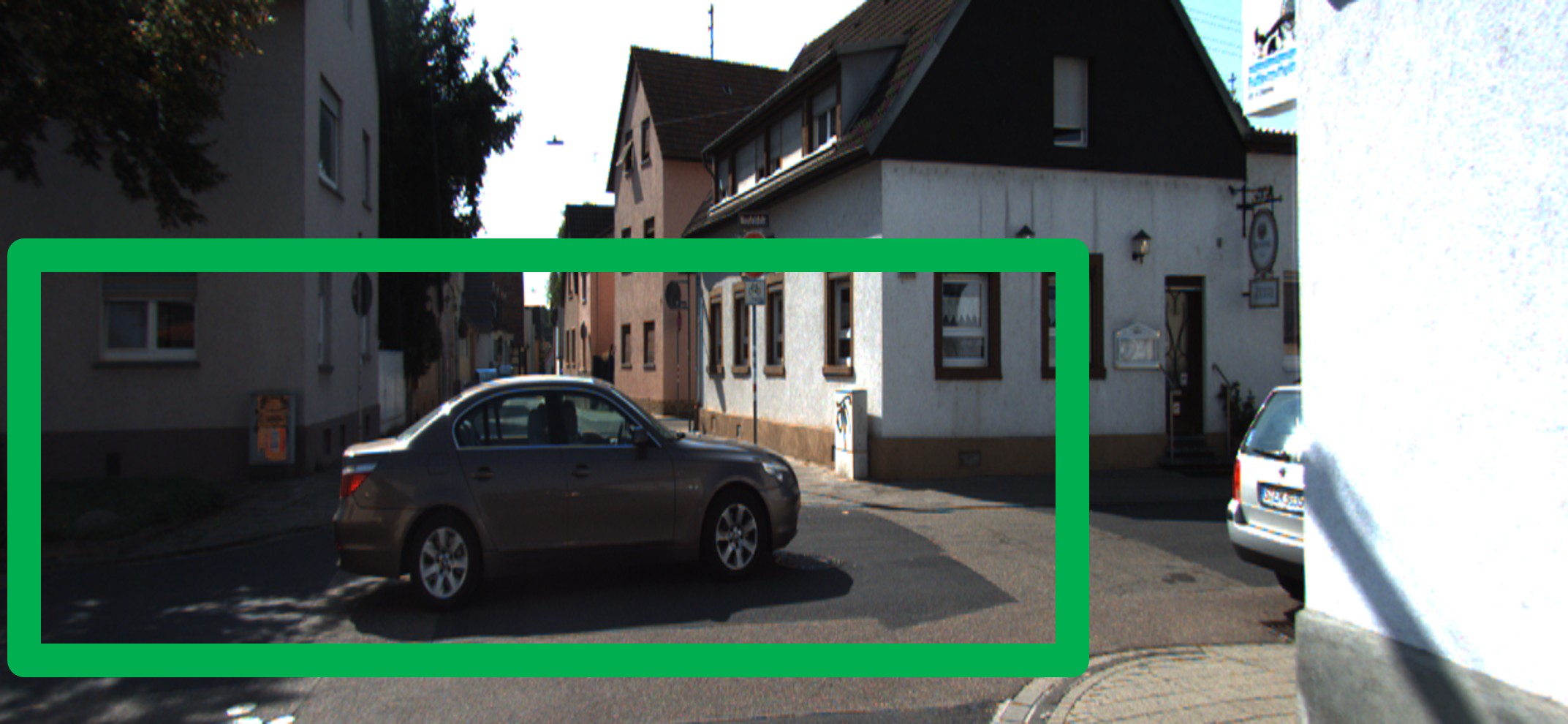} & \hspace{-0.15cm} 
\includegraphics[width=2.33cm,height=1.3cm]{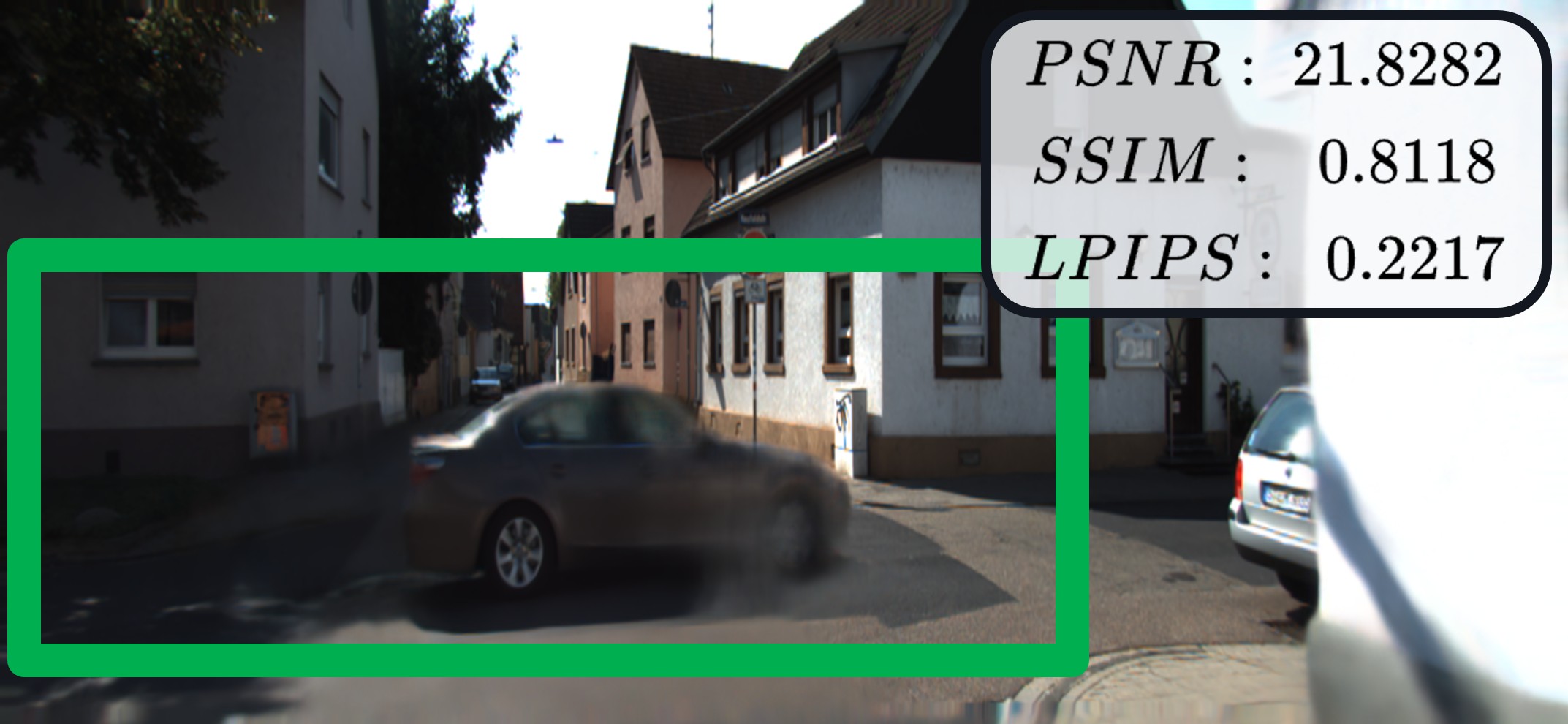}& \hspace{-0.15cm} 
\includegraphics[width=2.33cm,height=1.3cm]{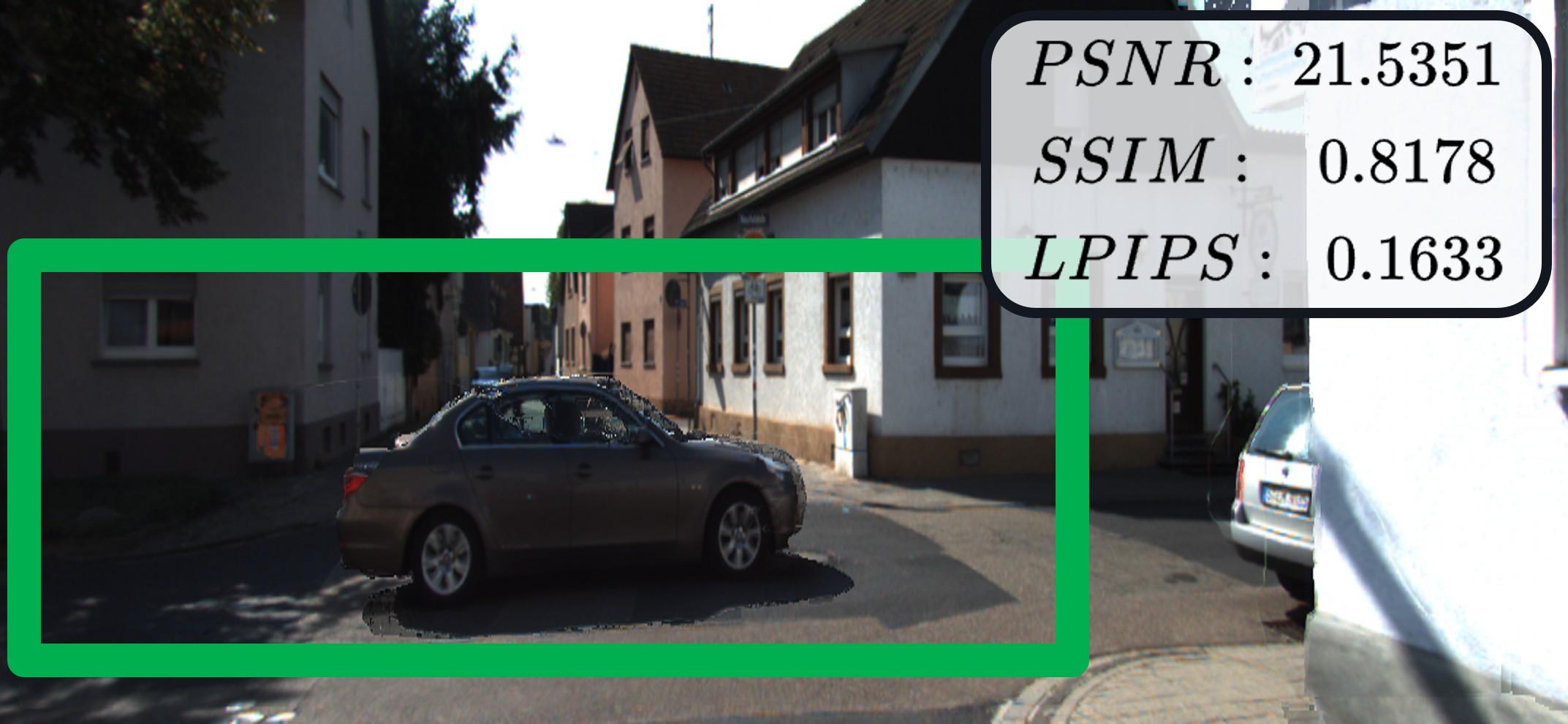}& \hspace{-0.15cm} 
\includegraphics[width=2.33cm,height=1.3cm]{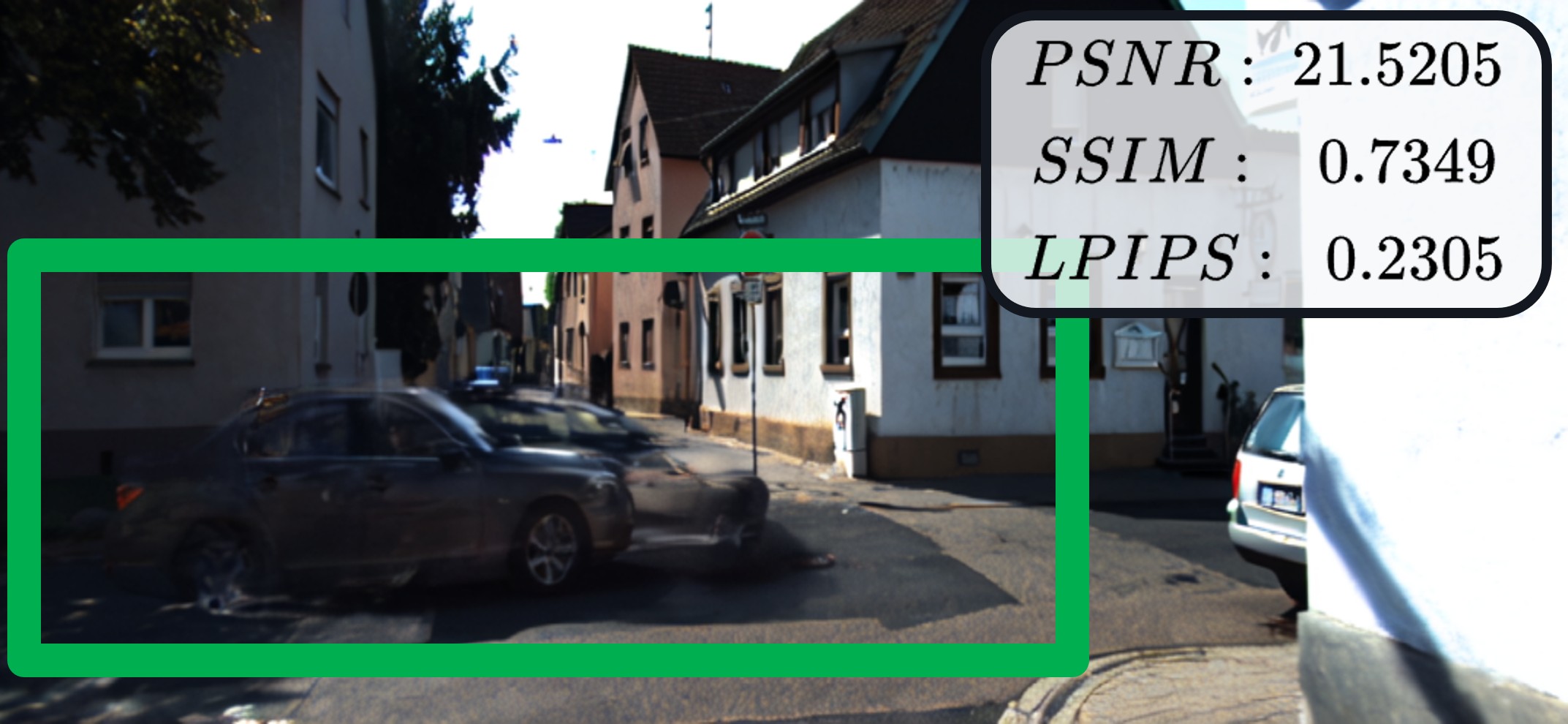}& \hspace{-0.15cm} 
\includegraphics[width=2.33cm,height=1.3cm]{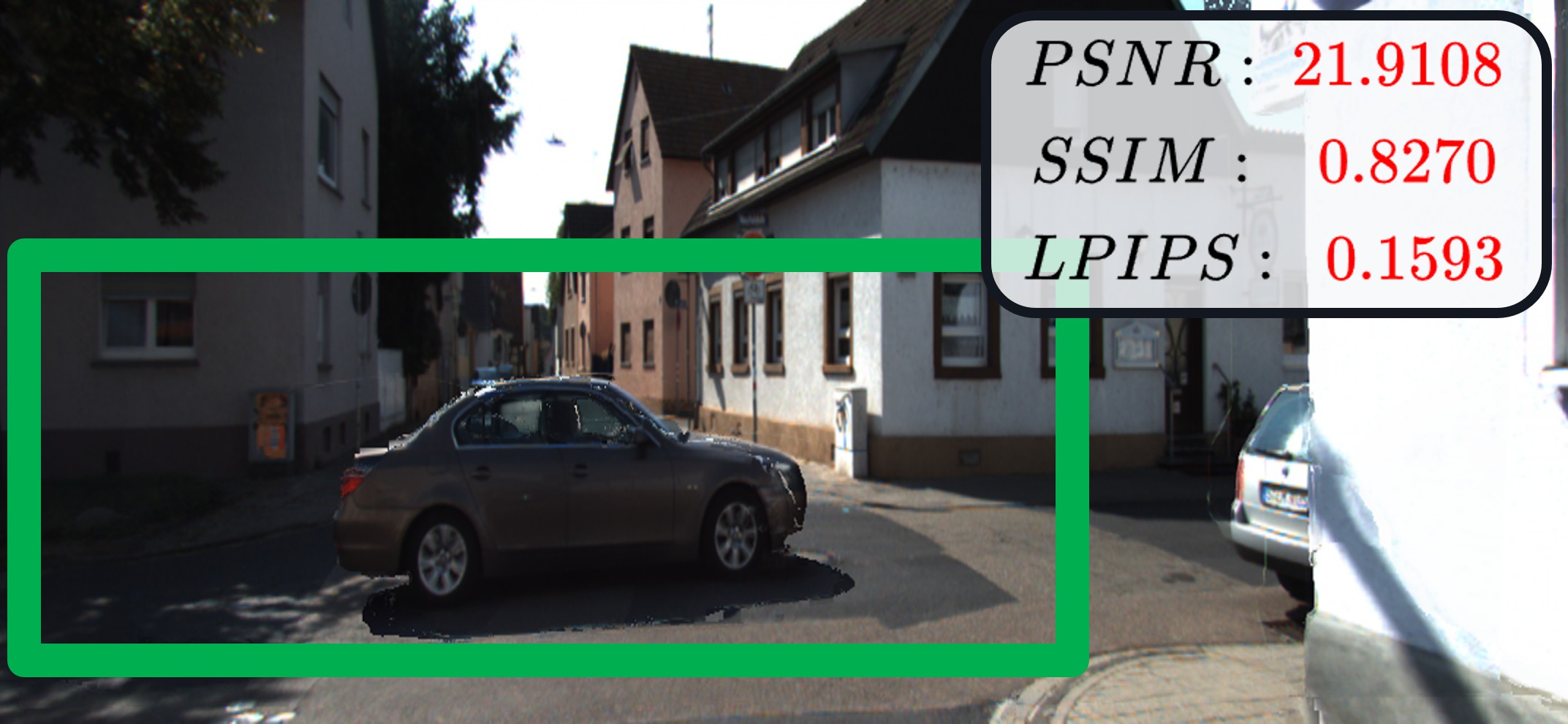}  \\

\vspace{-30pt}
\end{tabular}$
\end{center}
\caption{Video frame interpolation results ($\times$4 interpolation) on KITTI. The second and third columns show results from flow-based methods: VFIFormer (backward warping) and OCAI (forward warping). The fourth column presents result from diffusion-based method: GenIn. The final column displays the output of our \ours method. 
}
\vspace{-10pt}
\label{fig:experiment_vfi4x}
\end{figure}

In contrast, \ours effectively combines the strengths of optical flow and diffusion. It demonstrates strong robustness to large motion, generating coherent object trajectories while preserving smooth and accurate object boundaries. These results highlight the superior temporal consistency and perceptual quality achieved by our approach across diverse motion scenarios. Further qualitative results are available in the supplementary material.

Fig.~\ref{fig:experiment_vfi4x} shows the $\times$4 interpolation outputs produced by state of the art VFI methods and our proposed model. Similar to the $\times$2 interpolation setting, backward‑warping methods such as VFIFormer struggle with large‑motion objects, producing noticeably blurred frames. Forward‑warping methods like OCAI handle large motion more effectively but still introduce noise along object boundaries. Diffusion‑based method (GenIn) generates visually realistic frames, yet it fails to maintain correspondence between objects across frames. Objects may remain static until $I_{2/4}$ and then reappear at a different location, leading to temporal inconsistency.

In contrast, our method accurately captures object motion while producing clean and stable frames. Although our qualitative results are significantly better, there are a few cases where VFIFormer reports a higher PSNR. This highlights that PSNR does not always correlate with perceived visual quality.
\section{Discussion}
\label{sec:dis}

\begin{table*}[t!]
\begin{center}
\caption{Ablation study on \ours. We use common image evaluation metrics, including PSNR ($\uparrow$), SSIM ($\uparrow$), LPIPS ($\downarrow$), and FID ($\downarrow$). \textcolor{red}{Red}: \textcolor{red}{Best} result.}
\label{tab:ablation}
\vspace{-10pt}
\adjustbox{max width=0.98\textwidth}
{
\begin{tabular}{|l||c|c|c|}
\hline
\multirow{2}*{ \cellcolor{mycolor2} Methods} 
\cellcolor{mycolor2} & \cellcolor{mycolor2} DAVIS (4$\times$) & \cellcolor{mycolor2} Sintel (4$\times$) & \cellcolor{mycolor2} KITTI (4$\times$) \\
\cline{2-4}
\multirow{-2}*{ \cellcolor{mycolor2} Methods} & \cellcolor{mycolor2} \hspace{1pt}PSNR\hspace{1pt}/\hspace{1pt}SSIM\hspace{1pt}/\hspace{1pt}LPIPS\hspace{1pt}/ \ FID \ \ \ & \cellcolor{mycolor2} \hspace{1pt}PSNR\hspace{1pt}/\hspace{1pt}SSIM\hspace{1pt}/\hspace{1pt}LPIPS\hspace{1pt}/ \ FID \ \ \ & \cellcolor{mycolor2} \hspace{1pt}PSNR\hspace{1pt}/\hspace{1pt}SSIM\hspace{1pt}/\hspace{1pt}LPIPS\hspace{1pt}/ \ FID \ \ \ \\
\hline
\hline
\rowcolor{mycolor} \multicolumn{4}{|c|}{Symmetric Nonlinear motion based flow interpolation} \\
\hline
Baseline (Linear)~\cite{jeong2024ocai}  &23.7432/0.8063/0.1864/\textcolor{red}{51.2046}&25.8927/ 0.8402/0.1747/\ \textcolor{red}{93.8831}&18.8717/0.6650/ 0.2620/\textcolor{red}{29.3621}\\
Quadratic~\cite{xu2019quadratic} &23.8200/0.8134/0.1956/56.6269&\textcolor{red}{26.1876}/\textcolor{red}{0.8509}/0.1752/\ 97.2842&19.0454/0.6789/0.2606/30.0474\\
Enhanced Quadratic~\cite{liu2020enhanced} &23.6344/0.8082/0.2030/57.9580&26.1238/0.8469/0.1792/\ 97.7645&18.9036/0.6732/0.2657/31.0025\\
Ours w/o Eq~\ref{our_occ} &24.0158/0.8093/0.1865/55.6194&26.1069/0.8463/\textcolor{red}{0.1702}/\ 95.5088&19.0517/0.6757/0.2577/29.4991\\
\rowcolor{myrowcolor} Ours (Symmetric Nonlinear) &\textcolor{red}{24.0369}/\textcolor{red}{0.8157}/\textcolor{red}{0.1831}/52.2948&26.1153/0.8477/0.1733/\ 96.2919&\textcolor{red}{19.1260}/\textcolor{red}{0.6798}/\textcolor{red}{0.2575}/29.8282\\
\hline
\rowcolor{mycolor} \multicolumn{4}{|c|}{$\alpha$ hyperparameter in Eq.~\ref{eq:our0tapp}} \\
\hline
Nonlinear motion ($\alpha$=1/4) &23.9485/0.8130/0.1834/53.7636&26.0455/0.8446/\textcolor{red}{0.1727}/\ 98.2369&19.0456/0.6742/0.2581/\textcolor{red}{29.2620}\\
\rowcolor{myrowcolor} Nonlinear motion ($\alpha$=2/4) &\textcolor{red}{24.0369}/\textcolor{red}{0.8157}/\textcolor{red}{0.1831}/\textcolor{red}{52.2948}&\textcolor{red}{26.1153}/\textcolor{red}{0.8477}/0.1733/\ \textcolor{red}{96.2919}&\textcolor{red}{19.1260}/\textcolor{red}{0.6798}/\textcolor{red}{0.2575}/29.8282\\
Nonlinear motion ($\alpha$=3/4) &23.9424/0.8118/0.1870/55.4992&26.0261/0.8450/0.1763/100.0384&18.9847/0.6736/0.2614/30.4168\\
Nonlinear motion ($\alpha$=1.0) &23.7348/0.8054/0.1920/56.8010&25.7951/0.8394/0.1807/104.8856&18.8022/0.6643/0.2668/30.9530\\

\hline
\rowcolor{mycolor}  \multicolumn{4}{|c|}{Diffusion-based VFI ablation studies} \\
\hline
  Baseline~\cite{wang2024generative} &18.7968/0.6696/0.2967/54.9852&18.5285/0.6414/0.3443/121.9460&17.1569/0.5855/0.3186/25.6861\\
Baseline + flow init &19.2524/0.6908/0.3152/62.7582&20.3186/0.7061/0.3112/129.3442&17.1438/0.6057/0.3297/28.1318\\
\rowcolor{myrowcolor} Baseline + weight latent &\textcolor{red}{21.8624}/\textcolor{red}{0.7696}/\textcolor{red}{0.2344}/\textcolor{red}{51.3644}&\textcolor{red}{23.2930}/\textcolor{red}{0.7926}/\textcolor{red}{0.2215}/\textcolor{red}{110.9622}&19.2261/\textcolor{red}{0.6750}/\textcolor{red}{0.2803}/25.8986\\
Baseline + conf weight latent &21.8168/0.7693/0.2360/54.1466&23.1672/0.7908/0.2257/115.5835&\textcolor{red}{19.2286}/0.6736/0.2815/\textcolor{red}{25.2510}\\
\hline
\rowcolor{mycolor}  \multicolumn{4}{|c|}{Fusion of Symmetric Nonlinear VFI and Diffusion VFI} \\
\hline
Confidence Masking ($\delta$=0.5) &24.0245/0.8178/0.1837/49.5342&26.1647/0.8496/0.1711/\ 93.5126&19.5233/0.6987/0.2541/28.1704\\
Confidence Masking ($\delta$=1.0) &23.9757/0.8166/0.1902/49.7149&25.9308/0.8468/0.1818/\ 98.8999&19.3254/0.6913/0.2606/28.6371\\

\rowcolor{myrowcolor} Confidence Weighting (Eq.~\ref{final_fusion}) &\textcolor{red}{24.0859}/\textcolor{red}{0.8188}/\textcolor{red}{0.1814}/\textcolor{red}{48.5302}&\textcolor{red}{26.2414}/\textcolor{red}{0.8510}/\textcolor{red}{0.1688}/\ \textcolor{red}{92.4279}&\textcolor{red}{19.5978}/\textcolor{red}{0.7006}/\textcolor{red}{0.2519}/\textcolor{red}{27.6261}\\
\hline
\end{tabular}
}
\vspace{-20pt}
\end{center}
\end{table*}

\textbf{Nonlinear Motion:} 
The top section of Table~\ref{tab:ablation} compares our symmetric nonlinear motion approach with two alternatives: quadratic~\cite{xu2019quadratic} and enhanced quadratic motion~\cite{liu2020enhanced}. In this evaluation, we replace our formulation of $V_{0 \rightarrow t}$ with each variant. The quadratic motion variant utilizes current and past flows  $V_{0 \rightarrow 1}$ and $V_{0 \rightarrow -1}$, while the enhanced quadratic motion additionally incorporates $V_{0 \rightarrow 2}$.

Quadratic motion demonstrates improved performance over linear motion, yet generally underperforms our symmetric nonlinear method. An exception was observed on the Sintel dataset, where quadratic motion slightly outperforms in PSNR and SSIM. In contrast, enhanced quadratic motion consistently yields lower performance than both the baseline and our method. This decline can be attributed to the use of $V_{0 \rightarrow 2}$, which becomes less reliable in scenarios involving large temporal gaps, such as the $\times$4 interpolation setting, where predicting long-range motion is inherently more challenging.


Moreover, both quadratic and enhanced quadratic methods fail to account for occlusions, which can introduce significant errors in flow estimation and degrade interpolation quality. However, our method explicitly predicts occlusions and restricts nonlinear motion estimation to regions consistently visible across all four input frames. This occlusion-aware design helps mitigate flow errors (Ours w/o Eq.~\ref{our_occ} in Table~\ref{tab:ablation}) and contributes to the superior performance of our approach across diverse motion scenarios.

On the other hand, the linear model achieves the highest FID score. Although FID is categorized as a perceptual quality metric, it fundamentally measures dataset‑level realism rather than per‑frame accuracy. We hypothesize that the linear model, despite lower reconstruction accuracy, may generate smoother and more simplified intermediate frames, which can result in a better FID score.
However, our method explicitly addresses artifacts that arise from non‑linear motion (occluded regions) by applying a diffusion‑based refinement to low‑confidence areas. As a result, our final model achieves improved FID performance while maintaining strong reconstruction metrics.

\textbf{Hyperparameter $\alpha$: }
As shown in the second section of Table~\ref{tab:ablation}, we evaluate models using $\alpha$ values of $\frac{1}{4}$, $\frac{2}{4}$, $\frac{3}{4}$, 1.0 and observe that the setting of $\alpha$=$\frac{2}{4}$ generally yield the best performance, while the other values yield slightly worse but comparable performance. Based on this observation, we adopt $\alpha$=0.5 as our default and apply this setting across all experiments.

\textbf{Flow Guidance:} 
The third section of Table~\ref{tab:ablation} presents an ablation study on how flow-based cues are integrated into the diffusion model. As a first approach, we encode the flow-guided intermediate frames using a VAE encoder and use the resulting latent to replace the initial noise. This strategy leads to slight improvements in PSNR and SSIM compared to the baseline. However, since early diffusion steps introduce substantial changes, the influence of the flow-based priors may diminish as the denoising progresses.


To address this, we explore a second strategy in which the flow-guided latent features are fused with the diffusion latents via a weighted sum after each diffusion step. In this setup, the diffusion process starts from the flow-guided latent, and stronger flow-guidance is applied in the early denoising stages, while the influence of the diffusion updates is gradually increased in later steps. This dynamic fusion preserves motion cues throughout the denoising process, resulting in substantial improvements across PSNR, SSIM, and LPIPS metrics compared to the baseline.

Additionally, we experiment with confidence-based weighting, where the flow-guided latents are modulated using confidence maps derived from the flow estimation module. While this approach still outperforms the baseline, it is less effective than the full latent fusion strategy. These results suggest that directly blending the entire latent space with adaptive weighting yields more robust guidance than relying solely on confidence-based modulation.

\textbf{Fusion:} 
The final section of Table~\ref{tab:ablation} presents an ablation study on combining our symmetric nonlinear motion module with the flow-guided diffusion model. We estimate confidence using forward-backward consistency, enabling selective use of flow-based outputs in well-aligned regions, while occlusions and boundaries are handled by the diffusion model. The first fusion strategy applies confidence-based masking: flow-based outputs are used where confidence exceeds or equals a threshold ($\delta$), and diffusion outputs elsewhere. Experiments with thresholds of 0.5 and 1.0 show that 0.5 yields better results. We also evaluate confidence-based weighting (Eq.~\ref{final_fusion}), which outperforms masking by smoothly blending both sources. This selective fusion effectively integrates the structural precision of flow-based interpolation with the perceptual quality of diffusion, resulting in consistently high-quality outputs across diverse motion scenarios.


\section{Conclusion}
\label{sec:con}

We proposed a generative video frame interpolation framework that unifies flow-based precision with diffusion-based realism. By modeling nonlinear motion and guiding a diffusion model with flow-derived frames and confidence maps, our method achieves temporally coherent and visually consistent results. Experiments on DAVIS, Sintel, and KITTI confirm its superior performance across both accuracy and perceptual quality.

\newpage

\bibliographystyle{splncs04}
\bibliography{main}

@String(AAAI  = {AAAI})

@inproceedings{jeong2023distractflow,
  title={Distractflow: Improving optical flow estimation via realistic distractions and pseudo-labeling},
  author={Jeong, Jisoo and Cai, Hong and Garrepalli, Risheek and Porikli, Fatih},
  booktitle={Proceedings of the IEEE/CVF Conference on Computer Vision and Pattern Recognition},
  pages={13691--13700},
  year={2023}
}

@inproceedings{dosovitskiy2015flownet,
  title={Flownet: Learning optical flow with convolutional networks},
  author={Dosovitskiy, Alexey and Fischer, Philipp and Ilg, Eddy and Hausser, Philip and Hazirbas, Caner and Golkov, Vladimir and Van Der Smagt, Patrick and Cremers, Daniel and Brox, Thomas},
  booktitle={Proceedings of the IEEE/CVF International Conference on Computer Vision},
  pages={2758--2766},
  year={2015}
}

@article{hui2019lightweight,
  title={A lightweight optical flow CNN-revisiting data fidelity and regularization},
  author={Hui, Tak-Wai and Tang, Xiaoou and Loy, Chen Change},
  journal={arXiv preprint arXiv:1903.07414},
  year={2019}
}

@inproceedings{teed2020raft,
  title={Raft: Recurrent all-pairs field transforms for optical flow},
  author={Teed, Zachary and Deng, Jia},
  booktitle={Proceedings of the European Conference on Computer Vision},
  pages={402--419},
  year={2020},
  organization={Springer}
}

@inproceedings{mayer2016large,
  title={A large dataset to train convolutional networks for disparity, optical flow, and scene flow estimation},
  author={Mayer, Nikolaus and Ilg, Eddy and Hausser, Philip and Fischer, Philipp and Cremers, Daniel and Dosovitskiy, Alexey and Brox, Thomas},
  booktitle={Proceedings of the IEEE/CVF Conference on Computer Vision and Pattern Recognition},
  pages={4040--4048},
  year={2016}
}

@article{heusel2017gans,
  title={Gans trained by a two time-scale update rule converge to a local nash equilibrium},
  author={Heusel, Martin and Ramsauer, Hubert and Unterthiner, Thomas and Nessler, Bernhard and Hochreiter, Sepp},
  journal={Advances in neural information processing systems},
  volume={30},
  year={2017}
}

@inproceedings{butler2012naturalistic,
  title={A naturalistic open source movie for optical flow evaluation},
  author={Butler, Daniel J and Wulff, Jonas and Stanley, Garrett B and Black, Michael J},
  booktitle={Proceedings of the European Conference on Computer Vision},
  pages={611--625},
  year={2012},
  organization={Springer}
}

@article{geiger2013vision,
  title={Vision meets robotics: The kitti dataset},
  author={Geiger, Andreas and Lenz, Philip and Stiller, Christoph and Urtasun, Raquel},
  journal={The International Journal of Robotics Research},
  volume={32},
  number={11},
  pages={1231--1237},
  year={2013},
  publisher={Sage Publications Sage UK: London, England}
}

@inproceedings{menze2015object,
  title={Object scene flow for autonomous vehicles},
  author={Menze, Moritz and Geiger, Andreas},
  booktitle={Proceedings of the IEEE/CVF Conference on Computer Vision and Pattern Recognition},
  pages={3061--3070},
  year={2015}
}

@inproceedings{huang2022flowformer,
  title={FlowFormer: A Transformer Architecture for Optical Flow},
  author={Huang, Zhaoyang and Shi, Xiaoyu and Zhang, Chao and Wang, Qiang and Cheung, Ka Chun and Qin, Hongwei and Dai, Jifeng and Li, Hongsheng},
  booktitle={Proceedings of the European Conference on Computer Vision},
  year={2022}
}

@inproceedings{meister2018unflow,
  title={Unflow: Unsupervised learning of optical flow with a bidirectional census loss},
  author={Meister, Simon and Hur, Junhwa and Roth, Stefan},
  booktitle={Proceedings of the AAAI conference on artificial intelligence},
  volume={32},
  number={1},
  year={2018}
}

@inproceedings{huang2022real,
  title={Real-time intermediate flow estimation for video frame interpolation},
  author={Huang, Zhewei and Zhang, Tianyuan and Heng, Wen and Shi, Boxin and Zhou, Shuchang},
  booktitle={European Conference on Computer Vision},
  pages={624--642},
  year={2022},
  organization={Springer}
}

@inproceedings{zhang2023extracting,
  title={Extracting motion and appearance via inter-frame attention for efficient video frame interpolation},
  author={Zhang, Guozhen and Zhu, Yuhan and Wang, Haonan and Chen, Youxin and Wu, Gangshan and Wang, Limin},
  booktitle={Proceedings of the IEEE/CVF Conference on Computer Vision and Pattern Recognition},
  pages={5682--5692},
  year={2023}
}

@inproceedings{liu2020enhanced,
  title={Enhanced quadratic video interpolation},
  author={Liu, Yihao and Xie, Liangbin and Siyao, Li and Sun, Wenxiu and Qiao, Yu and Dong, Chao},
  booktitle={European conference on computer vision},
  pages={41--56},
  year={2020},
  organization={Springer}
}

@inproceedings{shang2023joint,
  title={Joint video multi-frame interpolation and deblurring under unknown exposure time},
  author={Shang, Wei and Ren, Dongwei and Yang, Yi and Zhang, Hongzhi and Ma, Kede and Zuo, Wangmeng},
  booktitle={Proceedings of the IEEE/CVF Conference on Computer Vision and Pattern Recognition},
  pages={13935--13944},
  year={2023}
}

@article{han2022realflow,
  title={RealFlow: EM-based Realistic Optical Flow Dataset Generation from Videos},
  author={Han, Yunhui and Luo, Kunming and Luo, Ao and Liu, Jiangyu and Fan, Haoqiang and Luo, Guiming and Liu, Shuaicheng},
  journal={arXiv preprint arXiv:2207.11075},
  year={2022}
}

@inproceedings{kong2022ifrnet,
  title={Ifrnet: Intermediate feature refine network for efficient frame interpolation},
  author={Kong, Lingtong and Jiang, Boyuan and Luo, Donghao and Chu, Wenqing and Huang, Xiaoming and Tai, Ying and Wang, Chengjie and Yang, Jie},
  booktitle={Proceedings of the IEEE/CVF Conference on Computer Vision and Pattern Recognition},
  pages={1969--1978},
  year={2022}
}

@article{xu2019quadratic,
  title={Quadratic video interpolation},
  author={Xu, Xiangyu and Siyao, Li and Sun, Wenxiu and Yin, Qian and Yang, Ming-Hsuan},
  journal={Advances in Neural Information Processing Systems},
  volume={32},
  year={2019}
}

@inproceedings{li2023amt,
  title={AMT: All-Pairs Multi-Field Transforms for Efficient Frame Interpolation},
  author={Li, Zhen and Zhu, Zuo-Liang and Han, Ling-Hao and Hou, Qibin and Guo, Chun-Le and Cheng, Ming-Ming},
  booktitle={Proceedings of the IEEE/CVF Conference on Computer Vision and Pattern Recognition},
  pages={9801--9810},
  year={2023}
}

@inproceedings{niklaus2020softmax,
  title={Softmax splatting for video frame interpolation},
  author={Niklaus, Simon and Liu, Feng},
  booktitle={Proceedings of the IEEE/CVF Conference on Computer Vision and Pattern Recognition},
  pages={5437--5446},
  year={2020}
}

@article{pont20172017,
  title={The 2017 davis challenge on video object segmentation},
  author={Pont-Tuset, Jordi and Perazzi, Federico and Caelles, Sergi and Arbel{\'a}ez, Pablo and Sorkine-Hornung, Alex and Van Gool, Luc},
  journal={arXiv preprint arXiv:1704.00675},
  year={2017}
}

@article{blattmann2023stable,
  title={Stable video diffusion: Scaling latent video diffusion models to large datasets},
  author={Blattmann, Andreas and Dockhorn, Tim and Kulal, Sumith and Mendelevitch, Daniel and Kilian, Maciej and Lorenz, Dominik and Levi, Yam and English, Zion and Voleti, Vikram and Letts, Adam and others},
  journal={arXiv preprint arXiv:2311.15127},
  year={2023}
}

@article{zhang2025motion,
  title={Motion-aware generative frame interpolation},
  author={Zhang, Guozhen and Zhu, Yuhan and Cui, Yutao and Zhao, Xiaotong and Ma, Kai and Wang, Limin},
  journal={arXiv preprint arXiv:2501.03699},
  year={2025}
}

@inproceedings{jain2024video,
  title={Video interpolation with diffusion models},
  author={Jain, Siddhant and Watson, Daniel and Tabellion, Eric and Poole, Ben and Kontkanen, Janne and others},
  booktitle={Proceedings of the IEEE/CVF Conference on Computer Vision and Pattern Recognition},
  pages={7341--7351},
  year={2024}
}

@inproceedings{danier2024ldmvfi,
  title={Ldmvfi: Video frame interpolation with latent diffusion models},
  author={Danier, Duolikun and Zhang, Fan and Bull, David},
  booktitle={Proceedings of the AAAI Conference on Artificial Intelligence},
  volume={38},
  number={2},
  pages={1472--1480},
  year={2024}
}

@article{xue2019video,
  title={Video enhancement with task-oriented flow},
  author={Xue, Tianfan and Chen, Baian and Wu, Jiajun and Wei, Donglai and Freeman, William T},
  journal={International Journal of Computer Vision},
  volume={127},
  pages={1106--1125},
  year={2019},
  publisher={Springer}
}

@inproceedings{lu2022video,
  title={Video frame interpolation with transformer},
  author={Lu, Liying and Wu, Ruizheng and Lin, Huaijia and Lu, Jiangbo and Jia, Jiaya},
  booktitle={Proceedings of the IEEE/CVF Conference on Computer Vision and Pattern Recognition},
  pages={3532--3542},
  year={2022}
}

@article{wang2004image,
  title={Image quality assessment: from error visibility to structural similarity},
  author={Wang, Zhou and Bovik, Alan C and Sheikh, Hamid R and Simoncelli, Eero P},
  journal={IEEE transactions on image processing},
  volume={13},
  number={4},
  pages={600--612},
  year={2004},
  publisher={IEEE}
}

@inproceedings{zhang2018unreasonable,
  title={The unreasonable effectiveness of deep features as a perceptual metric},
  author={Zhang, Richard and Isola, Phillip and Efros, Alexei A and Shechtman, Eli and Wang, Oliver},
  booktitle={Proceedings of the IEEE conference on computer vision and pattern recognition},
  pages={586--595},
  year={2018}
}

@inproceedings{jeong2024ocai,
  title={Ocai: Improving optical flow estimation by occlusion and consistency aware interpolation},
  author={Jeong, Jisoo and Cai, Hong and Garrepalli, Risheek and Lin, Jamie Menjay and Hayat, Munawar and Porikli, Fatih},
  booktitle={Proceedings of the IEEE/CVF Conference on Computer Vision and Pattern Recognition},
  pages={19352--19362},
  year={2024}
}

@article{wang2024generative,
  title={Generative inbetweening: Adapting image-to-video models for keyframe interpolation},
  author={Wang, Xiaojuan and Zhou, Boyang and Curless, Brian and Kemelmacher-Shlizerman, Ira and Holynski, Aleksander and Seitz, Steven M},
  journal={arXiv preprint arXiv:2408.15239},
  year={2024}
}

@inproceedings{feng2024explorative,
  title={Explorative inbetweening of time and space},
  author={Feng, Haiwen and Ding, Zheng and Xia, Zhihao and Niklaus, Simon and Abrevaya, Victoria and Black, Michael J and Zhang, Xuaner},
  booktitle={European Conference on Computer Vision},
  pages={378--395},
  year={2024},
  organization={Springer}
}

@inproceedings{seo2025bim,
  title={BiM-VFI: Bidirectional Motion Field-Guided Frame Interpolation for Video with Non-uniform Motions},
  author={Seo, Wonyong and Oh, Jihyong and Kim, Munchurl},
  booktitle={Proceedings of the Computer Vision and Pattern Recognition Conference},
  pages={7244--7253},
  year={2025}
}

@inproceedings{xu2022gmflow,
  title={Gmflow: Learning optical flow via global matching},
  author={Xu, Haofei and Zhang, Jing and Cai, Jianfei and Rezatofighi, Hamid and Tao, Dacheng},
  booktitle={Proceedings of the IEEE/CVF conference on computer vision and pattern recognition},
  pages={8121--8130},
  year={2022}
}

\newpage

\section{Implementation Details}
\label{supple:implementation}

\subsection{Model implementation details}

We compared our proposed method against several state-of-the-art approaches. For fair evaluation, all competing methods were implemented using their official releases and publicly available pretrained weights\footnote{IFR-Net~\cite{kong2022ifrnet}: \url{https://github.com/ltkong218/IFRNet} \\ VFIFormer~\cite{lu2022video}: \url{https://github.com/dvlab-research/VFIformer} \\ AMT~\cite{li2023amt}: \url{https://github.com/MCG-NKU/AMT}\\ EMA-VFI~\cite{zhang2023extracting}: \url{https://github.com/MCG-NJU/EMA-VFI}\\ Bim-VFI~\cite{seo2025bim}: \url{https://github.com/KAIST-VICLab/BiM-VFI}\\ RIPR~\cite{han2022realflow}: \url{https://github.com/megvii-research/RealFlow}\\ LDMVFI~\cite{danier2024ldmvfi}: \url{https://github.com/danier97/LDMVFI}\\ TRF~\cite{feng2024explorative}: \url{https://time-reversal.github.io/} \\ GenIn~\cite{wang2024generative}: \url{https://github.com/jeanne-wang/svd_keyframe_interpolation} }. Specifically, IFRNet-B~\cite{kong2022ifrnet} was employed with its baseline model configuration, while the AMT-L~\cite{li2023amt} was evaluated. For BiM-VFI~\cite{seo2025bim}, the \textit{$pyr\_level$} parameter was set to 7 during experimentation.

Using our symmetric nonlinear motion-based flow interpolation method, we generated a total of 23 intermediate frames at temporal positions t=1/24, 2/24, 3/34, ..., 23/24. These synthesized intermediate frames were subsequently employed in the video diffusion process.

Unlike GenIn, which initializes from random Gaussian noise, our approach benefits from a strong initialization, enabling us to reduce the total number of diffusion steps from 50 (about 680 sec using A100 GPU) to 20. Furthermore, whereas GenIn performs noise reinjection five times up to 25 diffusion steps, our method applies noise reinjection once at each steps throughout the entire diffusion process.

\subsection{Dataset details}
We conducted our experiments using three benchmark datsets: DAVIS~\cite{pont20172017}, Sintel~\cite{butler2012naturalistic}, and KITTI~\cite{geiger2013vision, menze2015object}. 
Unlike approaches such as DIVIM~\cite{jain2024video}, which resize sequences to certain resolution, we preserved the original resolution of each dataset for both inter-frame generation and evaluation.


For DAVIS, we employed the 480p resolution sequences. In this datasets, the $12^{th}$ frame of each video was designated as the target frame for the $\times$2 setting, while the $11^{th}$, $12^{th}$, and $13^{th}$ frames were used as the target frames for the $\times$4 setting. Under the $\times$2 setting, the $11^{th}$ and $13^{th}$ frames were used as inputs, while multi-frame configuration, four frames (the $9^{th}$, $11^{th}$, $13^{th}$ and $15^{th}$) were employed. For the $\times$4 setting, the $10^{th}$ and $14^{th}$ frames were used, and in the multi-frame configuration, four frames (the $6^{th}$, $10^{th}$, $14^{th}$, and $18^{th}$) were selected. 

For Sintel, we adopted the clean version of the datasets, which excludes noise such as motion blur and fog. For KITTI dataset, we utilized the multi-view sequences. In both Sintel and KITTI, the $10^{th}$ frame was designated as the target frame for the $\times$2 setting, while the $9^{th}$, $10^{th}$, and $11^{th}$ frames were used as the target frames for the $\times$4 setting. In the $\times$2 setting, the $9^{th}$ and $11^{th}$ frames were used, while in the multi-frame configuration, four frames (the $7^{th}$, $9^{th}$, $11^{th}$, and $13^{th}$) were employed. In the $\times$4 setting, the $8^{th}$ and $12^{th}$ frames were used, and in the multi-frame configuration, four frames (the $4^{th}$, $8^{th}$, $12^{th}$, and $16^{th}$) were selected.

\section{Additional Qualitative Video Frame Interpolation results}
\label{supple:additional_results}

\begin{figure*}[ht]
\begin{center}$
\centering
\begin{tabular}{ c ccc}

\rotatebox{90}{\qquad \textbf{GT}}
 &  
\includegraphics[width=3.9cm,height=1.8cm]{Fig/supple_results/kitti_gt2.jpg} & 
\hspace{-0.15cm} 
\includegraphics[width=3.9cm,height=1.8cm]{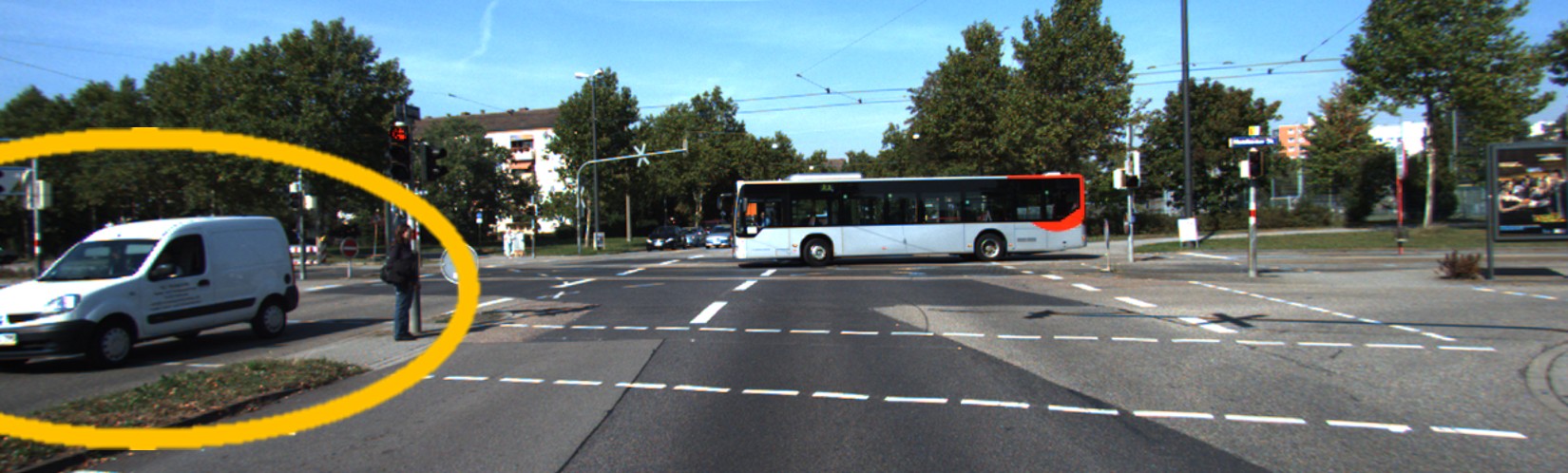} & 
\hspace{-0.15cm} 
\includegraphics[width=3.9cm,height=1.8cm]{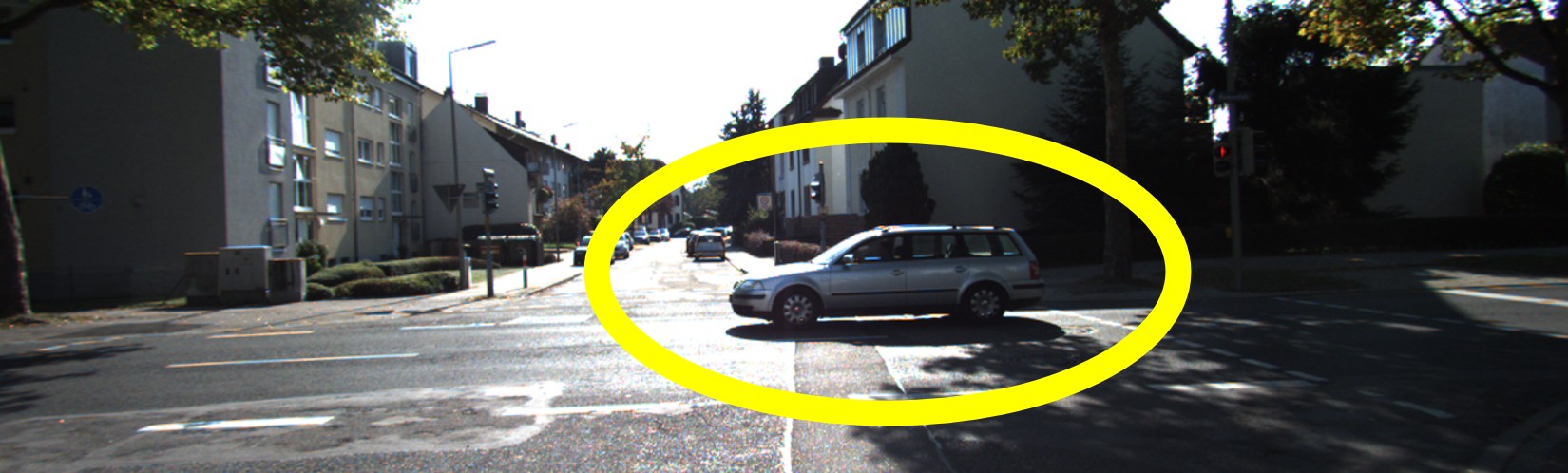} \\
\rotatebox{90}{\ \textbf{\scriptsize{VFIFormer}}}
&  
\includegraphics[width=3.9cm,height=1.8cm]{Fig/supple_results/kitti_vfiformer2.jpg} & \hspace{-0.15cm} 
\includegraphics[width=3.9cm,height=1.8cm]{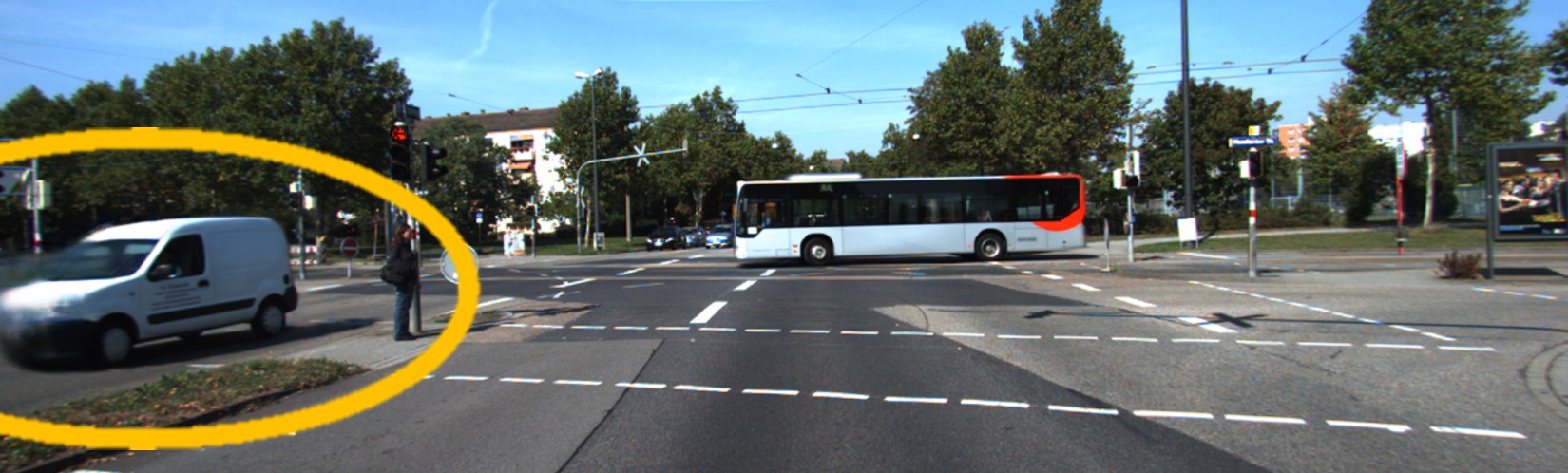} & \hspace{-0.15cm} 
\includegraphics[width=3.9cm,height=1.8cm]{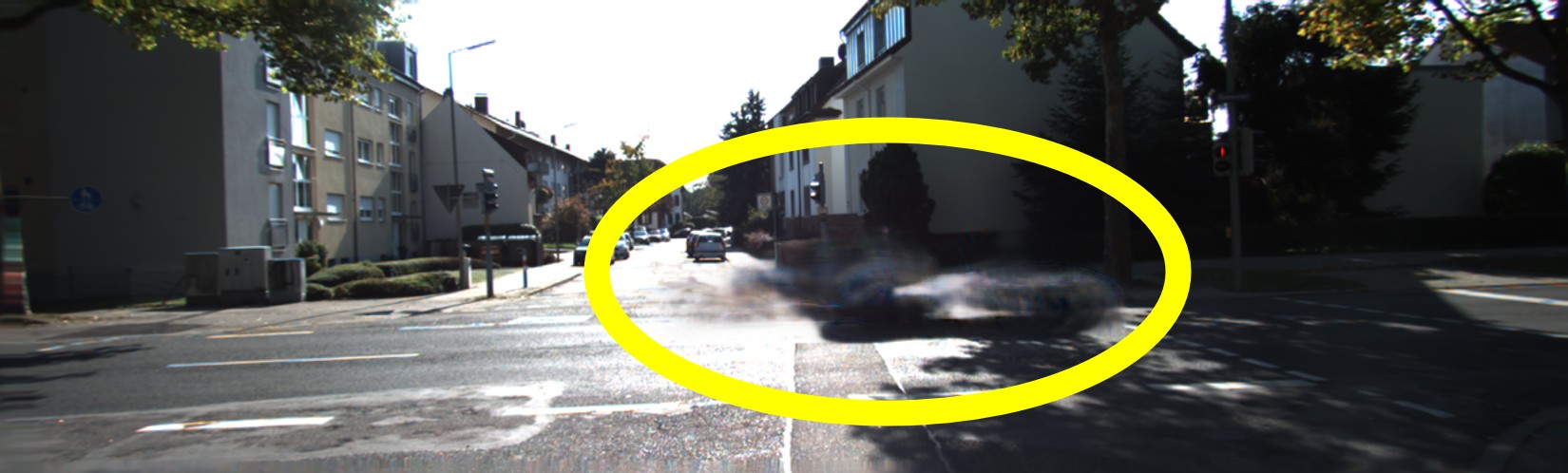} \\
\rotatebox{90}{ \ \textbf{\scriptsize{EMA-VFI}}}
& 
\includegraphics[width=3.9cm,height=1.8cm]{Fig/supple_results/kitti_ema2.jpg} & \hspace{-0.15cm} 
\includegraphics[width=3.9cm,height=1.8cm]{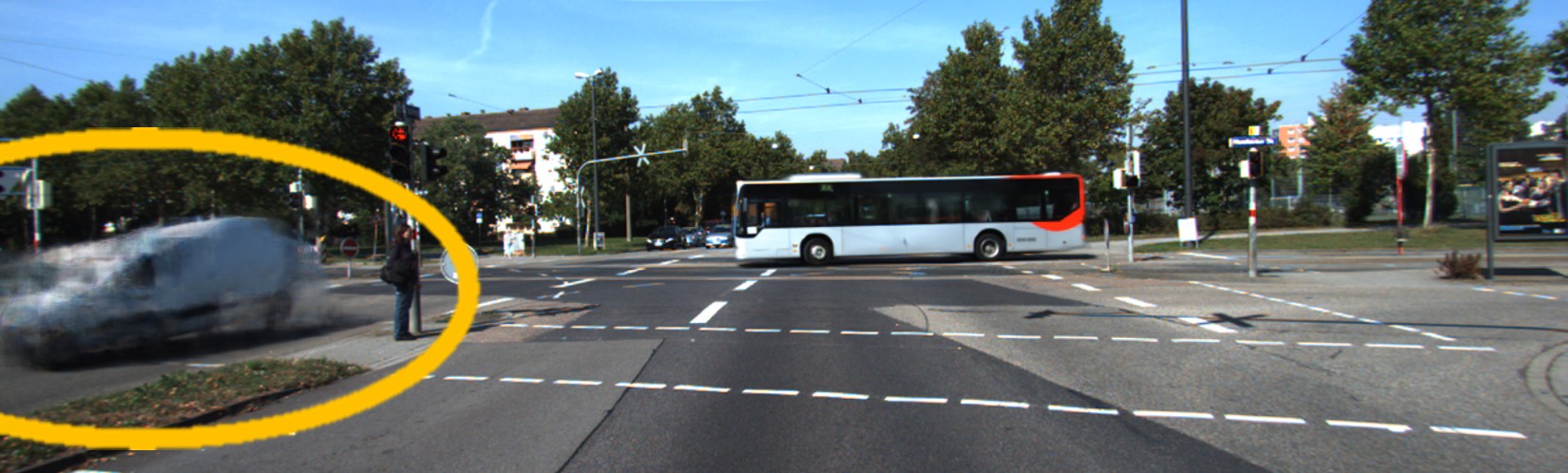} & \hspace{-0.15cm} 
\includegraphics[width=3.9cm,height=1.8cm]{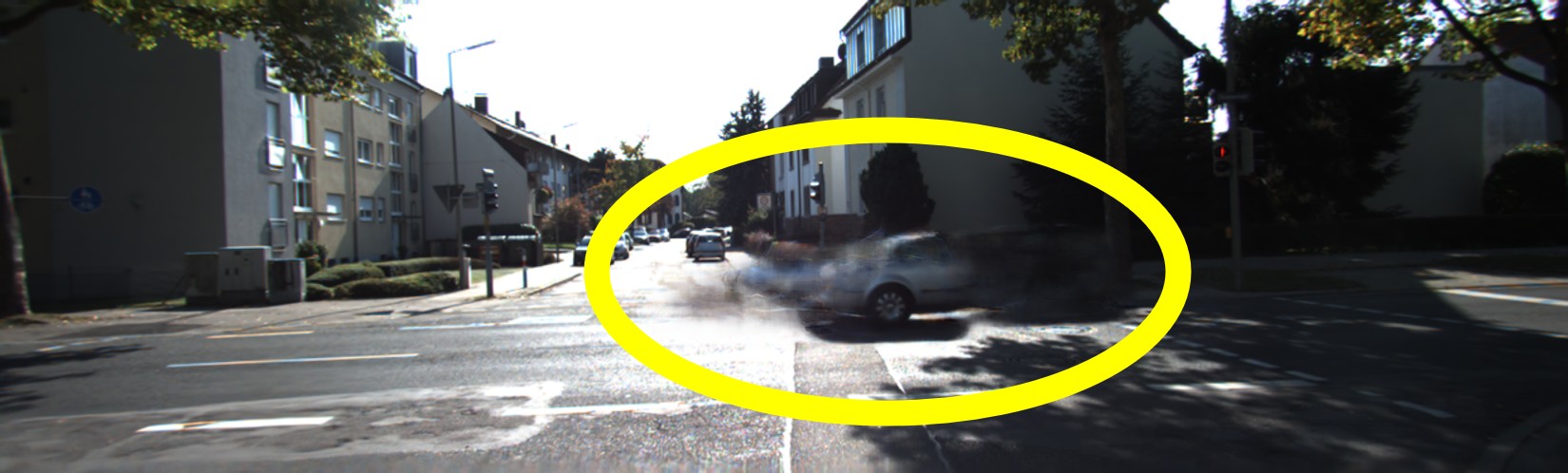} \\
 \rotatebox{90}{ \ \textbf{\scriptsize{BiM-VFI}}}
&  
\includegraphics[width=3.9cm,height=1.8cm]{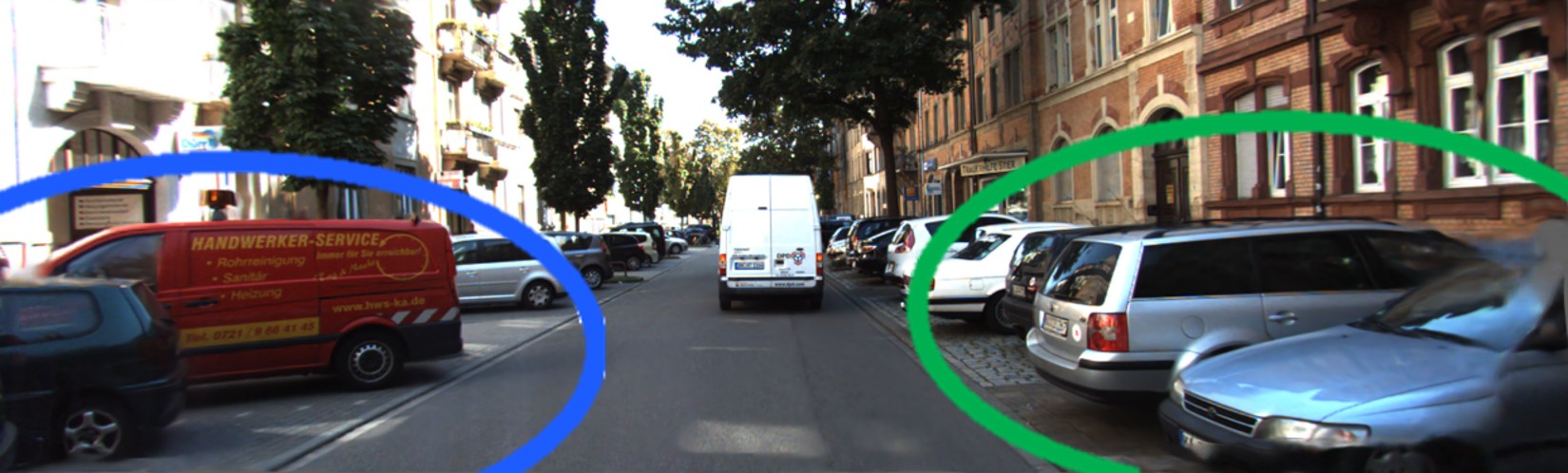} & \hspace{-0.15cm} 
\includegraphics[width=3.9cm,height=1.8cm]{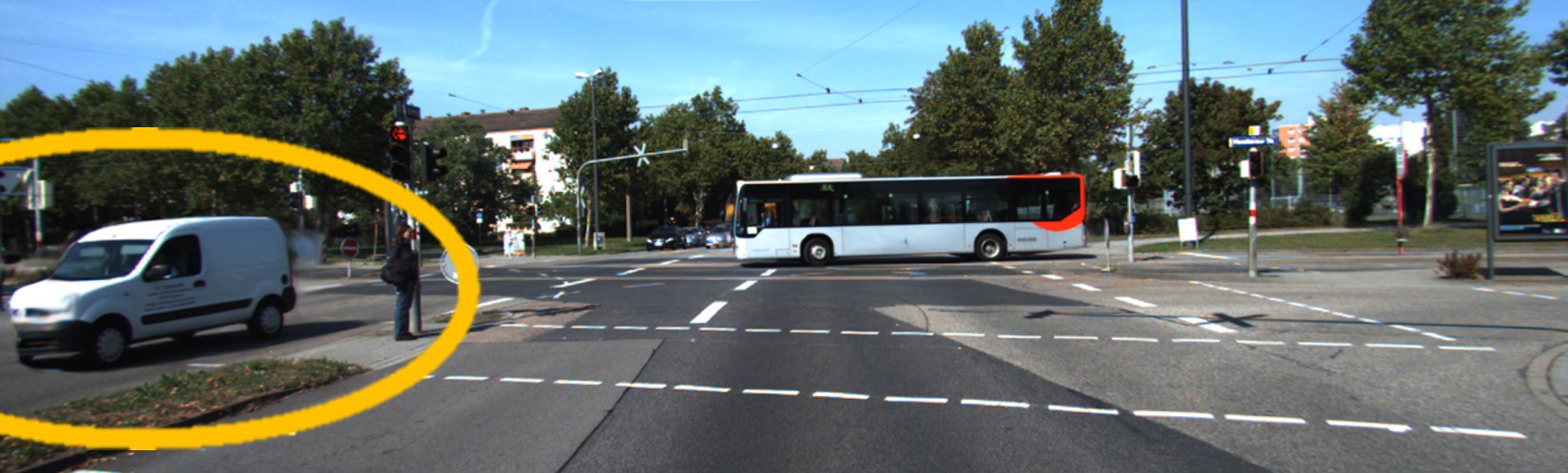} & \hspace{-0.15cm} 
\includegraphics[width=3.9cm,height=1.8cm]{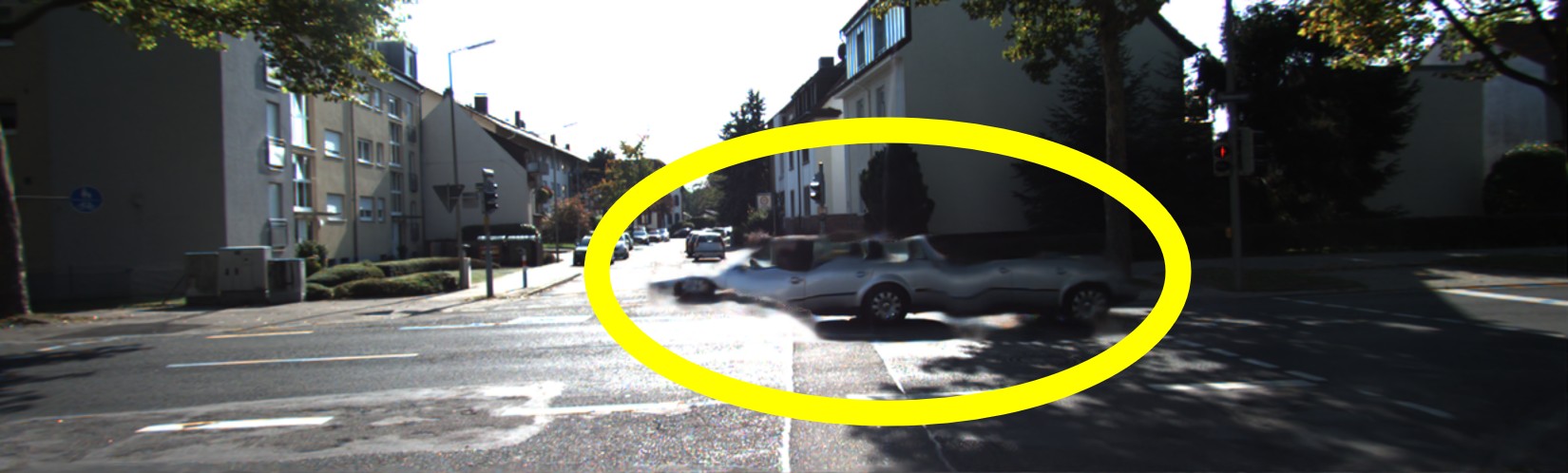} \\
 \rotatebox{90}{ \quad \textbf{\scriptsize{OCAI}}}
& 
\includegraphics[width=3.9cm,height=1.8cm]{Fig/supple_results/kitti_ocai2.jpg} & \hspace{-0.15cm}  
\includegraphics[width=3.9cm,height=1.8cm]{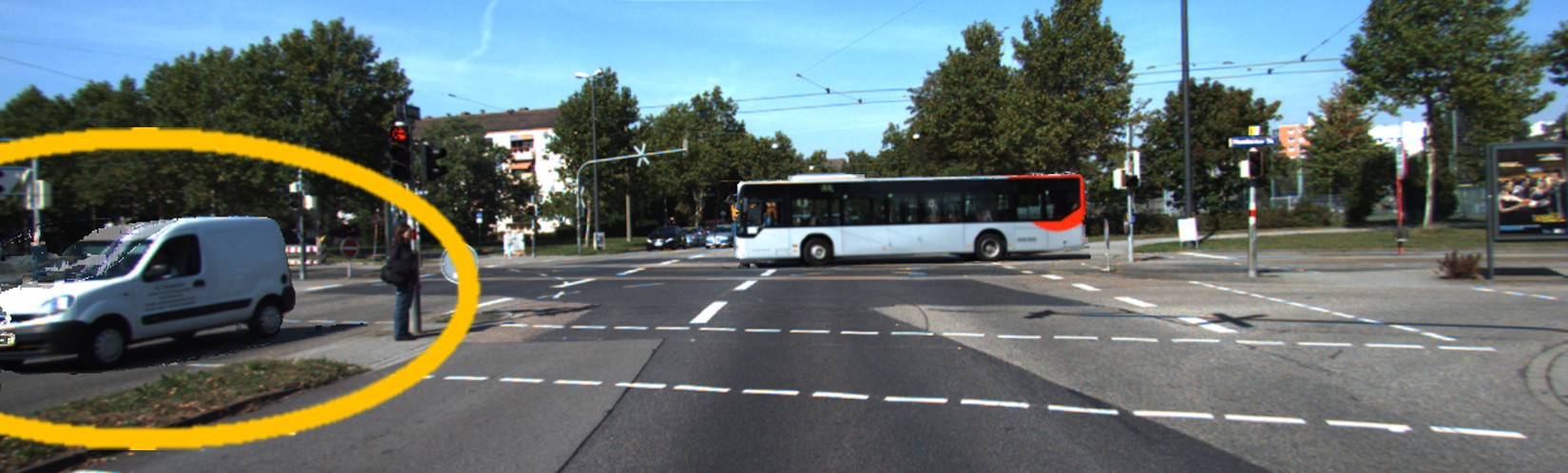} & \hspace{-0.15cm} 
\includegraphics[width=3.9cm,height=1.8cm]{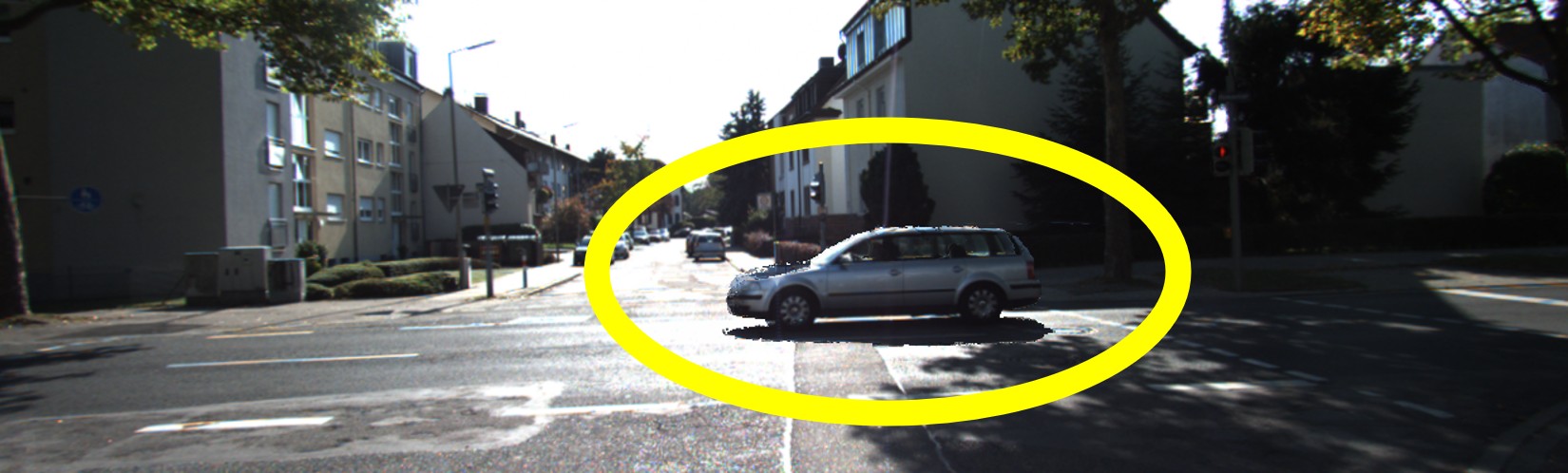} \\
 \rotatebox{90}{ \ \  \textbf{\scriptsize{LDMVFI}}}
&  
\includegraphics[width=3.9cm,height=1.8cm]{Fig/supple_results/kitti_ldmvfi2.jpg} & \hspace{-0.15cm} 
\includegraphics[width=3.9cm,height=1.8cm]{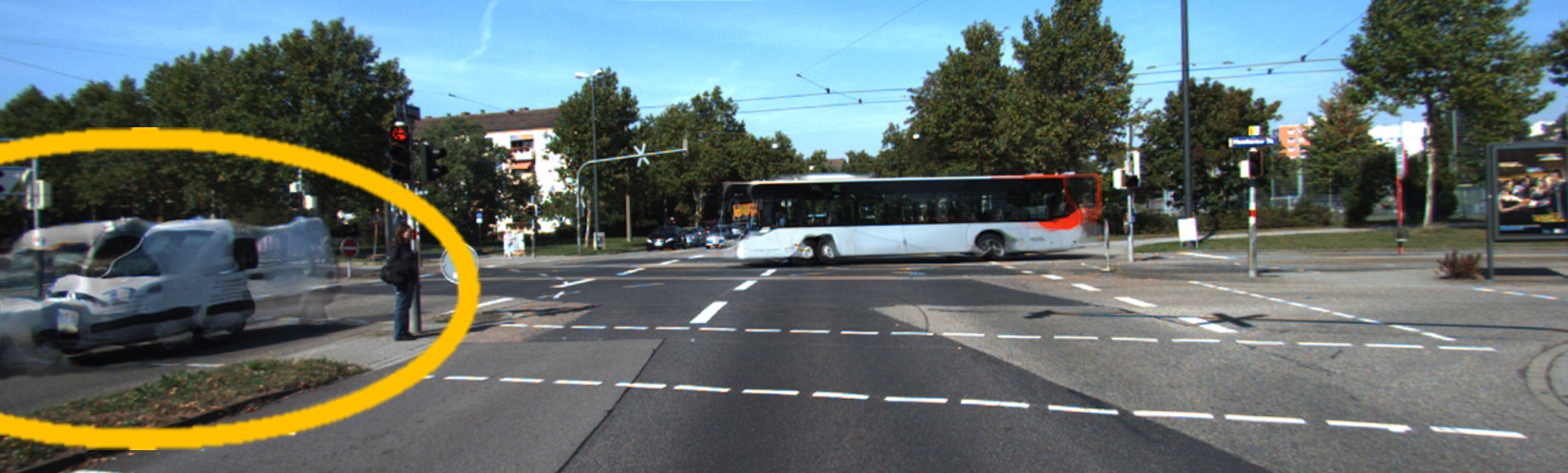} & \hspace{-0.15cm} 
\includegraphics[width=3.9cm,height=1.8cm]{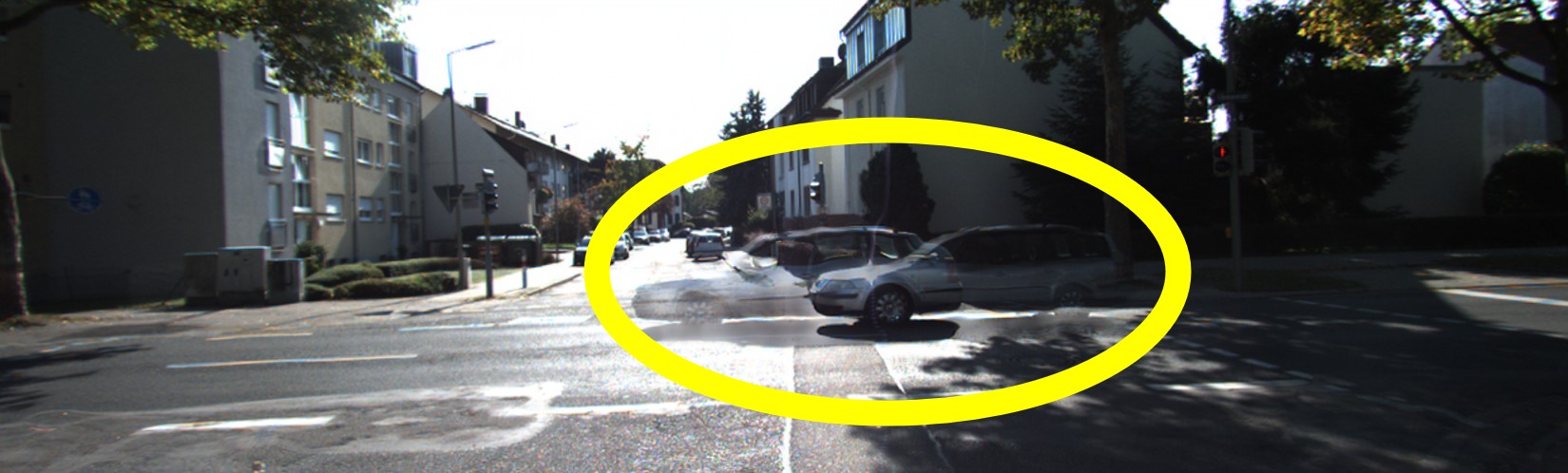} \\
 \rotatebox{90}{ \quad \ \textbf{\scriptsize{GenIn}}}
&  
\includegraphics[width=3.9cm,height=1.8cm]{Fig/supple_results/kitti_gen2.jpg} & \hspace{-0.15cm} 
\includegraphics[width=3.9cm,height=1.8cm]{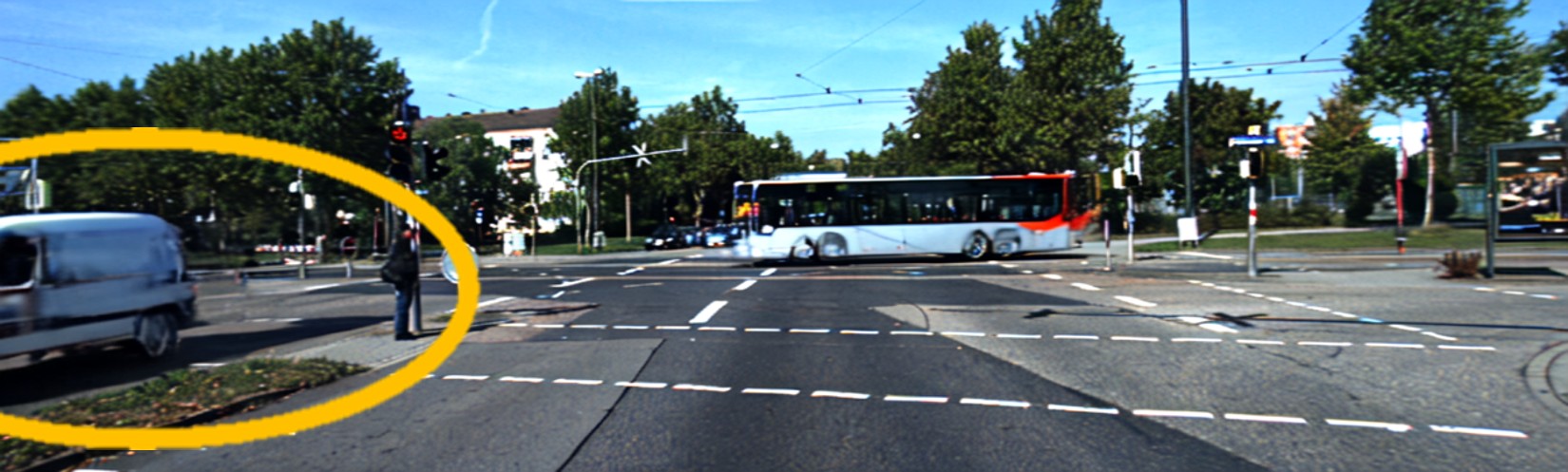} & \hspace{-0.15cm} 
\includegraphics[width=3.9cm,height=1.8cm]{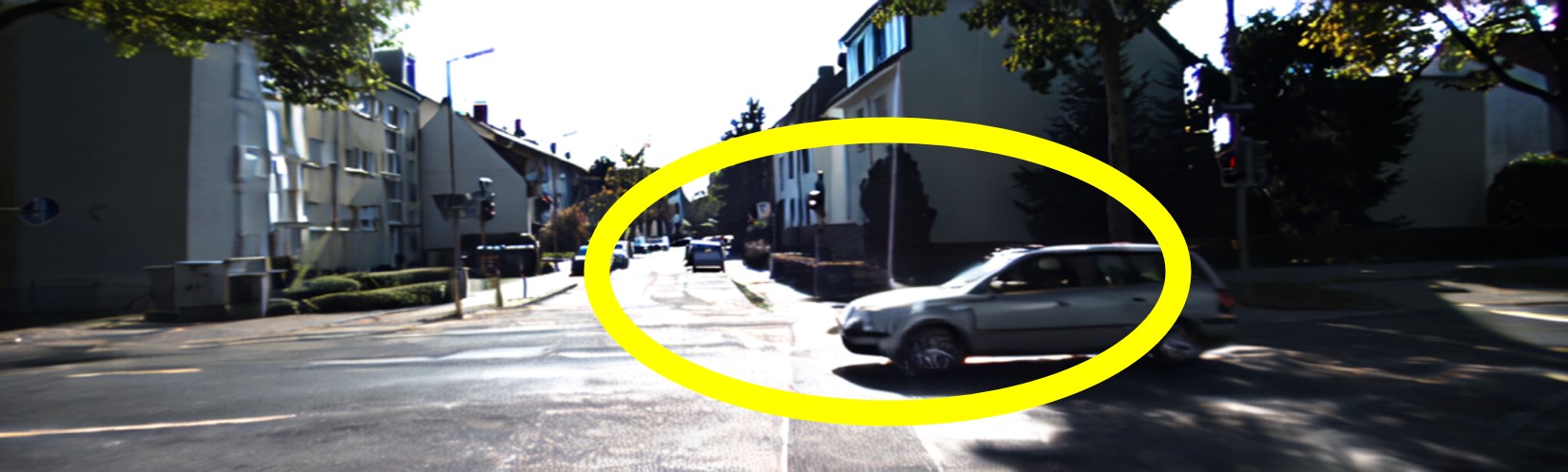} \\
 \rotatebox{90}{\quad \ \ \textbf{\scriptsize{Ours}}}
&  
\includegraphics[width=3.9cm,height=1.8cm]{Fig/supple_results/kitti_our2.jpg} & \hspace{-0.15cm} 
\includegraphics[width=3.9cm,height=1.8cm]{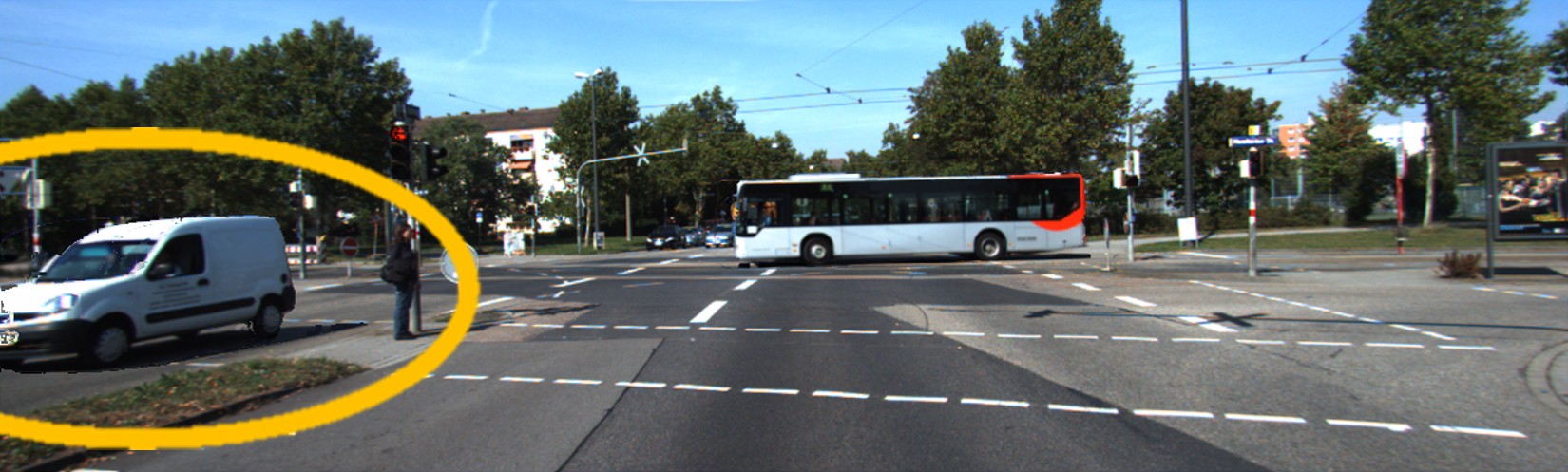} & \hspace{-0.15cm} 
\includegraphics[width=3.9cm,height=1.8cm]{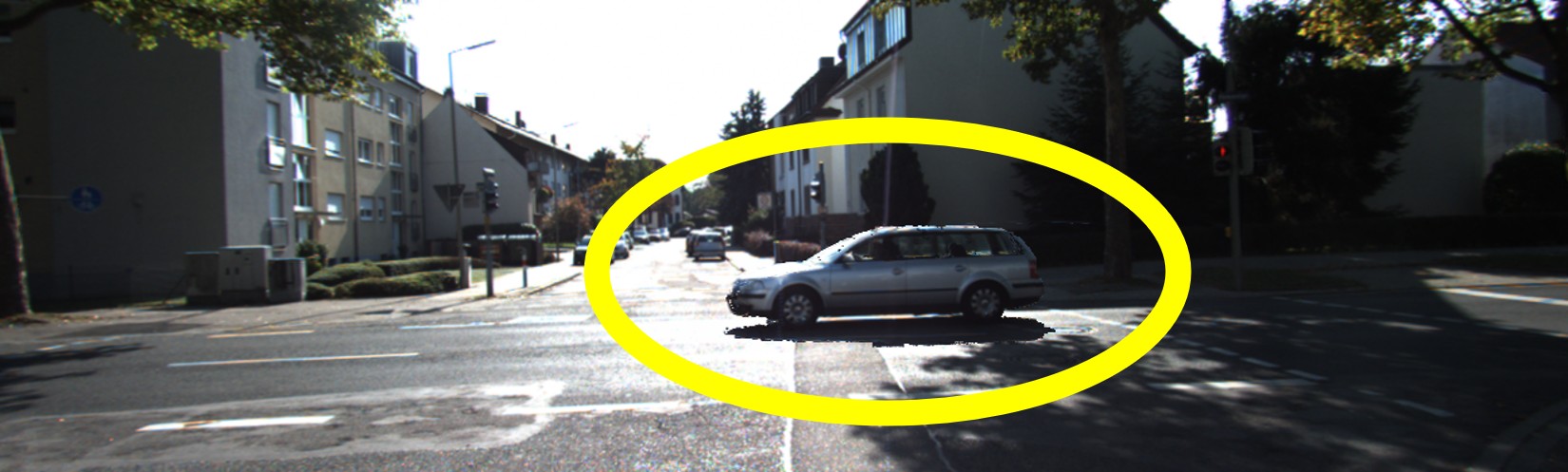} \\
\vspace{-18pt}
\end{tabular}$
\end{center}
\caption{Video frame interpolation results on KITTI ($\times$2). The second to fifth rows show results from flow-based methods: VFIFormer and EVA-VFI (backward warping) and OCAI (forward warping). The sixth and seventh rows present results from diffusion-based methods: LDMVFI and GenIn. The final row displays the output of our proposed \ours method. 
}
\vspace{-6pt}
\label{fig:experiment_vfi_kitti2x}
\end{figure*}

We provide additional qualitative results on the DAVIS, Sintel, and KITTI datasets (p5-9). Fig.~\ref{fig:experiment_vfi_kitti2x} illustrates the results for the KITTI dataset under the $\times$2 setting. As shown in the figure, particularly within the yellow circle on the right, most existing methods fail to handle large motion effectively, whereas OCAI and our proposed method successfully capture the motion. Moreover, in the right image, our method produces results with noticeably less noise. In the middle image, OCAI introduces significant noise, while our approach generates a much cleaner reconstruction.

\begin{figure*}[ht]
\begin{center}$
\centering
\begin{tabular}{ c ccc}

\rotatebox{90}{\qquad \textbf{GT}}
 &  
\includegraphics[width=3.9cm,height=1.8cm]{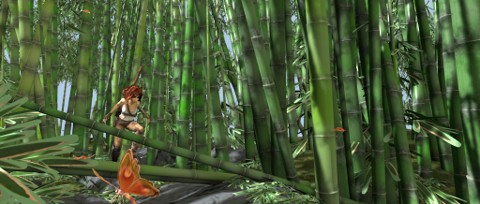} & \hspace{-0.15cm} 
\includegraphics[width=3.9cm,height=1.8cm]{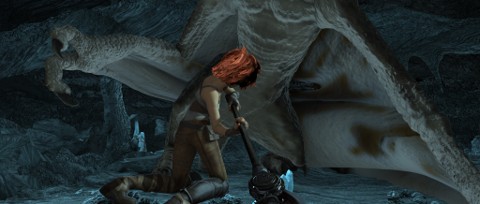} & \hspace{-0.15cm} 
\includegraphics[width=3.9cm,height=1.8cm]{Fig/supple_results/sintel_gt3.jpg} \\
\rotatebox{90}{\ \textbf{\scriptsize{VFIFormer}}}
& 
\includegraphics[width=3.9cm,height=1.8cm]{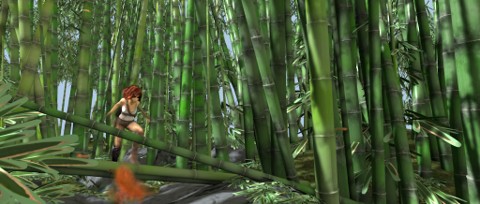} & \hspace{-0.15cm} 
\includegraphics[width=3.9cm,height=1.8cm]{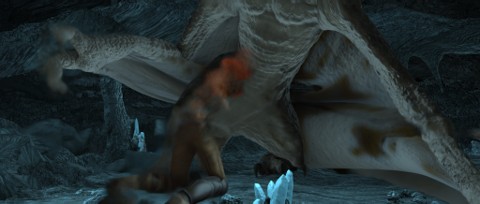} & \hspace{-0.15cm} 
\includegraphics[width=3.9cm,height=1.8cm]{Fig/supple_results/sintel_vfiformer3.jpg} \\
\rotatebox{90}{ \ \textbf{\scriptsize{EMA-VFI}}}
&  
\includegraphics[width=3.9cm,height=1.8cm]{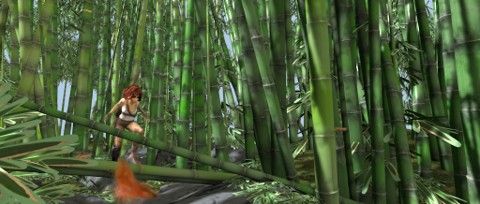} & \hspace{-0.15cm} 
\includegraphics[width=3.9cm,height=1.8cm]{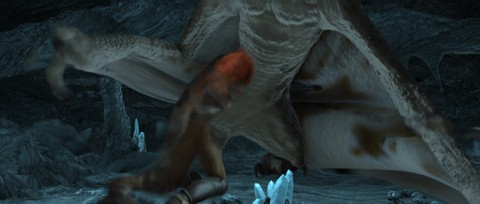} & \hspace{-0.15cm} 
\includegraphics[width=3.9cm,height=1.8cm]{Fig/supple_results/sintel_ema3.jpg} \\
 \rotatebox{90}{ \ \textbf{\scriptsize{BiM-VFI}}}
& 
\includegraphics[width=3.9cm,height=1.8cm]{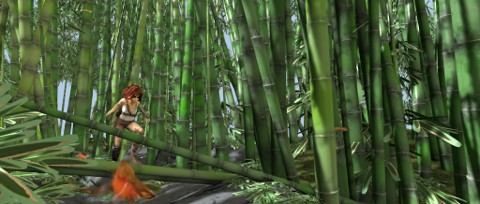} & \hspace{-0.15cm} 
\includegraphics[width=3.9cm,height=1.8cm]{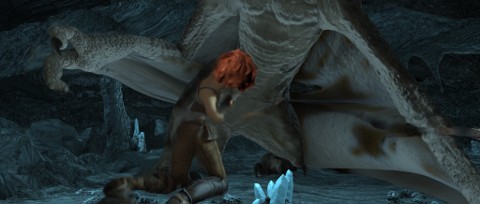} & \hspace{-0.15cm} 
\includegraphics[width=3.9cm,height=1.8cm]{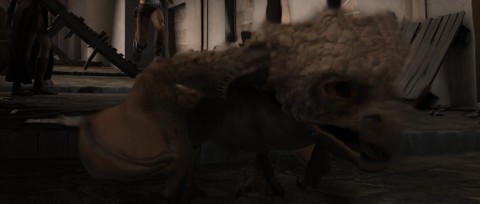} \\
 \rotatebox{90}{ \quad \textbf{\scriptsize{OCAI}}}
& 
\includegraphics[width=3.9cm,height=1.8cm]{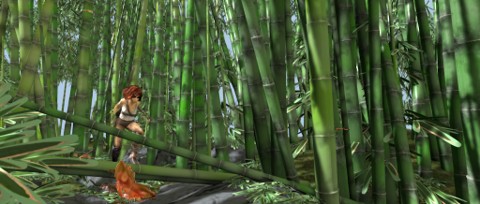} & \hspace{-0.15cm} 
\includegraphics[width=3.9cm,height=1.8cm]{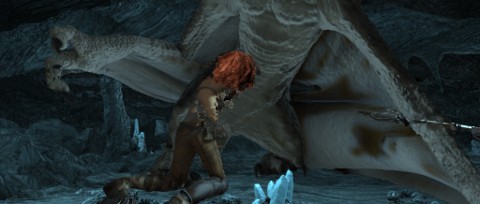} & \hspace{-0.15cm} 
\includegraphics[width=3.9cm,height=1.8cm]{Fig/supple_results/sintel_ocai3.jpg} \\
 \rotatebox{90}{ \ \  \textbf{\scriptsize{LDMVFI}}}
&  
\includegraphics[width=3.9cm,height=1.8cm]{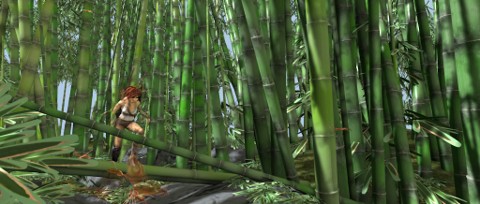} & \hspace{-0.15cm} 
\includegraphics[width=3.9cm,height=1.8cm]{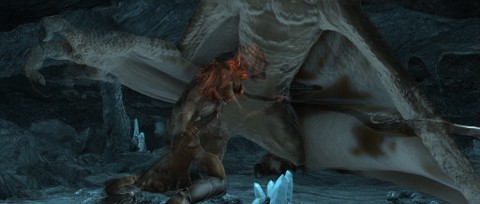} & \hspace{-0.15cm} 
\includegraphics[width=3.9cm,height=1.8cm]{Fig/supple_results/sintel_ldmvfi3.jpg} \\
 \rotatebox{90}{ \quad \ \textbf{\scriptsize{GenIn}}}
&  
\includegraphics[width=3.9cm,height=1.8cm]{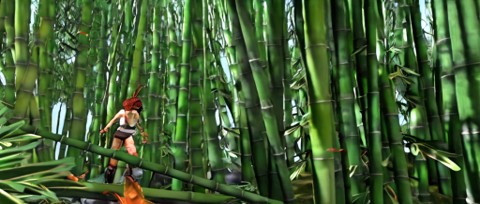} & \hspace{-0.15cm} 
\includegraphics[width=3.9cm,height=1.8cm]{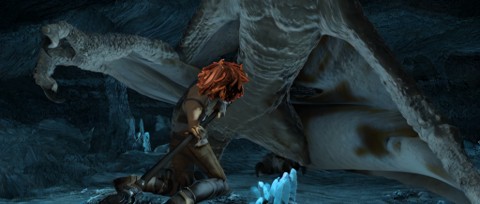} & \hspace{-0.15cm} 
\includegraphics[width=3.9cm,height=1.8cm]{Fig/supple_results/sintel_gen3.jpg} \\
 \rotatebox{90}{\quad \ \ \textbf{\scriptsize{Ours}}}
& 
\includegraphics[width=3.9cm,height=1.8cm]{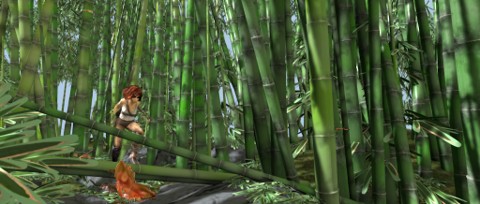} & \hspace{-0.15cm} 
\includegraphics[width=3.9cm,height=1.8cm]{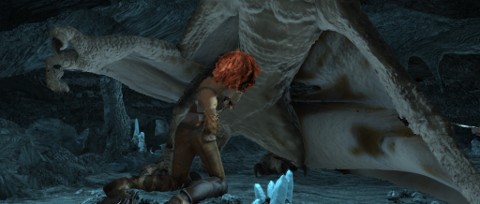} & \hspace{-0.15cm} 
\includegraphics[width=3.9cm,height=1.8cm]{Fig/supple_results/sintel_our3.jpg} \\
\vspace{-18pt}
\end{tabular}$
\end{center}
\caption{Video frame interpolation results on Sintel ($\times$2). The second to fifth rows show results from flow-based methods: VFIFormer and EVA-VFI (backward warping) and OCAI (forward warping). The sixth and seventh rows present results from diffusion-based methods: LDMVFI and GenIn. The final row displays the output of our proposed \ours method. 
}
\vspace{-6pt}
\label{fig:experiment_vfi_sintel2x}
\end{figure*}

\begin{figure*}[ht]
\begin{center}$
\centering
\begin{tabular}{ c ccc}

\rotatebox{90}{\qquad \textbf{GT}}
 & 
\includegraphics[width=3.9cm,height=1.8cm]{Fig/supple_results/davis_gt1.jpg} & \hspace{-0.15cm} 
\includegraphics[width=3.9cm,height=1.8cm]{Fig/supple_results/davis_gt2.jpg} & \hspace{-0.15cm} 
\includegraphics[width=3.9cm,height=1.8cm]{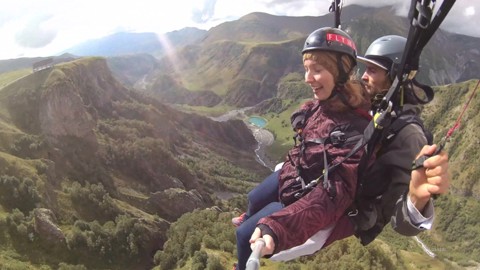} \\
\rotatebox{90}{\ \textbf{\scriptsize{VFIFormer}}}
&  
\includegraphics[width=3.9cm,height=1.8cm]{Fig/supple_results/davis_vfiformer1.jpg} & \hspace{-0.15cm} 
\includegraphics[width=3.9cm,height=1.8cm]{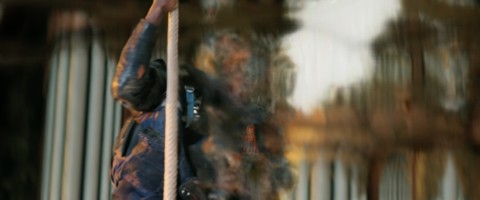} & \hspace{-0.15cm} 
\includegraphics[width=3.9cm,height=1.8cm]{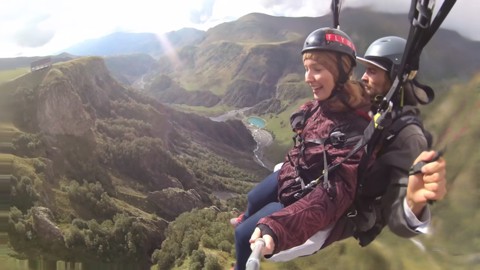} \\
\rotatebox{90}{ \ \textbf{\scriptsize{EMA-VFI}}}
&  
\includegraphics[width=3.9cm,height=1.8cm]{Fig/supple_results/davis_ema1.jpg} & \hspace{-0.15cm}  
\includegraphics[width=3.9cm,height=1.8cm]{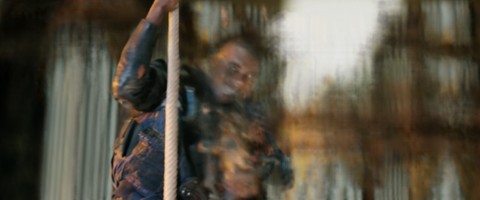} & \hspace{-0.15cm} 
\includegraphics[width=3.9cm,height=1.8cm]{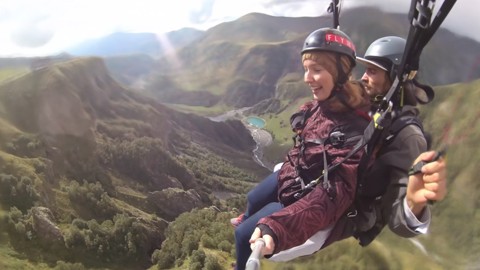} \\
 \rotatebox{90}{ \ \textbf{\scriptsize{BiM-VFI}}}
&  
\includegraphics[width=3.9cm,height=1.8cm]{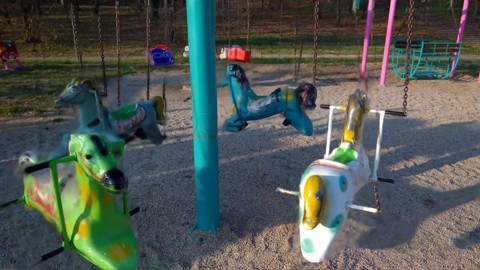} & \hspace{-0.15cm} 
\includegraphics[width=3.9cm,height=1.8cm]{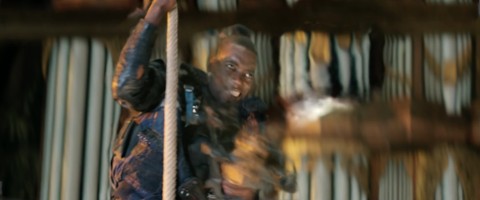} & \hspace{-0.15cm} 
\includegraphics[width=3.9cm,height=1.8cm]{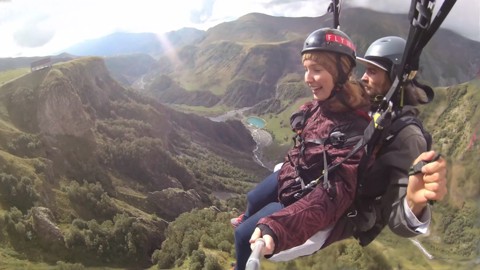} \\
 \rotatebox{90}{ \quad \textbf{\scriptsize{OCAI}}}
& 
\includegraphics[width=3.9cm,height=1.8cm]{Fig/supple_results/davis_ocai1.jpg} & \hspace{-0.15cm} 
\includegraphics[width=3.9cm,height=1.8cm]{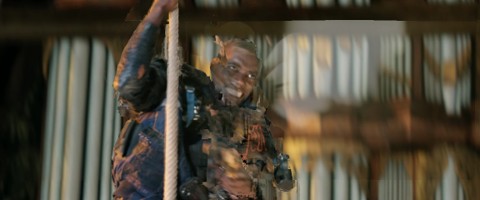} & \hspace{-0.15cm} 
\includegraphics[width=3.9cm,height=1.8cm]{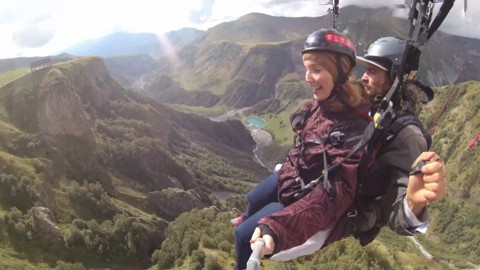} \\
 \rotatebox{90}{ \ \  \textbf{\scriptsize{LDMVFI}}}
&  
\includegraphics[width=3.9cm,height=1.8cm]{Fig/supple_results/davis_lvdvfi1.jpg} & \hspace{-0.15cm} 
\includegraphics[width=3.9cm,height=1.8cm]{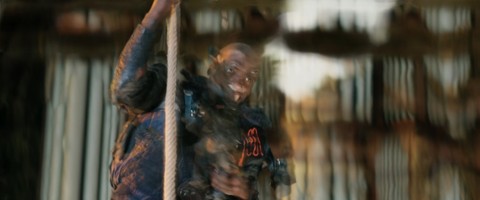} & \hspace{-0.15cm} 
\includegraphics[width=3.9cm,height=1.8cm]{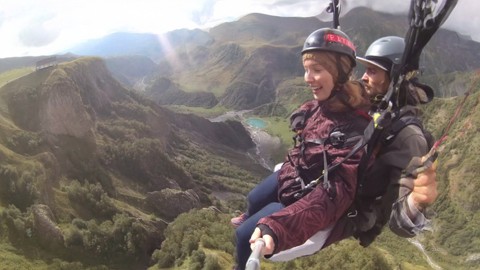} \\
\rotatebox{90}{ \quad \ \textbf{\scriptsize{GenIn}}}
& 
\includegraphics[width=3.9cm,height=1.8cm]{Fig/supple_results/davis_gen1.jpg} & \hspace{-0.15cm} 
\includegraphics[width=3.9cm,height=1.8cm]{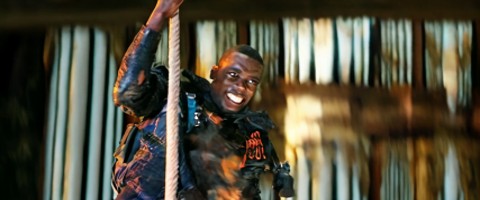} & \hspace{-0.15cm} 
\includegraphics[width=3.9cm,height=1.8cm]{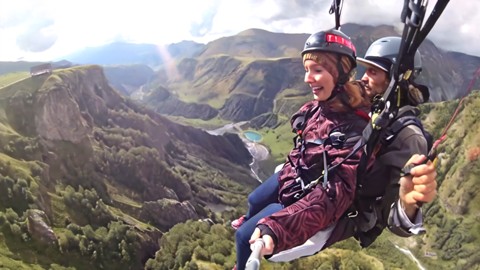} \\
\rotatebox{90}{\quad \ \ \textbf{\scriptsize{Ours}}}
&  
\includegraphics[width=3.9cm,height=1.8cm]{Fig/supple_results/davis_our1.jpg} & \hspace{-0.15cm} 
\includegraphics[width=3.9cm,height=1.8cm]{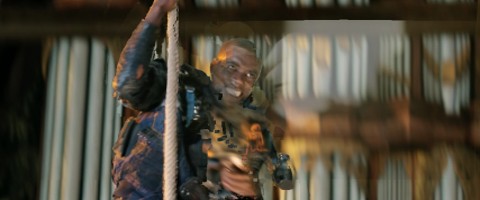} & \hspace{-0.15cm} 
\includegraphics[width=3.9cm,height=1.8cm]{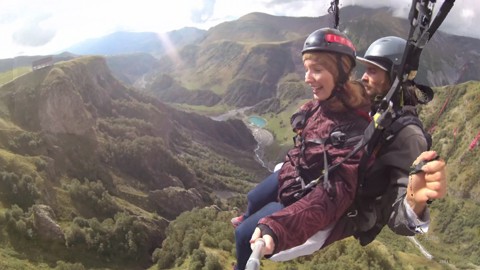} \\
\vspace{-18pt}
\end{tabular}$
\end{center}
\caption{Video frame interpolation results on DAVIS ($\times$2). The second to fifth rows show results from flow-based methods: VFIFormer and EVA-VFI (backward warping) and OCAI (forward warping). The sixth and seventh rows present results from diffusion-based methods: LDMVFI and GenIn. The final row displays the output of our proposed \ours method. 
}
\vspace{-6pt}
\label{fig:experiment_vfi_davis2x}
\end{figure*}

\begin{figure*}[ht]
\begin{center}$
\centering
\begin{tabular}{ c ccc}
& $\hat{I}_{1/4}$ & $\hat{I}_{2/4}$ &  $\hat{I}_{3/4}$ \\

\rotatebox{90}{\qquad \textbf{GT}}
 &  
\includegraphics[width=3.9cm,height=1.8cm]{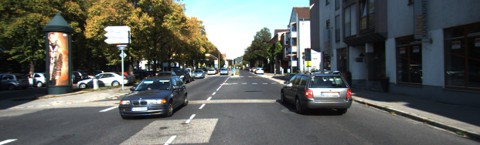} & \hspace{-0.15cm}
\includegraphics[width=3.9cm,height=1.8cm]{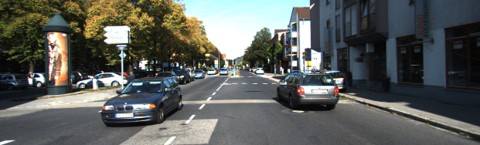} & \hspace{-0.15cm}
\includegraphics[width=3.9cm,height=1.8cm]{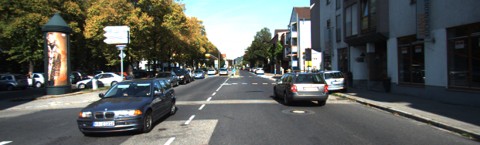} \\
\rotatebox{90}{ \ \textbf{\scriptsize{EMA-VFI}}}
& 
\includegraphics[width=3.9cm,height=1.8cm]{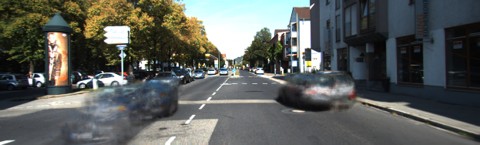} & \hspace{-0.15cm}
\includegraphics[width=3.9cm,height=1.8cm]{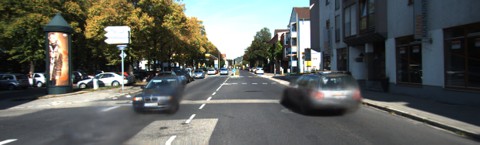} & \hspace{-0.15cm}
\includegraphics[width=3.9cm,height=1.8cm]{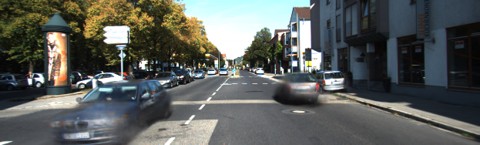} \\
 \rotatebox{90}{ \ \textbf{\scriptsize{BiM-VFI}}}
&  
\includegraphics[width=3.9cm,height=1.8cm]{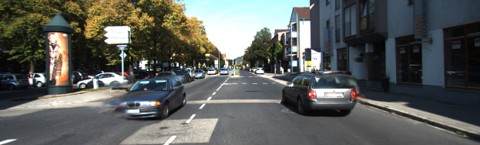} & \hspace{-0.15cm}
\includegraphics[width=3.9cm,height=1.8cm]{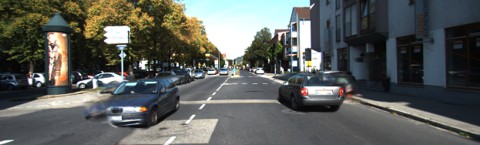} & \hspace{-0.15cm}
\includegraphics[width=3.9cm,height=1.8cm]{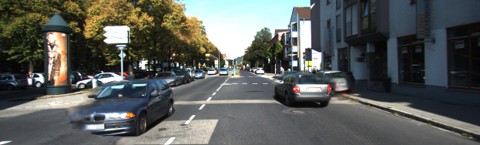} \\
 \rotatebox{90}{ \quad \textbf{\scriptsize{OCAI}}}
& 
\includegraphics[width=3.9cm,height=1.8cm]{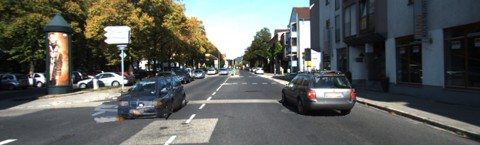} & \hspace{-0.15cm}
\includegraphics[width=3.9cm,height=1.8cm]{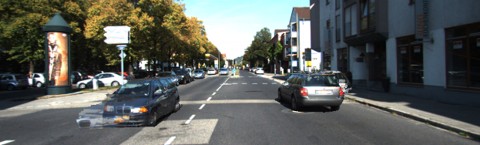} & \hspace{-0.15cm}
\includegraphics[width=3.9cm,height=1.8cm]{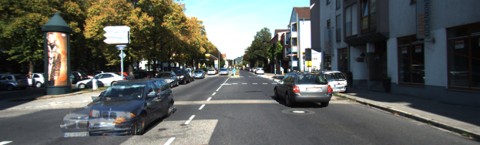} \\
\rotatebox{90}{ \ \  \textbf{\scriptsize{LDMVFI}}}
&  
\includegraphics[width=3.9cm,height=1.8cm]{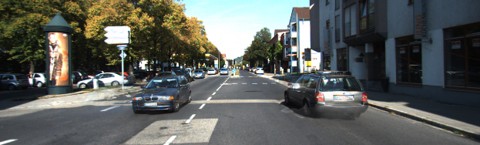} & \hspace{-0.15cm}
\includegraphics[width=3.9cm,height=1.8cm]{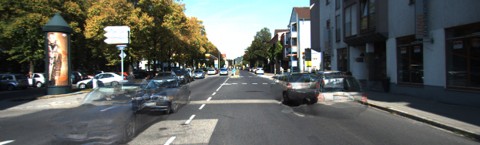} & \hspace{-0.15cm}
\includegraphics[width=3.9cm,height=1.8cm]{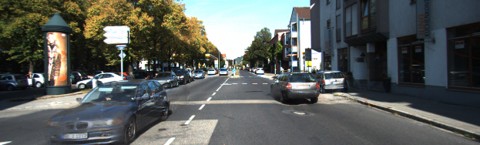} \\
\rotatebox{90}{ \quad \ \textbf{\scriptsize{GenIn}}}
&  
\includegraphics[width=3.9cm,height=1.8cm]{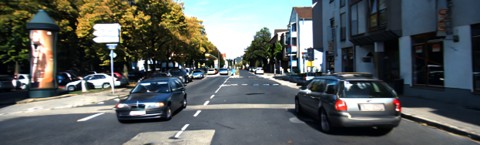} & \hspace{-0.15cm}
\includegraphics[width=3.9cm,height=1.8cm]{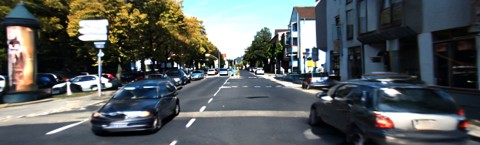} & \hspace{-0.15cm} 
\includegraphics[width=3.9cm,height=1.8cm]{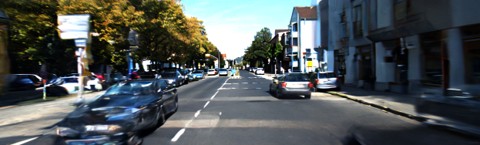} \\
\rotatebox{90}{ \ \textbf{\scriptsize{Quadratic}}}
& 
\includegraphics[width=3.9cm,height=1.8cm]{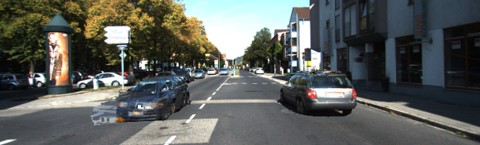} & \hspace{-0.15cm}
\includegraphics[width=3.9cm,height=1.8cm]{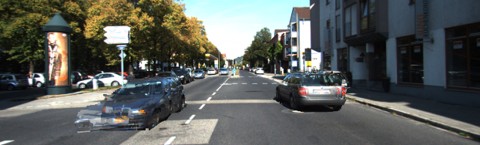} & \hspace{-0.15cm}
\includegraphics[width=3.9cm,height=1.8cm]{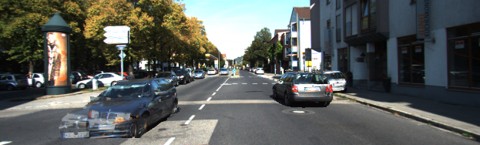} \\
\rotatebox{90}{\quad \ \ \textbf{\scriptsize{Ours}}}
& 
\includegraphics[width=3.9cm,height=1.8cm]{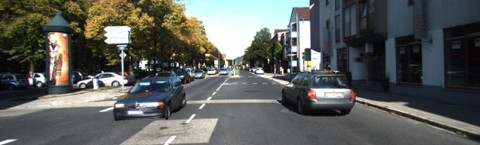} & \hspace{-0.15cm}
\includegraphics[width=3.9cm,height=1.8cm]{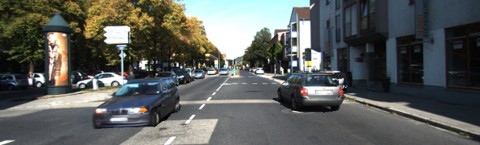} & \hspace{-0.15cm} 
\includegraphics[width=3.9cm,height=1.8cm]{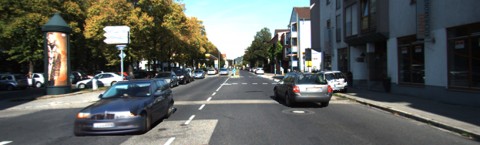} \\
\vspace{-18pt}
\end{tabular}$
\end{center}
\caption{Video frame interpolation results on KITTI ($\times$4). The second to fourth rows show results from flow-based methods: EVA-VFI and Bim-VFI (backward warping) and OCAI (forward warping). The fifth and sixth rows present results from diffusion-based methods: LDMVFI and GenIn. The seventh row show the quadratic motion VFI results. The final row displays the output of our proposed \ours method. 
}
\vspace{-6pt}
\label{fig:experiment_vfi_kitti4x}
\end{figure*}

\begin{figure*}[ht]
\begin{center}$
\centering
\begin{tabular}{ c ccc}
& $\hat{I}_{1/4}$ & $\hat{I}_{2/4}$ &  $\hat{I}_{3/4}$ \\
\rotatebox{90}{\qquad \textbf{GT}}
 &  
\includegraphics[width=3.9cm,height=1.8cm]{Fig/supple_4x_results/davis_gt1.jpg} & \hspace{-0.15cm} 
\includegraphics[width=3.9cm,height=1.8cm]{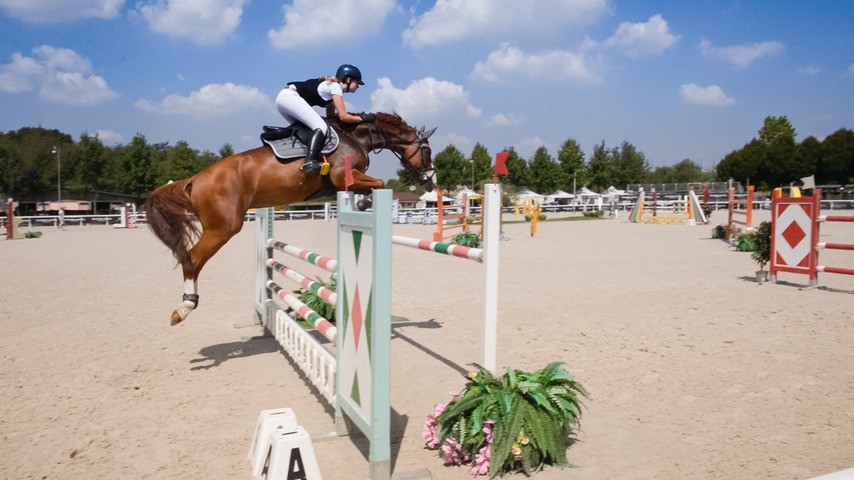} & \hspace{-0.15cm} 
\includegraphics[width=3.9cm,height=1.8cm]{Fig/supple_4x_results/davis_gt3.jpg} \\
\rotatebox{90}{ \ \textbf{\scriptsize{EMA-VFI}}}
&  
\includegraphics[width=3.9cm,height=1.8cm]{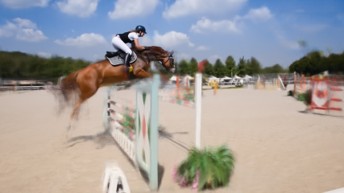} & \hspace{-0.15cm} 
\includegraphics[width=3.9cm,height=1.8cm]{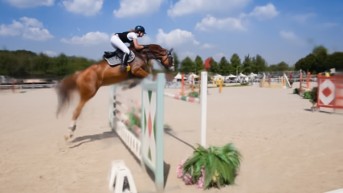} & \hspace{-0.15cm} 
\includegraphics[width=3.9cm,height=1.8cm]{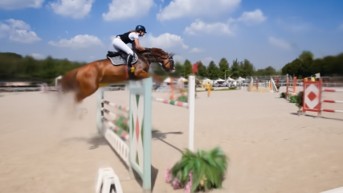} \\
\rotatebox{90}{ \ \textbf{\scriptsize{BiM-VFI}}}
&  
\includegraphics[width=3.9cm,height=1.8cm]{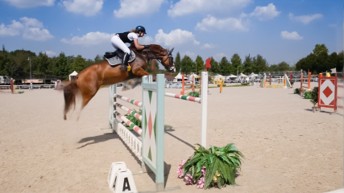} & \hspace{-0.15cm} 
\includegraphics[width=3.9cm,height=1.8cm]{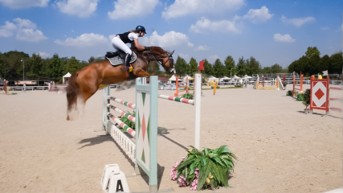} & \hspace{-0.15cm} 
\includegraphics[width=3.9cm,height=1.8cm]{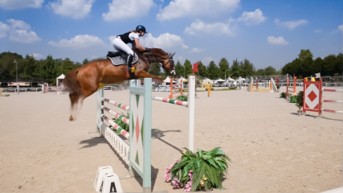} \\
\rotatebox{90}{ \quad \textbf{\scriptsize{OCAI}}}
&  
\includegraphics[width=3.9cm,height=1.8cm]{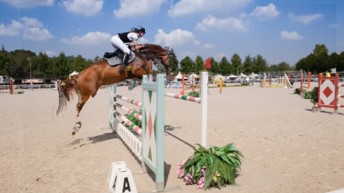} & \hspace{-0.15cm} 
\includegraphics[width=3.9cm,height=1.8cm]{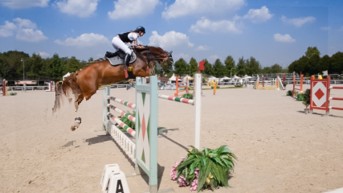} & \hspace{-0.15cm} 
\includegraphics[width=3.9cm,height=1.8cm]{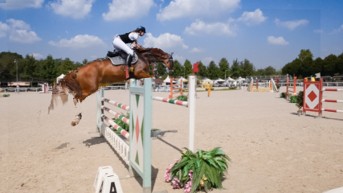} \\
\rotatebox{90}{ \ \  \textbf{\scriptsize{LDMVFI}}}
&  
\includegraphics[width=3.9cm,height=1.8cm]{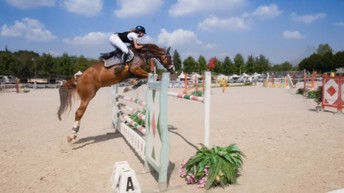} & \hspace{-0.15cm} 
\includegraphics[width=3.9cm,height=1.8cm]{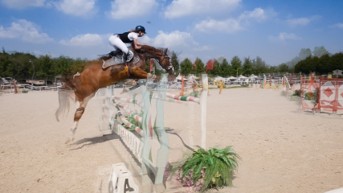} & \hspace{-0.15cm} 
\includegraphics[width=3.9cm,height=1.8cm]{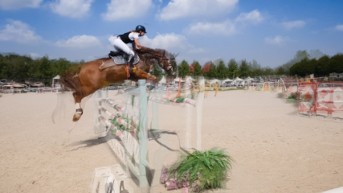} \\
\rotatebox{90}{ \quad \ \textbf{\scriptsize{GenIn}}}
& 
\includegraphics[width=3.9cm,height=1.8cm]{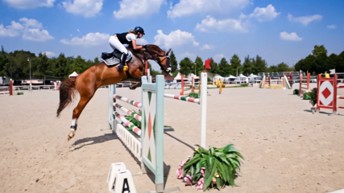} & \hspace{-0.15cm} 
\includegraphics[width=3.9cm,height=1.8cm]{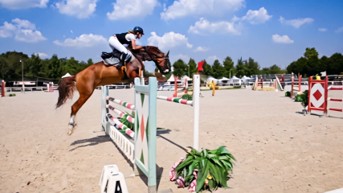} & \hspace{-0.15cm} 
\includegraphics[width=3.9cm,height=1.8cm]{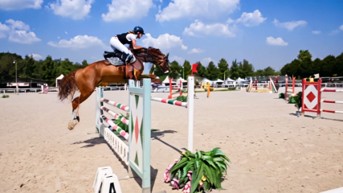} \\
\rotatebox{90}{ \ \textbf{\scriptsize{Quadratic}}}
&  
\includegraphics[width=3.9cm,height=1.8cm]{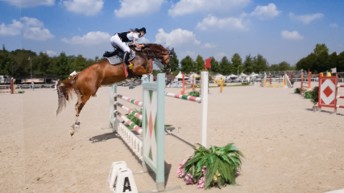} & \hspace{-0.15cm} 
\includegraphics[width=3.9cm,height=1.8cm]{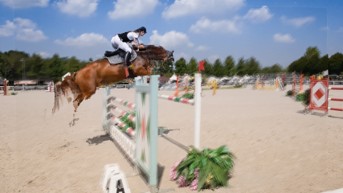} & \hspace{-0.15cm} 
\includegraphics[width=3.9cm,height=1.8cm]{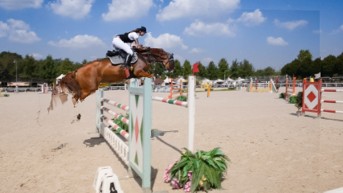} \\
\rotatebox{90}{\quad \ \ \textbf{\scriptsize{Ours}}}
& 
\includegraphics[width=3.9cm,height=1.8cm]{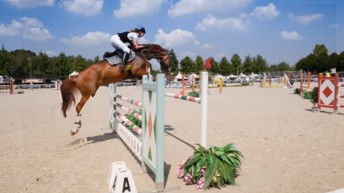} & \hspace{-0.15cm} 
\includegraphics[width=3.9cm,height=1.8cm]{Fig/supple_4x_results/davis_our2.jpg} & \hspace{-0.15cm}  
\includegraphics[width=3.9cm,height=1.8cm]{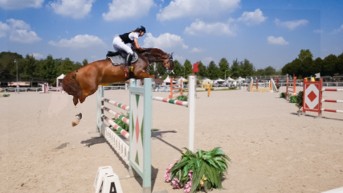} \\
\vspace{-18pt}
\end{tabular}$
\end{center}
\caption{Video frame interpolation results on DAVIS ($\times$4). The second to fourth rows show results from flow-based methods: EVA-VFI and Bim-VFI (backward warping) and OCAI (forward warping). The fifth and sixth rows present results from diffusion-based methods: LDMVFI and GenIn. The seventh row show the quadratic motion VFI results. The final row displays the output of our proposed \ours method. 
}
\vspace{-6pt}
\label{fig:experiment_vfi_davis4x}
\end{figure*}

Fig.~\ref{fig:experiment_vfi_sintel2x} presents the results for the Sintel dataset under the $\times$2 setting. Similar to KITTI, the right image demonstrates that other methods fail to accurately reproduce the ground-truth object. In contrast, our method generates a clean and faithful image.

Fig.~\ref{fig:experiment_vfi_davis2x} shows the results for the DAVIS dataset under the $\times$2 setting. Compared to other approaches, our method produces images with reduced blur and closer resemblance to the original frames. Interestingly, in the middle frame where occlusion makes image generation particularly challenging, GenIn produces a remarkably clean facial image. Furthermore, while GenIn performs well in generating the left and right frames, it introduces substantial variations in brightness and contrast. These artifacts, although visually appealing in some cases, can negatively affect quantitative metrics such as PSNR and SSIM.

In Fig.~\ref{fig:experiment_vfi_kitti4x} and~\ref{fig:experiment_vfi_davis4x} , we present additional qualitative results under the $\times$4 setting. Similar to the $\times$2 setting, most existing algorithms struggle to accurately reconstruct regions with large motion, often producing blurred results. In contrast, our method generates relatively precise and visually consistent images under such challenging conditions. Furthermore, we present results from quadratic motion–based VFI, where the trajectories $V_{0 \rightarrow t}$ and $V_{1 \rightarrow t}$ fail to converge at a same point. As a consequence, when the two images are combined, noticeable blur artifacts are introduced, which are particularly evident in the DAVIS dataset results.


\end{document}